%% file: main.tex
\documentclass[11pt,letterpaper]{article}
\pdfoutput=1
\usepackage[T1]{fontenc}
\usepackage{tcolorbox}

\AtBeginDocument{%
  \addtolength\abovedisplayskip{0.1\baselineskip}%
  \addtolength\belowdisplayskip{0.1\baselineskip}%
  \addtolength\abovedisplayshortskip{0\baselineskip}%
  \addtolength\belowdisplayshortskip{0\baselineskip}%
}

\usepackage{lmodern} \normalfont %
\usepackage{anyfontsize}
\DeclareFontShape{T1}{lmr}{bx}{sc} { <-> ssub * cmr/bx/sc }{}
\DeclareFontShape{T1}{lmr}{m}{scit}{ <-> ssub * cmr/m/sc }{}
\DeclareFontShape{T1}{lmr}{bx}{scit}{ <-> ssub * cmr/bx/sc }{}

\usepackage{etoolbox}
\usepackage{graphicx, color}

\usepackage{url}            %
\usepackage{booktabs}       %
\usepackage{amsfonts}       %
\usepackage{nicefrac}  
\usepackage{microtype}
\usepackage{graphicx}
\usepackage{booktabs} %
\usepackage{subcaption}
\usepackage{lipsum}
\usepackage{tkz-fct}
\input{headers}
\usetikzlibrary{patterns}
\usepackage{makecell}
\usepackage{arydshln}
\usepackage{thmtools}
\usepackage{thm-restate}
\usepackage[capitalize,noabbrev]{cleveref}
\usepackage{hyperref}
\hypersetup{
	linktocpage=true,
	colorlinks=true,				
	linkcolor=[rgb]{.7,0,0},				
	citecolor=[rgb]{0.5,0.0,0.13},				
	urlcolor=[rgb]{.7,0,.7},
}
\usepackage[tt=false]{libertine}
    \usepackage[libertine]{newtxmath}
    \usepackage[T1]{fontenc}
\usepackage{thm-restate}
\usepackage{amsmath,amssymb,array,booktabs,tabularx,colortbl,xcolor,caption}
\definecolor{darkblack}{rgb}{0,0,.75}
\usepackage{tocloft}
\title{
Should We Combine Our Data?\\ Algorithms and Limits for Privately Assessing Data Sharing}

\author{
\begin{tabular}{@{}c@{\hspace{4em}}c@{}}
Mahdi Haghifam &
Kumar Kshitij Patel \\[3pt]
{\small Toyota Technological Institute at Chicago} &
{\small Yale University}
\end{tabular}\\[2em]
\begin{tabular}{@{}c@{\hspace{2em}}c@{\hspace{2em}}c@{}}
Avrim Blum\textsuperscript{*} &
Giulia Fanti\textsuperscript{*} &
Juba Ziani\textsuperscript{*} \\[3pt]
{\small Toyota Technological Institute at Chicago} &
{\small Carnegie Mellon University} &
{\small Georgia Institute of Technology}
\end{tabular}
}
\date{}

\usepackage[letterpaper, margin=1in]{geometry}

\renewcommand{\epsilon}{\varepsilon}

\usepackage[colorinlistoftodos]{todonotes}
\usepackage{marginnote} %

\def\[#1\]{\begin{equation*}\begin{aligned}#1\end{aligned}\end{equation*}}
\def\*[#1\*]{\begin{align*}#1\end{align*}}

\let\originalleft\left
\let\originalright\right
\renewcommand{\left}{\mathopen{}\mathclose\bgroup\originalleft}
\renewcommand{\right}{\aftergroup\egroup\originalright}

\usepackage{caption}
\include{defs}
\include{defs-valuation}

\usepackage{cleveref}
\crefname{algorithm}{Algorithm}{algorithms}
\Crefname{algorithm}{Algorithm}{algorithms}

\newtheoremstyle{examplestyle}
  {\topsep}   %
  {\topsep}   %
  {\normalfont} %
  {}          %
  {\bfseries} %
  {.}         %
  {.5em}      %
  {}          %

\theoremstyle{examplestyle}

\newtheorem{myexample}{Example} 

\crefname{myexample}{example}{examples}
\Crefname{myexample}{Example}{Examples}

\renewcommand{\paragraph}[1]{\medskip\noindent{\bf #1}}

\begin{document}

\maketitle
\begingroup
\renewcommand{\thefootnote}{*}
\footnotetext[0]{The starred authors are listed in alphabetical order.}
\renewcommand{\thefootnote}{}
\footnotetext[0]{Part of this work was supported by and carried out while Mahdi Haghifam, Giulia Fanti, Kumar Kshitij Patel, and Juba Ziani were visiting the Simons Institute for the Theory of Computing. Juba Ziani was supported by NSF awards 2336236 (CAREER) and 2504990. Giulia Fanti was supported in part by NSF award CCF-2338772 (CAREER).}
\endgroup

\begin{abstract}

In this paper, we study when two parties should pool their data to train a machine learning model. Two agents, Alice and Bob, each hold a sensitive labeled dataset and want to train a classifier from a given hypothesis class $\mathcal{H}$. Even when each party's dataset can be perfectly classified on its own, the two parties may follow conflicting labeling standards or have selection biases that affect their features and label sets, such that no single classifier in $\mathcal{H}$ fits their combined data. Before committing to data sharing, Alice and Bob therefore want to test whether a single classifier in $\mathcal{H}$ can jointly fit both datasets---without revealing their individual data points to one another.

We formalize this problem from a learning-theoretic viewpoint via $\operatorname{opt}_\hypclass$, the minimum \emph{number of mistakes} made by any hypothesis in $\hypclass$ on their pooled data. Through interactive communication, the agents must distinguish $\operatorname{opt}_\hypclass=0$ (compatible datasets) from $\operatorname{opt}_\hypclass\geq t$ (conflicting datasets) under pure $\varepsilon$-two-party differential privacy. A trusted curator could do this with $t=O(1/\varepsilon)$, and we ask how small $t$ can be without one. For thresholds on the real line over a domain of size $m$, we show that the optimal $t$ is $\tilde{\Theta}(\log\log m/\varepsilon)$, exponentially smaller than the $\Theta(\log m/\varepsilon)$ samples needed to privately learn thresholds, even with a curator. For halfspaces in $\mathbb{R}^d$, the dependence on $m$ jumps to $\log(m)$: we give a protocol with $t=\tilde{O}(d^3\log m/\varepsilon)$ and show that some domains require $t=\Omega(d\log(m/d)/\varepsilon)$.
\end{abstract}

\section{Introduction}
Institutions often hold \emph{complementary} datasets that capture different aspects of the same downstream learning task. For example, a breast screening clinic may hold primarily negative cases, while a specialized diagnostic center may hold a much higher proportion of confirmed cancer cases \cite{lehman2017screening,sprague2017diagnostic}. 
Similarly, organizations trying to detect cybersecurity incidents frequently share data to provide a comprehensive view of the cybersecurity landscape, as attackers may target different organizations with different exploits at different times \cite{chang2023exploring}.
In such settings, pooling data may enable institutions to learn a useful classifier that no entity could learn on its own. 
Indeed, this potential was highlighted and analyzed in early theoretical works on distributed machine learning \cite{balcan2012distributed}.

Realizing the potential benefit of collaboration, however, often requires institutions to commit resources upfront through legal agreements, financial compensation, governance approvals, and dedicated infrastructure. Before incurring these costs, institutions need to assess the \emph{utility} of combining their datasets: will pooling improve model accuracy or other downstream metrics enough to justify the collaboration? Yet this assessment depends on joint properties of the datasets, generally hidden from each institution's local view. Institutions face a circular dilemma: they must judge whether collaboration is worthwhile before committing, yet doing so may require the very data sharing whose value they seek to assess.

Privacy requirements further complicate this dilemma. Hospitals, for example, must protect sensitive patient records even when sharing them would help evaluate a collaboration. Institutions may also hesitate to disclose valuable data to a prospective partner who could use them independently and decline to collaborate. When no trusted third party is available, each party must therefore assess whether collaboration is promising while limiting what it reveals about its data. 
This tension between the need to preserve data privacy while enabling agents to assess the utility of collaboration motivates the main conceptual question of this paper:  \emph{How can agents estimate whether it is beneficial to combine their datasets
while preserving the privacy of individual records? What algorithms can achieve this goal, and what are their fundamental limits?}

\subsection{How to Assess the Benefits of Data Sharing?}

In this section, we propose a learning-theoretic notion that agents can use to assess the value of data sharing and provide two real-world scenarios in which it can inform the decision to collaborate. Our focus throughout the paper is on binary classification on a data space (or feature space) $\dataspace\subseteq \Reals^d$ of size $m$ for some $d \geq 1$ and a fixed publicly known hypothesis class $\hypclass \subseteq \{\pm 1\}^{\dataspace}$. We consider a setting with two agents, Alice ($A$) and Bob ($B$) who respecitvely hold a labeled dataset $S_A, S_B \in (\dataspace \times \{\pm1\})^n$ of size $n \in \Naturals$.

We focus on a basic question that arises before joint training: can a single classifier in a given hypothesis class explain both parties' data? We formalize this question using their joint empirical loss defined as
\begin{equation}
    \opt_{\hypclass}(S_A,S_B)
    :=
    \min_{h\in\hypclass}
    \sum_{U\in\{A,B\}}
    \sum_{(x,y)\in S_U}
    \indic{h(x)\neq y}.
    \label{eq:opt-general}
\end{equation}
We use \(\opt_{\hypclass}(S_A,S_B)\), the number of mistakes made on the pooled data by the best classifier in \(\hypclass\), to assess the value of combining the data. 
It is zero when one hypothesis explains both datasets, and large when every hypothesis in $\hypclass$ makes many mistakes on their union. A large joint loss may indicate conflicting labeling practices, selection biases causing missing features and labels, or that the hypothesis class is not complex enough to fit the combined data. %

We therefore focus on distinguishing between the following: 
\begin{align}
\opt_{\hypclass}(S_A,S_B)=0 &  & \opt_{\hypclass}(S_A,S_B)\geq t
\label{eq:hypotheses}
\end{align}
for some $t>0$. When $\opt_{\hypclass}(S_A,S_B)=0$, Alice and Bob agree to collaborate. When $\opt_{\hypclass}(S_A,S_B) \geq t$, Alice and Bob must reassess\footnote{This could take the form of Alice and Bob realizing they do not have the expertise or computing power to fit a good model, in which case they refuse to collaborate; or, this could mean that Alice and Bob need to reconcile their labeling practices, jointly rectify their selection biases to account for missingness, potentially re-label data, or choose a more complex hypothesis class.}. Finally, if $0<\opt_{\hypclass}(S_A,S_B)<t$, we make no claims, as is standard in distribution testing. We now present two examples that further motivate the property-testing problem in \eqref{eq:hypotheses}.

\begin{myexample}[Learning from disparate label sets]\label{example:1}
Suppose Alice is a screening clinician and Bob is a diagnostician. They want to develop a classifier to determine whether a patient in a target population should be referred for further diagnostic testing. Their datasets reflect different stages of care: Alice's screening dataset $S_A$ mainly contains records of patients who were not referred, whereas Bob's diagnostic dataset $S_B$ contains only records of patients who were referred and received more thorough testing \cite{de2011verification}. Since the records contain sensitive patient information, neither agent is willing to disclose individual records to assess a possible collaboration, and no trusted third party is available. To isolate this selection bias, suppose Alice's labels are all $-1$, and Bob's are all $+1$. Then, neither can train a useful model alone; they must assess $\opt_{\hypclass}(S_A,S_B)$ to determine whether they can train a low-error model on their joint data. 
\end{myexample}

\begin{myexample}[Identifying conflicting labeling practices]\label{exp:2}
Now suppose Alice has labeled spam and non-spam emails in her dataset. However, her data may be limited or imbalanced, so she wants to augment it with Bob's curated, labeled email dataset. The challenge is that Bob may have a different view of what constitutes spam, so his labels could conflict with Alice's data \cite{klie2024analyzing}. For example, Alice may label unsolicited promotional emails as spam, while Bob uses that label for fraudulent or malicious messages. Before joint training, Alice wants to know whether a single classifier can achieve a sufficiently low test error under both labeling policies. To isolate conflicting labeling practices, suppose the agents encounter the same distribution of emails but may assign different labels to the same email. In this case, $\opt_{\hypclass}(S_A,S_B) = 0$ indicates that, under the hypothesis class $H$, Alice's and Bob's labeling practices are compatible, implying a collaboration may be worthwhile. On the other end, $\opt_{\hypclass}(S_A,S_B) \geq t$ points to differing labeling practices.
\end{myexample}
\begin{remark}[From in-sample to in-population error]
Note that our insights extend beyond the in-sample case and hold when considering generalization error over the underlying population distributions. We provide more details using standard arguments in Appendix~\ref{app:generalization}.
\end{remark}

\subsection{Problem Formalization}
In this section, we formalize our problem and state our main technical question. We first consider the disjoint-label setting as in \Cref{example:1}, where Alice holds only negatively labeled examples, and Bob holds only positively labeled ones. Since the labels are fixed by the party holding each record, we suppress them and write $S_A,\ S_B\in\dataspace^n$. This setting isolates the joint aspect of the problem: when $\hypclass$ contains the constant predictors, each local dataset is perfectly realizable, but their union need not be. Thus, the local optimal errors alone do not determine whether a common classifier can explain both datasets.
For our algorithms, we later show that they extend to datasets with arbitrary label mixtures (see \cref{remark:mixed-label-thr,remark:mixed-label-hs}). 

The definition of $\opt_\hypclass$ in \cref{eq:opt-general} in the disjoint-label setting can be rewritten as
\begin{equation} \label{def:opt}
\opt_{\hypclass}(S_A,S_B) \triangleq \min_{h \in \hypclass}\left\{ \sum_{x_A \in S_A}\indic{h(x_A)\neq -1} + \sum_{x_B \in S_B}\indic{h(x_B)\neq +1} \right\}.
\end{equation}

To assess a potential collaboration, the agents first agree on an error threshold $t\in[2n]$. They seek a private assessment algorithm that rejects with high probability when every classifier in $\hypclass$ makes at least $t$ mistakes on the pooled data, i.e., $\opt_{\hypclass}(S_A,S_B)\ge t$. We call this requirement \underline{soundness}. However, a trivial algorithm that always rejects also satisfies soundness. We therefore also require the algorithm to accept with high probability when the pooled data are jointly realizable, i.e., $\opt_{\hypclass}(S_A,S_B)=0$, a requirement we call \underline{completeness}. Finally, to protect the \underline{privacy} of individual records throughout the interaction, we use differential privacy~\citep{dwork2006calibrating} in its two-party form~\citep{beimel2008distributed,McGregorMironovPitassiReingoldTalwarVadhan10}, which requires each agent's view to be differentially private with respect to the other agent's dataset (see~\cref{def:two-party-dp}). We combine these pieces and formally define the two-party algorithm for dataset valuation as follows:

\begin{definition}[$\epsilon$-DP tester with resolution $t$]
\label{def:private-tester}
Fix a data space $\dataspace$, a hypothesis class $\hypclass\subseteq\{\pm1\}^{\dataspace}$, a dataset size $n\in\Naturals$, a privacy parameter $\varepsilon>0$, and $t\in[2n]$. Let $\Pi$ be an almost-surely terminating interactive two-party protocol on all pairs of finite multisets over $\dataspace$, with common output in $\{\mathsf{Match},\mathsf{Far}\}$. We interpret these outputs as acceptance or rejection of the proposed collaboration at this preliminary stage, respectively. We call $\Pi$ an \emph{ $\varepsilon$-private tester with resolution $t$} for $\hypclass$ at dataset size $n$ if it satisfies the following requirements:
\begin{enumerate}
    \item \emph{Privacy.}
    $\Pi$ satisfies $\varepsilon$-two-party differential privacy
    as in \cref{def:two-party-dp} for every input $S_A,S_B \in \dataspace^n$
    \item \emph{Soundness.}
    For every $S_A,S_B\in\dataspace^n$ s.t.  $ \opt_{\hypclass}(S_A,S_B)\geq t$, we have $ \Pr\left[\Pi(S_A,S_B)=\mathsf{Far}\right]\geq 0.9.$
    \item \emph{Completeness.}
    For every $S_A,S_B\in\dataspace^n$, s.t. $\opt_{\hypclass}(S_A,S_B)=0$, we have $\Pr\left[\Pi(S_A,S_B)=\mathsf{Match}\right]\geq 0.9.$
\end{enumerate}
The probabilities are over the protocol's private and public
randomness.
\end{definition}

We seek the smallest achievable resolution $t$, since smaller values let agents enforce stricter error tolerances. Without privacy constraints, the agents can exchange their datasets and compute $\opt_{\hypclass}(S_A,S_B)$ exactly, achieving $t=1$, the best possible resolution. A \emph{trusted curator} could also ensure differential privacy by releasing $\opt_{\hypclass}(S_A,S_B)+\mathrm{Lap}(1/\varepsilon)$, since the joint loss has sensitivity one. This gives a tester with resolution $t_{\text{central}}=O(1/\varepsilon)$, independent of the dimension and hypothesis class. However, we require information-theoretic two-party differential privacy without a trusted curator. This motivates the \emph{main technical question} that our paper seeks to answer:

\begin{tcolorbox}[
  colback=gray!5,
  colframe=black!65,
  colbacktitle=black!65,
  halign title=center,
  boxrule=1pt,
  arc=1mm,
  left=2mm, right=2mm,
  top=2mm, bottom=2mm,
]
\centering
\bfseries For fixed $\hypclass$ and $\epsilon$, what is the smallest resolution $t$ that an $\varepsilon$-DP private tester can guarantee?
\end{tcolorbox}

\subsection{Our Results}
We answer the above technical question for two fundamental classes of hypotheses: thresholds on the real line and half-spaces in $\Reals^d$. For each class, we establish both upper and lower bounds on the optimal resolution $t$ of any $\varepsilon$-private tester (summarized in Table~\ref{tab:valuation-results}). Notably, our main interactive protocols achieve a resolution $t$ that is \emph{completely independent of the dataset size $n$}, and all of our upper bounds extend immediately from the disjoint-label setting to arbitrary mixed-label datasets (see \cref{remark:mixed-label-thr,remark:mixed-label-hs}).

\subsubsection{Thresholds on Real Line}

We start with the simplest nontrivial class, thresholds on the real
line, where each hypothesis labels a point positive exactly when it
lies at or above a cutoff $\theta$:
\begin{equation}\label{eq:H_thr}
\hypclass_{\text{thr}}=\{h_\theta: \Reals \to \{\pm 1\}~|~ \exists\  \theta \in \Reals~\text{s.t.}~\text{$h_\theta(x)=1$~for $x\geq \theta$ and $h_\theta(x)=-1$ for $x<\theta$}\}.
\end{equation}
Our main result for this class is as follows:
\begin{theorem}[Informal Version of \cref{thm:threshold-upper-bound,thm:threshold-lower-bound}] \label{thm:informal-thr}
Fix $m, n \in \Naturals$ and $\epsilon \in (0,1]$. For every domain $\dataspace \subset \mathbb{R}$ with $|\dataspace|=m$, provided $n, t \gtrsim \frac{\log\log m}{\epsilon} \log\left(\frac{\log\log m}{\epsilon}\right)$, there exists an interactive, randomized, and polynomial-time algorithm that is an $\epsilon$-DP tester with resolution $t$ for $\hypclass_{\mathrm{thr}}$. Moreover, for every domain $\dataspace \subset \mathbb{R}$ with $|\dataspace|=m$, every $\epsilon$-DP tester with resolution $t\in[n]$ for $\hypclass_{\mathrm{thr}}$ requires $t \gtrsim \frac{\log\log m}{\epsilon}$.
\end{theorem}

Next, we mention a few implications of this result:
\begin{itemize}[leftmargin=*]
    \item \textbf{Minimal overhead compared to the centralized/non-private setting:} Recall that a trusted curator achieves resolution $O(1/\varepsilon)$ for every hypothesis class. In comparison, two-party differential privacy introduces only an $\tilde{O}(\log\log m)$ dependence on the domain size $m$, negligible in most practical settings.
    \item \textbf{Exponential improvement using an interactive protocol:} Interaction is provably required to achieve the near optimal $\Theta\left(\frac{\log\log m}{\epsilon} \log\left(\frac{\log\log m}{\epsilon}\right)\right)$ resolution: without interaction, any $\epsilon$-private tester requires $t \gtrsim \frac{\log m}{\epsilon}$, yielding an exponential separation between interactive and non-interactive private testing (see \cref{remark:no-interaction}).
    \item \textbf{Exponential separation between private testing and private learning:} A natural approach to the testing problem in \cref{def:opt} is to privately learn a good classifier and then privately evaluate its error on each dataset. However, any pure $\epsilon$-DP threshold learner, whether proper or improper, requires $\Omega\left(\frac{\log m}{\epsilon}\right)$ samples at constant accuracy and confidence, even when a trusted curator holds the pooled data~\citep{beimel2013private,feldman2014sample}. In contrast, \cref{thm:informal-thr} gives a private tester using only $O\left(\frac{\log\log m}{\epsilon} \log\left(\frac{\log\log m}{\epsilon}\right)\right)$ samples, yielding an exponential separation in the dependence on $m$. Qualitatively similar separations arise in non-private settings, where testing or estimating optimal prediction error can require fewer samples or label queries than learning a predictor~\citep{kearns2000testing,balcan2012active,blum2018active,kong2018estimating}.
\end{itemize}

\subsubsection{Halfspaces in Higher Dimension}
The next fundamental hypothesis class we consider is halfspaces in $\Reals^d$ defined as follows:
\begin{equation}\label{eq:H_hs}
\hypclass_{\text{hs}}=\{h_{w}: \Reals^d \to \{\pm 1\}~|~ \exists\ w \in \Reals^{d}~\text{s.t.}~\text{$h_{w}(x)=\text{sign}\left(\inner{w}{x}\right)$}\}.
\end{equation}

Our main result for this class is as follows:

\begin{theorem}[Informal Version of \cref{thm:halfspace-upper-bound,thm:hs-lower-bound,thm:hs-sample-size}]\label{thm:informal-hs}
Fix $d \ge 2$, $m,n \in \Naturals$, and $\epsilon \in (0,1]$. For every domain $\dataspace \subset \mathbb{R}^d$ with $|\dataspace|=m$, provided $n,t \gtrsim \frac{d^2 \log m}{\epsilon}\left(d+\log\log(m)\right)$, there exists an interactive $\epsilon$-DP private algorithm with resolution $t$ for $\hypclass_{\mathrm{hs}}$. 

Moreover, for every sufficiently large $d$ and every $m\gtrsim d$, there exists a public domain
$\dataspace_{d,m}\subseteq\Reals^d$ with
$|\dataspace_{d,m}|=m$, chosen only from $(d,m)$, such that for every sufficiently small $\epsilon>0$ and every $n\in\Naturals$ with $n\epsilon\gtrsim1$, every
$\epsilon$-private tester over $\dataspace_{d,m}$ with resolution $t\in[n]$ requires
$
t\gtrsim
\frac{\min\{d,\sqrt{n\epsilon}\}}{\epsilon}
\log\left(
2+\min\left\{\frac md,\frac{n\epsilon}{d^2}\right\}
\right).
$
\end{theorem}
Next, we mention a few implications of this result:
\begin{itemize}[leftmargin=*]
    \item \textbf{Sharp jump in domain-size dependence:} Comparing \cref{thm:informal-thr} and \cref{thm:informal-hs} reveals that, as we move from one-dimensional thresholds ($d=1$) to halfspaces in any fixed dimension $d\ge d_0$ for some constant $d_0$, the dependence of the testing resolution $t$ on the domain size $m$ jumps exponentially from $\log\log m$ to $\log m$. Our lower bound confirms that this $\log m$ dependence is tight up to secondary logarithmic factors in the stated parameter regime.
    \item \textbf{Dimension dependence despite a one-bit output:} Our lower bound implies that, even though $\opt_\hypclass(S_A,S_B)$ is a scalar, two-party pure DP testing incurs a resolution error that scales with $d$, whereas a centralized algorithm incurs only $t_{\text{central}} \;=\; O\left(\frac{1}{\varepsilon}\right)$. This gap reflects the privacy cost of decentralization, as the agents must replace a trusted curator with a protocol that protects each party's records from the other throughout their interaction.

    \item \textbf{The privacy cost even with abundant data:} Even with arbitrarily large datasets, two-party private testing can require resolution $\Omega(d\log(m/d)/\epsilon)$ on the same fixed public domain. Thus, increasing the number of records does not eliminate this worst-case cost in absolute resolution, although the corresponding fraction of errors decreases.
\end{itemize}

\begin{figure}
    \centering
\includegraphics[width=0.7\linewidth]{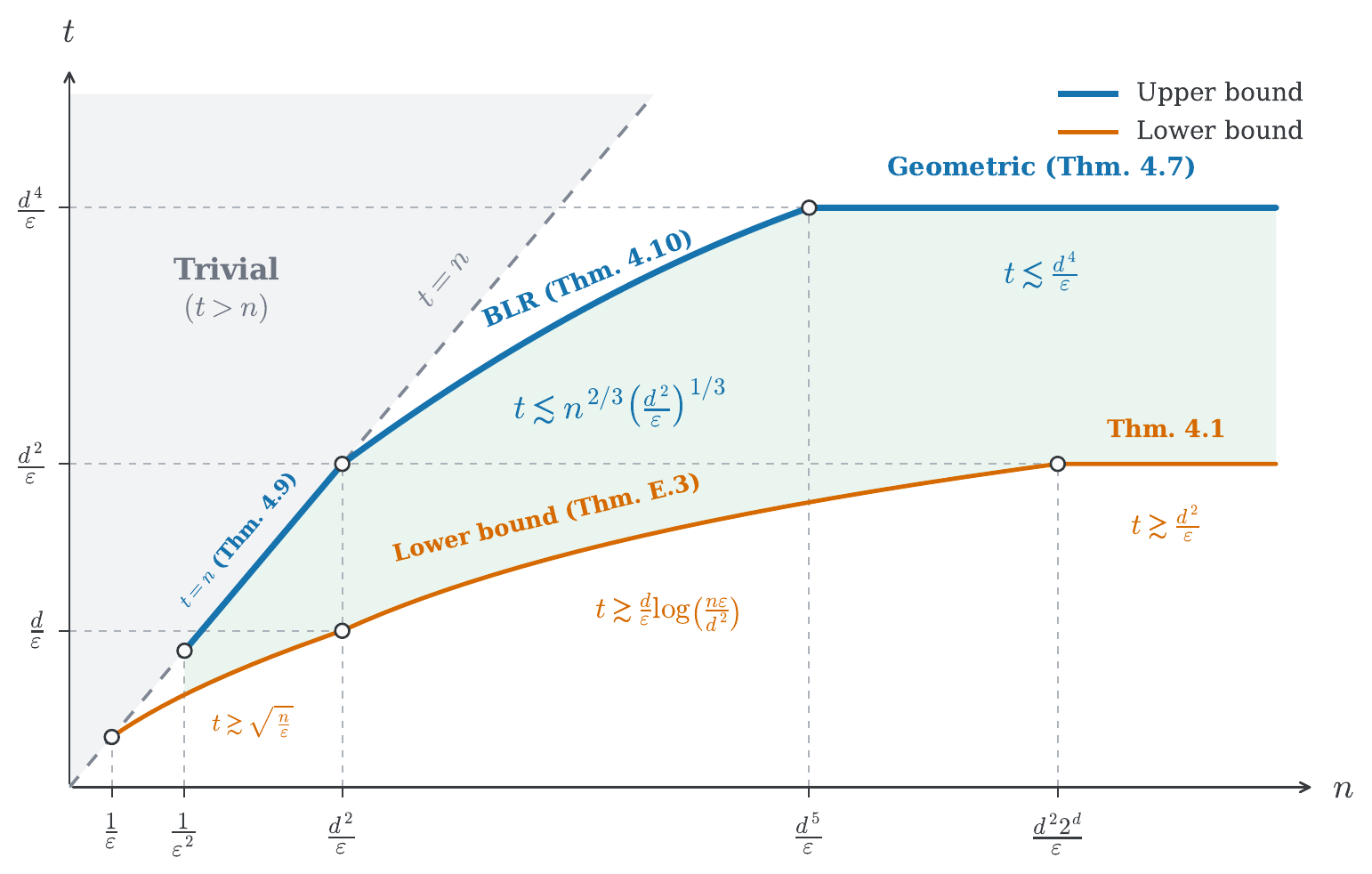}
    \caption{\textbf{Upper and lower bounds on the resolution $t$ for private halfspace testing as a function of dataset size $n$, illustrated for $m = 2^d$} (for $d > 1/\varepsilon$, hiding constant and $\log d$ factors). 
Since $\mathrm{opt}_{\mathcal{H}_{\mathrm{hs}}}(S_A,S_B) \le n$, the region $t > n$ is trivial; the upper bound (blue) starts at $n \asymp \frac{1}{\varepsilon^2}$ with $t = n$ (\cref{thm:hs-t-equals-n-mb}), transitions to the synthetic-data bound at $n \asymp \frac{d\log m}{\varepsilon} = \frac{d^2}{\varepsilon}$ (\cref{thm:halfspace-query-release}), and reaches the $n$-independent geometric bound at $n \asymp \frac{d^4\log m}{\varepsilon}$ (\cref{thm:halfspace-upper-bound}). 
The lower bound (orange) starts at $n \asymp \frac{1}{\varepsilon}$ (where $t \asymp \frac{1}{\varepsilon}$) and grows across two regimes (\cref{thm:hs-sample-size}) before saturating at $\frac{d\log(m/d)}{\varepsilon}$ for $n \gtrsim \frac{md\log(m/d)}{\varepsilon}$ (\cref{thm:hs-lower-bound}).}
    \label{fig:placeholder}
\end{figure}

\paragraph{Complementary bounds for smaller datasets.} An interesting aspect of both of our main results is that the achievable resolution $t$ does not depend on $n$. For halfspaces, however, our main tester from \cref{thm:informal-hs} provides a nontrivial guarantee only when $n\gtrsim  \frac{d^2 \log m}{\epsilon}\left(d+\log\log(m)\right)$. For smaller datasets, we complement it with two simple testers whose resolution grows with $n$. The first one is an application of the net mechanism of \cite{blum2013learning}: each party privately releases a small synthetic dataset, which gives a non-interactive tester with resolution $t=O\bigl((n^2d\log(2m)/\epsilon)^{1/3}\bigr)$ (see \cref{thm:halfspace-query-release}). The second one targets the coarsest nontrivial resolution, i.e.,  $t=n$. It turns out that $\opt_{\hypclass_{\text{hs}}}(S_A,S_B)=n$ happens only when Alice and Bob have the same number of records on every ray through the origin, and this can be tested with shared randomness as long as $n\gtrsim 1/\epsilon^2$ (see \cref{thm:hs-t-equals-n-mb}). Putting the three testers together, for every $n\gtrsim 1/\epsilon^2$ we get (ignoring smaller logarithmic factors),
\[
t=O\left(\min\left\{\frac{d^2\log(2m)}{\epsilon}
\left(d+\log(\log(2m))\right),\ \left(\frac{n^2d\log(2m)}{\epsilon}\right)^{1/3},\ n\right\}\right).
\]
The $t=n$ tester is the best of the three when $n\lesssim d\log(2m)/\epsilon$, the synthetic-data tester when $d\log(2m)/\epsilon\lesssim n\lesssim d^4\log(2m)/\epsilon$, and the geometric protocol for larger $n$; see \cref{fig:placeholder}.

\begin{table*}[t]
\small
\centering
\begingroup
\setlength{\tabcolsep}{4pt}
\setlength{\extrarowheight}{7pt}
\renewcommand{\arraystretch}{0.9}
\captionsetup{
    font=small,
    labelfont=bf,
    justification=raggedright,
    singlelinecheck=false,
    skip=9pt
}
\begin{tabular*}{\textwidth}
{@{\extracolsep{\fill}}ccc@{}}
\toprule
\textbf{Class}
&
\textbf{Upper bound}
&
\textbf{Lower bound}
\\[6pt]
\midrule

\textbf{Thresholds in $\Reals$}
&
{\footnotesize
$\displaystyle
\frac{\log\log m}{\epsilon}
\log\left(\frac{\log\log m}{\epsilon}\right)
$}
&
{\footnotesize
$\displaystyle
\frac{\log\log m}{\epsilon}
$}
\\[12pt]

\textbf{Halfspaces in $\Reals^d$}
&
{\footnotesize
$\displaystyle
\min\left\{
\frac{d^2\log(2m)}{\epsilon}
\left(d+\log(\log(2m))\right),\
\left(\frac{n^2d\log(2m)}{\epsilon}\right)^{1/3},\
n
\right\}
$}
&
{\footnotesize
$\displaystyle
\frac{\min\{d,\sqrt{n\epsilon}\}}{\epsilon}
\log\left(
2+\min\left\{
\frac{m}{d},\frac{n\epsilon}{d^2}
\right\}
\right){}^{\dagger}
$}
\\[12pt]

\bottomrule
\end{tabular*}

\par\vspace{7pt}
\caption{Bounds on the error gap $t$ for pure $\epsilon$-two-party DP testing with success probability at least $0.9$, up to universal constant factors. $\dagger$~The halfspace lower bound holds for some input domains, $d\geq d_0$, $m\gtrsim d$, and sufficiently small $\epsilon$.}
\label{tab:valuation-results}
\endgroup
\end{table*}

\subsection{Technical Overview}

In this section, we outline the main ideas behind our lower bounds and private testing algorithms. We first explain in \Cref{sec:overview_lower} the idea behind our lower bound, which is based on the connection between $\opt_\hypclass$ and error-correcting codes. We then describe in \Cref{sec:overview_algo_threshold}, our threshold algorithm  which uses private binary search over prefix lengths to compare two ordered blocks of data. Finally, in \Cref{sec:overview_algo_halfspace}, we explain the ideas behind our algorithm for halfspaces, which is based on the cutting-plane method in the dual space, i.e., classifier space.

\subsubsection{Lower Bounds: Coding Against Erasures}\label{sec:overview_lower}
Our lower bounds use the connection between information complexity and DP~\citep{McGregorMironovPitassiReingoldTalwarVadhan10}. We consider a communication problem in which Alice and Bob receive secrets $v_A,v_B\in\mathcal{V}$ in a finite domain $\mathcal{V}$, and exchange messages to compute a binary function $F:\mathcal{V}\times\mathcal{V}\to\{0,1\}$. For a protocol computing $F$ on inputs $(V_A,V_B)\sim\mu$, let $\Pi$ denote its public transcript, including public randomness, exchanged messages, and final output. 
The protocol's internal information cost is defined as $\operatorname{IC}_{\mu}(\Pi,F):=I(V_A;\Pi\mid V_B)+I(V_B;\Pi\mid V_A)$. Intuitively, it measures what the transcript reveals about each input, given the other.

To force a private tester to incur a large information cost, we choose $F$ and $\mu$ such that every protocol $\Pi$ computing $F$ with bounded error has large $\operatorname{IC}_{\mu}(\Pi,F)$. To relate the communication problem to our problem, we construct local encoders $E_A,E_B:\mathcal{V}\to\mathcal{X}^n$, producing datasets $S_A=E_A(v_A)$ and $S_B=E_B(v_B)$ that satisfy
\begin{align*}
F(v_A,v_B)=0 \implies \opt_\hypclass(S_A,S_B)=0,\qquad \text{and}\qquad F(v_A,v_B)=1 \implies \opt_\hypclass(S_A,S_B)\geq t.
\end{align*}
Running a private tester with resolution $t$ on these datasets therefore yields a bounded-error protocol $\Pi$ for $F$. To capture the sensitivity of the encoders to the secrets, we define the encoding diameters $\Delta_U(n,t):=\max_{v,v'\in\mathcal{V}}\mathrm{d}_{\mathrm{ham}}(E_U(v),E_U(v'))$ for $U\in\{A,B\}$, and write $\Delta=\Delta(n,t):=\Delta_A(n,t)+\Delta_B(n,t)$. The delicacy is that we need to design an encoding function that relates the encoding diameters to $n,t$. Then, we can invoke existing results \citep{McGregorMironovPitassiReingoldTalwarVadhan10} (also see \cref{lem:metric-dp-to-information}) that show that $\operatorname{IC}_{\mu}(\Pi,F)\le O(\varepsilon\Delta)$. Comparing this upper bound with the information required to compute $F$ forces $\Delta$ to be large. Our encoders upper-bound $\Delta$ in terms of the loss gap $t$, turning this requirement into a lower bound on $t$.

To design encoders for Alice and Bob, we link joint empirical loss to robustness against record erasures. For any hypothesis class $\hypclass$, $\opt_\hypclass(S_A,S_B)$ equals the minimum number of records that must be removed from the pooled data to make it realizable. Thus, \textbf{ensuring $\opt_\hypclass(S_A,S_B)\geq t$ requires the pooled data to remain non-realizable after removing any set of fewer than $t$ records.} This suggests using error-correcting codes to make non-realizability robust to erasures while keeping the encoding diameter small. Repetition suffices for thresholds, while our halfspace construction uses an expander code.

For \textbf{thresholds} on a domain $\dataspace\subset\mathbb{R}$ of size $m$, Alice and Bob hold inputs $v_A,v_B\in\mathcal{V}=\dataspace$. We use the \emph{Greater-Than} predicate $F(v_A,v_B)=\mathbf{1}\{v_A\ge v_B\}$. Alice creates $t$ copies of $(v_A,-1)$, and Bob creates $t$ copies of $(v_B,+1)$. The pooled data are realizable when $v_A<v_B$, whereas every threshold misclassifies an entire block when $v_A\ge v_B$. This gives a loss gap of $t$ with encoding diameter $\Delta=O(t)$. Since Greater-Than has information complexity $\Omega(\log\log m)$ under a suitable input distribution~\citep{BravermanWeinstein12}, we use prior results connecting DP to information complexity \cite{McGregorMironovPitassiReingoldTalwarVadhan10} (\Cref{lem:metric-dp-to-information}) to compare
this lower bound with an $O(\varepsilon t)$ privacy bound, yielding $t=\Omega(\log\log(m)/\varepsilon)$. 

For \textbf{halfspaces}, Alice and Bob hold vectors $v_A,v_B\in\mathcal{V}=\mathbb{F}_q^k$, where $k=\Theta(d)$ and $q$ is a power of two. Our encoders map these inputs to labeled data points in a public feature domain $\mathcal{X}\subset\mathbb{R}^d$ of size $m$. We use the \textit{trace inner-product} predicate $F(v_A,v_B)=\operatorname{Tr}_{\mathbb{F}_q/\mathbb{F}_2}(\sum_{i=1}^k v_{A,i}v_{B,i})$, whose information complexity is $\Omega(k\log q)$~(\Cref{lem:ic-trace}). To construct the datasets, we view halfspace consistency as a linear feasibility problem, with classifier parameters as variables and one constraint per labeled record. We use this correspondence in reverse, building an LP and converting its constraints into records. For each $i\in[k]$, Alice supplies one input-dependent, or \emph{private}, constraint fixing the LP representation of $v_{A,i}$, and Bob does the same for $v_{B,i}$. The input-independent \emph{public} constraints force an output variable $z$ to equal $F(v_A,v_B)$ once the private constraints fix the inputs. They also require $z=0$, making the LP feasible iff $F(v_A,v_B)=0$.

To obtain a loss gap of $t$, we could repeat every constraint $t$ times, as for thresholds. However, the resulting diameter $\Delta=O(kt)$ cancels the dimension factor in the information comparison $k\log q\lesssim\varepsilon kt$ (see~\Cref{sec:hs-lower-overview}). We instead encode $v_A$ and $v_B$ using an expander code with $\Theta(k)$ symbols and constant-size checks~\citep{feldman2005using,feldman2007lp}. Private constraints fix the codeword symbols, while public constraints enforce the code checks. Our key step shows that, with public constraints enforced, fewer than $D=\Theta(d)$ private deletions cannot change the codewords, even through fractional LP solutions. Indeed, unique-neighbor expansion gives any nonempty set of fewer than $D$ deviating symbols a check touching exactly one of them; the unchanged symbols force it back to its original value. Thus, repeating each private constraint $\lceil t/D\rceil$ times and each public constraint $t$ times prevents fewer than $t$ record deletions from restoring realizability when $F(v_A,v_B)=1$. Only private copies affect the diameter, giving $\Delta=O(d\lceil t/D\rceil)=O(t+d)$. Finally, we represent field elements as polygon vertices in $\mathbb{R}^2$; restricting each symbol to this polygon lets one supporting inequality fix its value. Together with sparse checks, this keeps the dimension $O(d)$ and the universe size $O(dq^{O(1)})$. Choosing $q=(m/d)^{\Theta(1)}$ and applying the information bound yields $t=\Omega(d\log(m/d)/\varepsilon)$ under the parameter conditions of \Cref{thm:hs-lower-bound}.

\subsubsection{Algorithm for Thresholds: A Robust and Private Greater-Than Protocol}\label{sec:overview_algo_threshold}

\begin{figure} 
  \centering
 \includegraphics[width=4.5in]{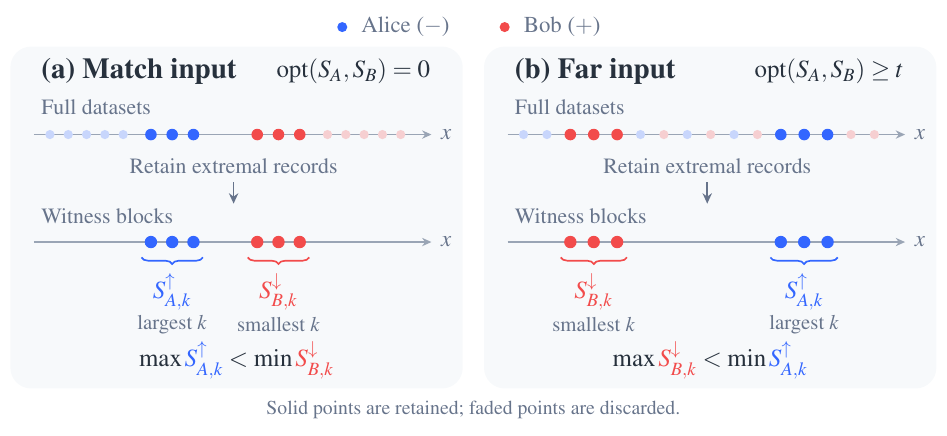}
 \caption{Reduction to witness blocks for thresholds with disjoint supports. Alice retains her largest $k=\lfloor t/2\rfloor$ records, and Bob retains his smallest $k$ records. Solid points are retained and used by our algorithm; faded points are discarded and never be used by our algorithm. In $\mathsf{Match}$, Alice's retained block lies entirely below Bob's; in $\mathsf{Far}$, their order is reversed. The illustration uses $n=8$, $t=6$, and $k=3$. 
 } \label{fig:threshold-fig}
\end{figure}

We describe the algorithm in the simpler setting where Alice holds only negative records, Bob holds only positive records, and their feature supports are disjoint (\Cref{lem:far-crossing} handles overlapping supports, and \Cref{remark:mixed-label-thr} extends the algorithm to mixed labels.) Also, recall that thresholds label points below their respective cutoffs as negative and those at or above as positive (see~\cref{eq:H_thr}). For thresholds with $|\dataspace|=m$ and $t\ge 2$, we reduce the promise problem to comparing two witness blocks of $k=\lfloor t/2\rfloor$ records from the size-$n$ datasets (\cref{fig:threshold-fig}). We prove a simple structural result that implies that if $\opt_{\hypclass_{\text{thr}}}(S_A,S_B)=0$, then all of Alice's points are smaller than all of Bob's points, and if $\opt_{\hypclass_{\text{thr}}}(S_A,S_B)\geq t$, then every retained point of Alice exceeds every retained point of Bob. See \cref{fig:threshold-fig} for intuition. 
Therefore, $k$ points are sufficient to distinguish between $\mathsf{Match}$ and $\mathsf{Far}$. Note that if $0 < \opt_{\hypclass_{\text{thr}}} < t$, both answers can happen, as is standard in distribution testing.

Given this reduction, a natural idea is for Alice and Bob to compare their blocks using a privately chosen point. Alice applies the exponential mechanism over the full public domain $\mathcal{X}$, favoring points near her block's median, and sends her choice to Bob, who privately counts how many of his records fall below it. Selecting among $O(m)$ candidates incurs score error $O(\log(m)/\epsilon)$. Taking $k$ large enough to dominate this error and Bob's counting noise ensures, with constant failure probability, that Alice's choice lies within her block's range and Bob's comparison identifies the correct order. This yields resolution $O(\log(m)/\epsilon)$, but leaves a gap with respect to our $\Omega\bigl(\log\log(m)/\epsilon\bigr)$ lower bound.

\begin{figure} 
\centering \includegraphics[width=0.8\textwidth]{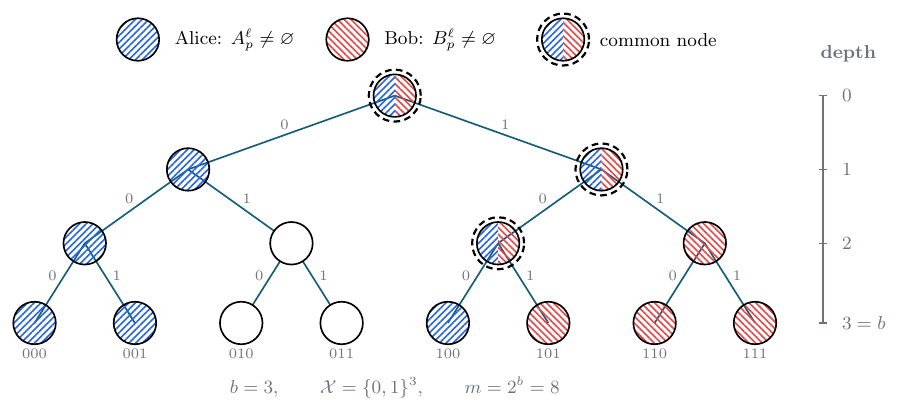} \caption{\textbf{A binary trie for the ordered witness blocks.} A node $p$ at depth $\ell$ represents a prefix; $A_p^\ell$ and $B_p^\ell$ contain the records in Alice's and Bob's blocks with that prefix. Blue and red shading indicate Alice's and Bob's records, respectively, and dashed circles mark the common nodes shared by both. Because the blocks are strictly ordered, at most one node at each depth is common.} \label{fig:tree} 
\end{figure}

To compare the blocks with a better dependence on $m$, we build on the standard Greater-Than protocol for comparing two numbers~\citep[Chapter~3]{rao2020communication}. We adapt its search over prefix lengths to handle blocks of records while preserving privacy. Specifically, we encode each point's rank in the public domain as a binary string of length $b=O(\log m)$ and organize these strings into a binary trie; see \cref{fig:tree}. Each node at depth $\ell$ contains the strings sharing a particular prefix of length $\ell$. When each party holds a single distinct string, their first differing bit determines their order. Their prefixes agree before that bit and disagree from that bit onward, so binary search locates it using $O(\log b)=O(\log\log m)$ prefix-equality tests. At that position, the string with bit $0$ precedes the string with bit $1$.

For blocks of records, however, the first separation may involve too few points to overcome privacy noise. We therefore search for a depth at which sufficiently many records have left the shared prefix. To measure this, let $A_p^\ell$ and $B_p^\ell$ contain the records in the respective blocks with prefix $p\in\{0,1\}^{\ell}$, and define $M(\ell):=\max_p\min\{|A_p^\ell|,|B_p^\ell|\}$ and $F(\ell):=k-M(\ell)$. Since the blocks are strictly ordered, at most one node at each depth contains records from both parties. When this common node exists, $F(\ell)$ is the larger of the two counts of records outside it. As each child contains only records from its parent, $M(\ell)$ cannot increase with depth; consequently, $F(\ell)$ is nondecreasing from $F(0)=0$ to $F(b)=k$. We can therefore use noisy binary search to find adjacent depths $L,H=L+1$ whose values of $F$ bracket a target level $\alpha$ up to the estimation error, with $\alpha$ chosen well above that error.

To estimate $F(\ell)$ without incurring the unwanted $\log m$ cost, we estimate $M(\ell)$ through a small hashed histogram. A direct noisy histogram over all $2^\ell$ prefixes would incur an error that grows with $\ell$, but at most $2k$ prefixes are occupied. The parties therefore use the same public hash into $O(k^2/\beta_0)$ buckets, which keeps distinct occupied prefixes in different buckets with probability at least $1-\beta_0/2$. They then add Laplace noise to their bucket counts and take the maximum over buckets of the smaller noisy count. With privacy budget $\epsilon_0$ per party, this estimates $M(\ell)$, and hence $F(\ell)$, to error $O(\log(k/\beta_0)/\epsilon_0)$ with failure probability at most $\beta_0$. The error thus depends logarithmically on $k$ and $1/\beta_0$, rather than on the full prefix universe.

These estimates locate a useful depth, but we must still determine which block comes first. If the estimated increase from $F(L)$ to $F(H)$ is large, the common node at depth $L$ splits enough records between its two children to reveal the order. Otherwise, $F(H)$ is already of order $\alpha$, while the common node at depth $H$ still contains most of each block. The records accumulated outside this node then reveal the order: in $\mathsf{Match}$, Alice's lie to its left and Bob's to its right, with these directions reversed in $\mathsf{Far}$. In both cases, each party computes a comparison statistic from weighted maxima of counts. These statistics have constant sensitivity even when their maximizing prefixes change, and their difference distinguishes the two orders with a margin of $\Omega(\alpha)$. We can therefore add noise without identifying the common prefix itself.

The search uses only $T=\lceil\log_2 b\rceil=O(\log\log m)$ adaptive queries. We give each query a privacy budget $\epsilon_0=\epsilon/(2T)$ and a failure budget $\beta_0=\beta/(2T)$, leaving half of each party's privacy budget and half of the total failure budget for the final comparison. With these allocations, a margin $\alpha=O\bigl((T/\epsilon)\log(kT/\beta)\bigr)$ suffices to dominate the noise in both stages. Taking $k\ge 4\alpha$ makes the final comparisons reliable; since $k=\lfloor t/2\rfloor$, this condition holds when $t$ is a sufficiently large constant multiple of $(T/\epsilon)\log(T/(\epsilon\beta))$. Thus, at constant failure probability, we obtain an $\epsilon$-private tester with resolution $t=\tilde{O}\bigl(\log\log(m)/\epsilon\bigr)$, independent of the original dataset size $n$, provided $n\ge t$.

\subsubsection{Algorithm for Halfspaces: Cutting Plane in the Dual Space}\label{sec:overview_algo_halfspace}
\begin{figure}
\centering
\includegraphics[width=0.65\textwidth, height=0.4\textheight, keepaspectratio]{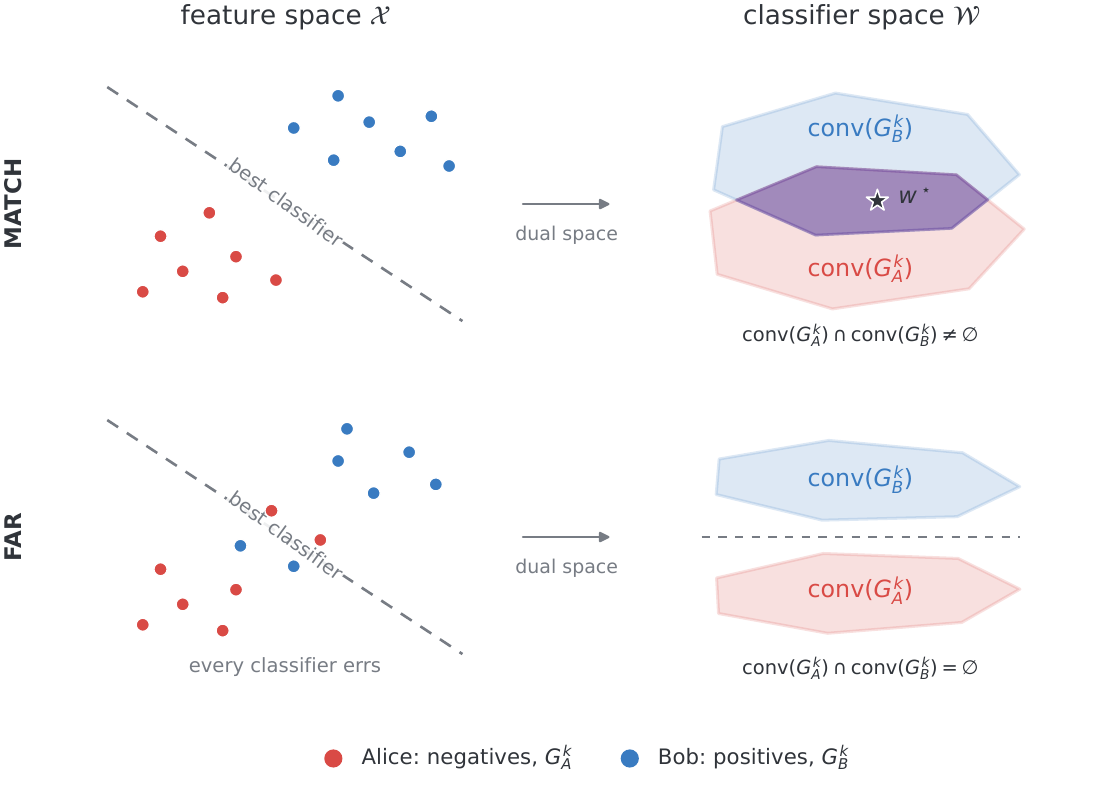}
\caption{The halfspace promise problem viewed in the dual 
space.  Our algorithm works in the dual space and tests which case holds.}
\label{fig:dual-space}
\end{figure}
For the class of halfspaces in $\mathbb{R}^d$ with
$|\dataspace|=m$, our main idea is to move to the dual space,
where each classifier $h_w$ is represented by a point
$w\in\mathbb{R}^d$.
Since only the predictions on the public domain matter, the
Sauer--Shelah lemma allows us to restrict attention to a finite
public set of candidate classifiers
$\mathcal{W}_{\mathrm{disc}}$ of size at most $m^d$.

Let $G_A^k$ and $G_B^k$ denote the candidates that make at most
$k=\lfloor (t-1)/(2(d+1))\rfloor$ mistakes on Alice's and Bob's
data, respectively. Unlike the ordered witness blocks for thresholds, these sets of
low-error classifiers can be very unstructured.
Our key observation is that their convex hulls distinguish the
two cases.
In $\mathsf{Match}$, the sets share a consistent classifier
$w^\star$.
In $\mathsf{Far}$, even their convex hulls are disjoint:
\[
    \mathrm{conv}(G_A^k)\cap\mathrm{conv}(G_B^k)=\emptyset.
\]
To see why, suppose a point belongs to both convex hulls.
By Carath\'eodory's theorem, it is a convex combination of at
most $d+1$ low-error classifiers from each party.
A convex combination can misclassify a record only if at least
one of its constituent classifiers misclassifies that record.
Thus, this point would define a classifier with at most
$2(d+1)k<t$ mistakes on the pooled data, contradicting
$\mathsf{Far}$.

This geometric separation suggests an iterative elimination
algorithm, inspired by \cite{braverman2019convex}.
The parties maintain a public set of active candidates.
In $\mathsf{Far}$, a hyperplane separates the convex hulls of
their remaining low-error classifiers.
At least one side of this hyperplane contains at most half of
the active candidates.
Thus, at least one party has a halfspace cut that retains all
of its low-error classifiers while discarding at least half
of the candidates.

The next challenge is to choose such a cut privately.
Searching directly over all possible halfspace cuts would be costly in terms of privacy.
Instead, we use \emph{halfspace containers}
\citep{braverman2019convex,dutta2021uniform} to obtain a small
public menu of $2^{O(d)}$ subsets of the active candidates.
Whenever a halfspace contains at most half of the active
candidates, one of these containers covers its candidates
while containing at most $3/4$ of the active set.
We can therefore search over this small menu while still
eliminating a constant fraction of the candidates.

Alice and Bob use the exponential mechanism with a low-sensitivity cost to select a container or abstain.
With high probability, at least one party selects a shrinking
container in each $\mathsf{Far}$ round, while every selected
container preserves $w^\star$ in $\mathsf{Match}$.
Thus, within $O(d\log m)$ rounds, the active set becomes empty
in $\mathsf{Far}$ and remains nonempty in $\mathsf{Match}$.

Considering all the factors, we can show this algorithm is an $\epsilon$-private tester with
$t=\tilde{O}\bigl(d^3\log(m)/\epsilon\bigr)$. Intuitively, the resolution $\tilde{O}(d^3\log(m)/\varepsilon)$ factors as
$\frac{d\log m}{\varepsilon}\cdot d\cdot d$, where $\frac{d\log m}{\varepsilon}$
arises from splitting the privacy budget across $O(d\log m)$
elimination rounds, the second factor of $d$ is the entropy of the containers that we need to pay due to exponential mechanism,
and the third factor of $d$ comes from Carath\'eodory theorem when relating the global mistake gap $t$ to the local error threshold $k=\Theta(t/d)$.

\subsection{Related Work}
\label{sec:related-work}
Our work lies at the intersection of three lines of research:
communication-efficient distributed learning, testing whether a good model
exists without learning one, and data valuation. 

\paragraph{Distributed Communication-Efficient Learning.}
A line of work studies communication-efficient distributed learning \citep{balcan2012distributed,chen2016communication,kane2019communication}. Broadly speaking, in the two-party setting, each agent holds a set of labeled examples, and the players communicate to learn a classifier with small error. In much of this work, the algorithms communicate with each other using samples, and the goal is to minimize the number of samples communicated.  However, \cite{balcan2012distributed} also propose and analyze the more general setting in which communication can be arbitrary, and the goal is to minimize the total number of bits communicated, possibly subject to privacy constraints.  Our setting can be viewed as a property-testing version of the distributed communication-efficient learning problem. 

More closely related to our halfspace algorithm is the work of \cite{braverman2019convex}, who study distributed halfspace learning and convex set disjointness over a known finite domain and measure communication in bits. Their goal is to minimize the number of bits exchanged. Our focus is on the smallest resolution $t$ for testing whether $\opt_\hypclass=0$ or $\opt_\hypclass\ge t$ under two-party DP. On the technical side, our halfspace algorithm builds on their container approach, using the improved container bounds of \cite{dutta2021uniform}. We apply this approach to sets of low-error classifiers in the dual space and select containers privately to eliminate candidates while preserving a shared, consistent classifier when one exists.

\paragraph{Testing and Estimating Learnability.} A related line of work asks whether one can test if the optimal population error using classifiers in a hypothesis class is small or large without learning a classifier \citep{balcan2012active,kothari2014testing,blum2018active,kong2018estimating}. In the active testing setting, the learner has access to a large pool of unlabeled samples and chooses which points to label. The goal is to answer this question using as few label queries as possible. A related question with access to i.i.d. labeled samples is studied in \citep{kong2018estimating}. Under distributional assumptions, they show that the optimal population loss can be estimated using fewer samples than are needed to learn an accurate predictor. For example, in isotropic regression with suitable moment bounds, they show a quadratic gap in the dependence on dimension: $O(\sqrt{d})$ samples suffice to estimate the optimal squared loss to constant accuracy, whereas learning can require $\Omega(d)$ samples. They also discuss using such estimates for data valuation and comparing separate and pooled datasets to assess whether joint training is promising.

These works share our goal of assessing prediction error without learning a classifier. They focus on using as few labels or samples as possible to assess population error. Our main constraint is privacy: two agents hold labeled datasets and want to determine whether a single classifier in their hypothesis class fits all the pooled records, or whether every classifier makes at least $t$ mistakes. The challenge is to achieve a small resolution $t$ while protecting individual records throughout the agents' interaction. We provide upper and lower bounds on $t$ that quantify the cost of two-party DP. 

Similar to this line of work, our threshold result also distinguishes between private testing and private learning. For fixed fractional empirical testing gap, learning accuracy, and confidence, our tester needs only $\widetilde{O}(\log\log m/\varepsilon)$ samples per party, whereas distribution-free pure-private threshold learning requires $\Omega(\log m/\varepsilon)$ samples even with a trusted curator.

\paragraph{Data Valuation and Selection.} Data valuation supports decisions throughout the machine learning pipeline, from selecting and pricing training data to debugging trained models~\citep{wang2024rethinking,tian2022private,koh2017understanding}. Popular approaches guide these decisions by assigning scores to individual records or entire datasets using Shapley values or leave-one-out contributions~\citep{ghorbani2019data,jia2019efficient,garridolucero2024du}. These methods must, however, account for non-monotonicity, since adding training data can worsen downstream test performance, including when pooling data from heterogeneous sources~\citep{nakkiran2020deep,compton2023more}. Our $\opt_\hypclass(S_A,S_B)>t$ scenario captures a related obstacle when individually realizable datasets admit no shared classifier that fits both perfectly. For data selection, such non-monotonicity can give adaptive methods a strict advantage over fixed score-based methods under equal value-query budgets~\citep{patel2026limits}, paralleling the benefit of interaction between agents in private threshold testing noted in \Cref{remark:no-interaction}.

\paragraph{Private Data Appraisal and Valuation.} Existing private appraisal methods differ in both what they protect and what their assessments establish. Secure-computation methods, for example, estimate learning gains or Shapley values while permitting leakage through the released appraisal~\citep{xu2022data,tian2022private}. Differential privacy can address such leakage, but its scope varies across methods. Protocols combining homomorphic encryption with label differential privacy leave features public~\citep{asghar2023practical}, while private KNN-based valuation assumes a trusted server and excludes validation-data privacy~\citep{wang2023privacy}. Similar asymmetries arise in dataset valuation that protects sellers without providing corresponding guarantees for the buyer~\citep{jahaninezhad2024private}. Protecting both sides therefore remains a challenge, as broader analyses of private valuation also emphasize~\citep{fu2026challenges}. Beyond these privacy constraints, the assessment itself may not certify compatibility. Similarity-based methods estimate dataset divergence or correlation without establishing whether a shared classifier fits the combined data~\citep{fuentes2025dataset,christensen2023semiprivate,jahaninezhad2024private}. Likewise, protecting fixed valuation scores does not itself ensure that they identify useful data combinations~\citep{patel2026limits}. These limitations leave unresolved the worst-case accuracy limits of compatibility testing under two-party record-level differential privacy without a trusted curator or computational assumptions.

\section{Preliminaries}

\subsection{Notations}

For a positive integer $r$, let $[r]:=\{1,\ldots,r\}$.  The public data domain is denoted by
$\dataspace\subseteq\Reals^d$; when it is finite, we write
$m:=|\dataspace|$. For $S,S'\in\dataspace^n$, we write $S\sim S'$ when they are
\emph{neighboring}, meaning that one record is replaced; the corresponding
replacement distance is denoted by $\mathrm d_{\text{ham}}(S,S')$.
For two vectors $u,v \in \Reals^d$, we write $\langle u,v\rangle$ for the Euclidean inner product,
$\conv(G)$ for the convex hull of $G\subset \Reals^d$.  Finally, $\Lap(b)$ is the mean-zero Laplace distribution with scale $b$.

\subsection{Two-Party DP and Connection to Information Complexity}

In an interactive protocol, privacy must hold not only for the final $\mathsf{Match}$ or $\mathsf{Far}$ output, but for the entire interaction transcript observed by each party.
For a protocol $\Pi$ and inputs $(S_A,S_B)$, let $\mathrm{View}_A^\Pi(S_A,S_B)$ denote Alice's complete view of the execution, consisting of her input $S_A$, her private randomness, the public randomness, all exchanged messages, and the common output; define $\mathrm{View}_B^\Pi(S_A,S_B)$ similarly.

\begin{definition}[Differential privacy for two-party protocols \citep{McGregorMironovPitassiReingoldTalwarVadhan10}] \label{def:two-party-dp}
A two-party protocol $\Pi$ is \emph{$\varepsilon$-differentially private}
if each party's view is $\varepsilon$-differentially private with respect
to the other party's dataset. More precisely, for every fixed $S_A$,
every pair of neighboring datasets $S_B,S_B'$, and every measurable
event $E$,
\[
\Pr \left[\mathrm{View}_A^\Pi(S_A,S_B)\in E\right]
\leq
e^\varepsilon
\Pr\left[\mathrm{View}_A^\Pi(S_A,S_B')\in E\right],
\]
and, symmetrically, for every fixed $S_B$, every pair of neighboring
datasets $S_A,S_A'$, and every measurable event $E$,
\[
\Pr\left[\mathrm{View}_B^\Pi(S_A,S_B)\in E\right]
\leq
e^\varepsilon
\Pr\left[\mathrm{View}_B^\Pi(S_A',S_B)\in E\right].
\]

We require differential privacy on the full input domain, including inputs that do not satisfy the promise of the testing problem.
\end{definition}

We also make a repeated use of this lemma which is a simple modification of the connection between DP and internal information complexity from \citep{McGregorMironovPitassiReingoldTalwarVadhan10}.

\begin{lemma}
\label{lem:metric-dp-to-information}
Let $\mathcal{A}$ and $\mathcal{B}$ be finite sets, and let $E_A:\mathcal{A}\to\dataspace^n$ and $E_B:\mathcal{B}\to\dataspace^n$ be deterministic maps that encode inputs in $\mathcal{A}$ and $\mathcal{B}$ as size-$n$ datasets over $\dataspace$. Define the \emph{encoding diameter} of $E_A$ and $E_B$ as follows: 
\[
\Delta_A
:=
\max_{a,a'\in\mathcal{A}}
\mathrm{d}_{\text{ham}}\bigl(E_A(a),E_A(a')\bigr),
\qquad
\Delta_B
:=
\max_{b,b'\in\mathcal{B}}
\mathrm{d}_{\text{ham}}\bigl(E_B(b),E_B(b')\bigr).
\]
Suppose $(A,B) \sim \mu$ for a distribution $\mu$ over $\mathcal{A}\times \mathcal{B}$, and Alice and Bob execute an $\varepsilon$-differentially private protocol $\Pi$ on the encoded datasets $\bigl(E_A(A), E_B(B)\bigr)$.
Let $T$ denote the resulting public transcript (consisting of the public randomness, all exchanged messages, and the final output). Then the internal information cost of $T$ (in bits) satisfies
\[
I(A;T\mid B)+I(B;T\mid A)
\le
\frac{\varepsilon(\Delta_A+\Delta_B)}{\ln 2}.
\]
\end{lemma}

\section{Threshold in  One Dimension}

In this section, we focus on the class of thresholds on real line. In particular, define 
\begin{equation} \label{eq:thr-def}
\hypclass_{\text{thr}}=\{h_\theta: \Reals \to \{\pm 1\}~|~ \exists \theta \in \Reals~\text{s.t.}~\text{$h_\theta(x)=1$~for $x\geq \theta$ and $h_\theta(x)=-1$ for $x<\theta$}\}.
\end{equation}
First notice that $\opt_{\hypclass_{\text{thr}}}(S_A,S_B)\leq n$, therefore, the non-vacuous regime for the resolution will be $1\leq t\leq n$. We first provide an information-theoretic lower bound on $t$ for this class, and then present an algorithm that (nearly) matches the lower bound.

\subsection{Lower bound for thresholds}

The next result presents a fundamental lower bound on $t$ for the class of thresholds. The main idea is to embed a binary function with a high information complexity and then use the connection between information complexity and two-view DP from \citep{McGregorMironovPitassiReingoldTalwarVadhan10} to provide a lower bound.

\begin{theorem}[Lower bound for thresholds] \label{thm:threshold-lower-bound}
There exist universal constants $c>0$ and $m_0\in\Naturals$ such that the following holds. For every $m\ge m_0$, every
$\dataspace\subseteq\Reals$ with $|\dataspace|=m$, every
$\varepsilon>0$, and every $n,t\in\Naturals$ such that $t \leq n$, we have that every 
$\varepsilon$-private tester for
$\hypclass_{\mathrm{thr}}$ with resolution $t$ for dataset of size $n$  (from \cref{def:private-tester}) requires
\[
t\ge c\frac{\log\log m}{\varepsilon}.
\]
\end{theorem}

\begin{remark}
Fix $\varepsilon\in(0,1]$ and $n\in\Naturals$.
\cref{thm:threshold-lower-bound} implies that no
$\varepsilon$-private tester for thresholds over $\Reals$ at
dataset size $n$ can achieve a resolution $1\le t\le n$.
Indeed, restricting such a tester to a finite domain of size $m$
would give $t\ge c\log\log(m)/\varepsilon$ for arbitrarily large
$m$, contradicting the fact that $t$ is fixed.

A trusted curator can achieve resolution $t=O(1/\varepsilon)$,
independently of the domain size, by adding Laplace noise to the
sensitivity-one quantity $\opt$. This resolution is nontrivial
provided $n\ge C/\varepsilon$ for a sufficiently large universal
constant $C$. Thus, on finite domains of size $m$, the two-party
setting incurs an unavoidable multiplicative
$\Omega(\log\log m)$ overhead relative to the centralized
resolution.
\end{remark}

\begin{remark}
Under cryptographic assumptions, secure function evaluation
can emulate the trusted curator and achieve
$t=O(1/\varepsilon)$ with computational differential
privacy \citep{McGregorMironovPitassiReingoldTalwarVadhan10}. However, \cref{thm:threshold-lower-bound} shows that information-theoretic view-based privacy
requires $t=\Omega\left(\log\log(m)/{\varepsilon}\right)$. Thus,  the two models exhibit a separation in the achievable promise gap.
\end{remark}
\begin{remark}\label{remark:no-interaction}
The lower bound in \cref{thm:threshold-lower-bound} allows arbitrary interaction between Alice and Bob. For one-way protocols with bounded worst-case error, however, the standard information-theoretic argument underlying the Greater-Than lower bound~\citep[Chapter~6]{rao2020communication} shows that Alice's message must reveal $\Omega(\log m)$ bits about her input when it is uniformly random. Indeed, given any prefix of Alice's input, Bob can choose a comparison that predicts her next bit with constant error; Fano's inequality and the chain rule then give the information lower bound. Applying this argument to the encoding in \cref{thm:threshold-lower-bound}, group privacy bounds the information revealed by Alice's message by $O(\epsilon t)$, yielding $t=\Omega(\log(m)/\epsilon)$. The private-interior-point baseline in \cref{sec:main-alg-thresh} achieves the matching $O(\log(m)/\epsilon)$ resolution. Thus, interaction is necessary to achieve our improved resolution of $\widetilde{O}(\log\log(m)/\epsilon)$.
\end{remark}

\begin{proof}[Proof of \cref{thm:threshold-lower-bound}]
Without loss of generality write $\dataspace=\{x_1,\cdots,x_m\}$ where $x_1\leq \cdots\leq x_m$. Fix an arbitrary integer $n\ge t$ and let $\ell=\lfloor\log_2(m-2)\rfloor$. For $i,j\in \{2,\cdots,2^{\ell}+1\}$, define the size-$n$ multisets
\[
S_A^{(i)}=(\underbrace{x_1,\dots,x_1}_{\text{$n-t$ times}},\underbrace{x_{i},\cdots,x_i}_{\text{$t$ times}}),
\qquad
S_B^{(j)}=(\underbrace{x_{j},\dots,x_{j}}_{\text{$t$ times}},\underbrace{x_m,\cdots,x_m}_{\text{$n-t$ times}}),
\]
where Alice's records are negatively labeled and Bob's records are positively labeled. Then, 
\begin{equation}
\label{eq:threshold-opt}
\opt_{\hypclass_{\mathrm{thr}}}(S_A^{(i)},S_B^{(j)})
=
\begin{cases}
0,&i<j,\\
t,&i\ge j.
\end{cases}
\end{equation}

There exists a distribution $\mu_\ell$ over $\in \{2,\cdots,2^{\ell}+1\}^2$ such that
every protocol computing $\mathrm{GT}_\ell(i,j)=\indic{i\ge j}$ with error at most $1/10$ has internal information cost
$\Omega(\log\ell)$ \citep{BravermanWeinstein12,viola2015communication}.  More precisely, let $\Pi$ be a private tester from \cref{def:private-tester}. Then, given $(I,J)\sim\mu_\ell$, run $\Pi$ on $(S_A^{(I)},S_B^{(J)})$. By
\cref{eq:threshold-opt}, the resulting protocol computes $\mathrm{GT}_\ell$ with error at most $1/10$. Therefore, if $T$ consists of the shared
randomness, exchanged messages, and common output, then $I(I;T\mid J)+I(J;T\mid I)=\Omega(\log\ell)$ by \citep{BravermanWeinstein12,viola2015communication}. The padding is independent of $i$ and $j$, and hence both encoding
diameters are at most $2t$. By
\cref{lem:metric-dp-to-information}, $I(I;T\mid J)+I(J;T\mid I)
\le
\frac{4\varepsilon t}{\ln 2}$. Then, it follows that
$
t = \Omega\left(\frac{\log\ell}{\varepsilon}\right) = \Omega\left(\frac{\log\log m}{\varepsilon}\right),
$
as claimed.
\end{proof}

\subsection{Algorithm for Thresholds}

In this part, we present our algorithm that achieves the lower bound up to $\log\log\log(m)$ factor.  

\subsubsection{Structural Properties and Binary Trie}
First, we introduce a notation: For a multiset $S$ of size $n$ over a totally ordered domain, write its elements, with multiplicity, as $s_{(1)}\le\cdots\le s_{(n)}$. For $k\in[n]$, define the multisets of
the $k$ smallest and $k$ largest elements by $S_k^{\downarrow}:=\{s_{(1)},\ldots,s_{(k)}\},$ and $S_k^{\uparrow}:=\{s_{(n-k+1)},\ldots,s_{(n)}\}.$  Now, we are ready to present a structural property.
\begin{lemma}
\label{lem:far-crossing}
Let $\dataspace=\{0,1\}^b$ where $m=2^b$. Define $\psi_A(x)=x\|1$ and $\psi_B(x)=x\|0$. In particular, $\psi_A$ and $\psi_B$  add a bit $1$ and $0$  after the least significant bit of the input, respectively.  Let $S_A,S_B$ be multisets of size $n$ over $\dataspace$, let
$k\in[n]$, and write $\widetilde S_A=\psi_A(S_A)$ and  $\widetilde S_B=\psi_B(S_B)$. Then
$
\opt_{\hypclass_{\mathrm{thr}}}
    (\widetilde S_A,\widetilde S_B)
=
\opt_{\hypclass_{\mathrm{thr}}}(S_A,S_B).
$
Moreover:
\begin{enumerate}
    \item if
    $\opt_{\hypclass_{\mathrm{thr}}}(S_A,S_B)=0$, then
    $
    \max\widetilde S_{A,k}^{\uparrow}
    <
    \min\widetilde S_{B,k}^{\downarrow};
    $

    \item if
    $\opt_{\hypclass_{\mathrm{thr}}}(S_A,S_B)\ge 2k-1$, then
    $
    \min\widetilde S_{A,k}^{\uparrow}
    >
    \max\widetilde S_{B,k}^{\downarrow}.
    $
\end{enumerate}
\end{lemma}

\begin{proof}
Reading strings as integers, $\psi_A(x)=2x+1$ and $\psi_B(x)=2x$. Thus both maps are strictly increasing, and their images are disjoint
by parity.

At an even threshold $\widetilde\theta=2s$, the
Alice and Bob errors are, respectively,
\[
\{x\in S_A:x\ge s\},
\qquad
\{x\in S_B:x<s\},
\]
which are exactly the errors at threshold $s$ in the original
instance. At an odd threshold $\widetilde\theta=2s+1$, the error
multisets are
\[
\{x\in S_A:x\ge s\},
\qquad
\{x\in S_B:x\le s\}.
\]
Thus, compared with $2s$, the odd threshold $2s+1$ makes exactly the same errors except that it may additionally misclassify Bob’s records at s. It can therefore never be better than $2s$. It follows that the two optimal errors are equal.

If the original instance is realizable, then $\max S_A<\min S_B$. So every element of $\widetilde S_A$ lies below every element of
$\widetilde S_B$, proving part~(i). For part~(ii), suppose instead that
\[
a:=\min\widetilde S_{A,k}^{\uparrow}
\le
b:=\max\widetilde S_{B,k}^{\downarrow}.
\]
Since $a$ is odd and $b$ is even, we have $a<b$. At threshold $b$,
every Alice error is strictly larger than $a$, so there are at most
$k-1$ Alice errors. Likewise, every Bob error is strictly smaller than
$b$, so there are at most $k-1$ Bob errors. Therefore,
\[
\opt_{\hypclass_{\mathrm{thr}}}
    (\widetilde S_A,\widetilde S_B)
\le 2k-2,
\]
contradicting the assumption. This proves part~(ii).
\end{proof}

The lemma reduces the promise problem to a comparison between two
extremal blocks. Fix $k=\lfloor t/2\rfloor$. In the Match case, every
point in $\widetilde{S}_{A,k}^{\uparrow}$ lies below every point in
$\widetilde{S}_{B,k}^{\downarrow}$. In the Far case, their order is reversed: every
point in $\widetilde{S}_{A,k}^{\uparrow}$ lies above every point in
$\widetilde{S}_{B,k}^{\downarrow}$. In particular, being far from realizable creates
not merely one misordered pair, but two blocks of $k$ records with
opposite order. This introduces a stability that we later exploit:
the promised cases can be distinguished using robust order statistics
rather than the maximum of $S_A$ and the minimum of $S_B$, which may each depend on a single record.

\begin{remark} \label{rem:lifting-far}
To avoid ties between records owned by different parties, we apply the lifting function $\psi_A$ and $\psi_B$ from \cref{lem:far-crossing} and suppress the tildes. Since this transformation is local, it entails no
loss of generality. Henceforth, we assume that the supports of $S_{A,k}^{\uparrow}$ and
$S_{B,k}^{\downarrow}$ are disjoint, so the preceding order comparisons are strict. Without loss of generality to reduce notation clutter, we assume Alice and Bob only use $S_{A,k}^{\uparrow}$ and $S_{B,k}^{\downarrow}$ during the execution of the algorithm. It is easy to see that these subsets also respect the neighboring condition we require for the definition of DP.
\end{remark}

\begin{definition}[Binary Trie and Common Node]
\label{def:tree}
Let $\mathcal{X}=\{0,1\}^{b}$ with $2^b=m$. Let $\mathcal{T}$ be the complete binary trie whose nodes are the binary strings
$v\in\{0,1\}^{\le b}=\bigcup_{r=0}^{b}\{0,1\}^{r}$. The \emph{depth} of a node is its length $|v|$:
the root is the empty string at depth $0$, and the leaves are the depth-$b$ strings,
which we identify with $\mathcal{X}$.  For each $\ell \in [b]$ and  $p \in \{0,1\}^\ell$ define $A_p^{\ell}=\{x_i \in S_A: \mathrm{prefix}_\ell(x_i)=p\}$ and similarly  define $B_p^{\ell}=\{x_i \in S_B: \mathrm{prefix}_\ell(x_i)=p\}$. Also, we have the convention that $A_p^{0}=S_A$ and $B_p^{0}=S_B$. At depth $\ell$ a node is called \emph{common} if it contains both Alice and Bob points.
\end{definition}

Next lemma shows an important property of the binary trie node, that we use in the sequel to design an algorithm.

\begin{lemma}\label{lem:unique-common-node}
Fix $\dataspace =\{0,1\}^b$ and $n \in \Naturals$. Let $S_A,S_B \subseteq \dataspace^n$ be two sequence of length $n$ and assume the promise holds. Recall the definition of the trie node from \cref{def:tree}. Then, for every given depth $\ell \in [b]$, there exists at most on node that contains both Alice and bob's data, i.e., $|\{p \in \{0,1\}^l : B^\ell_p \cdot A^\ell_p\neq 0\}|\leq 1$. Also,  at depth $b$, we have $|\{p \in \{0,1\}^b : B^\ell_p \cdot A^\ell_p\neq 0\}|= 0 $.
\end{lemma}
\begin{proof}
Under the promise from \cref{lem:far-crossing} and  \cref{rem:lifting-far}, either $\max S_{A} < min S_{B}$ or $\max S_{B,k} < min S_{A}$. First, we prove the last claim: by \cref{lem:far-crossing} (recall also \cref{rem:lifting-far}), we have that under the promise, the block of Alice and Bob's data are disjoint. 
For the first claim, we prove by contradiction. Lets assume  at depth $\ell$, there exists two nodes that are both common and denoted by $p_1 \in \{0,1\}^\ell$ and $p_2 \in \{0,1\}^\ell$ such that $A_{p_1}^\ell . B_{p_1}^\ell\neq 0$ and $A_{p_2}^\ell . B_{p_2}^\ell\neq 0$. Assume without loss of
generality that $p_1<p_2$ in lexicographic order. Every leaf below $p_1$ is smaller than every leaf below $p_2$. Choose
$a_{p_1}\in A_{p_1}^\ell$ $b_{p_1}\in B_{p_1}^\ell$ $a_{p_2}\in A_{p_2}^\ell$ $b_{p_2}\in B_{p_2}^\ell$. It follows that
$ a_{p_1}<b_{p_2}$ and $b_{p_1}<a_{p_2}$.
If $\max S_A<\min S_B$, then $a_q<b_p$, contradicting $b_p<a_q$.
If $\min S_A>\max S_B$, then $a_p>b_q$, contradicting $a_p<b_q$.
\end{proof}

\begin{example}
To gain some intuition, behind how the binary trie structure, consider \cref{fig:tree} where it shows $\dataspace=\{0,1\}^3$. Here, $\opt\left(S_A,S_B\right)=0$. Note that at depth zero, the single node is always common and the last depth doesn't have any common nodes.
\end{example}

\begin{lemma}[escaped points]
Fix $\dataspace =\{0,1\}^b$ and $n \in \Naturals$. Let $S_A,S_B \subseteq \dataspace^n$ be two sequence of length $n$ and assume the promise holds. Define the  function $F: [b]\to \Naturals$ as $F(\ell) = n - \max_{p \in \{0,1\}^\ell}\min\{A_p^\ell,B_p^\ell\}$. Then $F_\ell$ is a non-decreasing function with $F(0) = 0$ and $F(b)=n$.
\end{lemma}
\begin{proof}
Set $M(\ell):=
\max_{p\in\{0,1\}^{\ell}}
\min\bigl\{|A_p^\ell|,|B_p^\ell|\bigr\}$, so that $F(\ell)=n-M(\ell)$. If $q$ is a node at depth $\ell+1$ and
$p$ is its parent, then $A_q^{\ell+1}\subseteq A_p^\ell$ and
$B_q^{\ell+1}\subseteq B_p^\ell$. Hence
\[
\min\bigl\{|A_q^{\ell+1}|,|B_q^{\ell+1}|\bigr\}
\le
\min\bigl\{|A_p^\ell|,|B_p^\ell|\bigr\}
\le M(\ell).
\]
Maximizing over $q$ gives $M(\ell+1)\le M(\ell)$, and therefore
$F(\ell+1)\ge F(\ell)$. At the root, $|A_{\emptyset}^0|=|B_{\emptyset}^0|=n$, so $F(0)=0$.
At depth $b$, no leaf is common by
\cref{lem:unique-common-node}; hence $M(b)=0$ and $F(b)=n$.
\end{proof}

\subsubsection{Distributed Private Common-Frequency Estimation}

In this section, we describe an algorithm for a problem that we refer to as \emph{distributed common heavy hitter}. This subroutine later used in our algorithm for the thresholds. 

The task is simple: Alice and Bob hold a categorical dataset of size $n$ over a finite universe of size $m$. Their goal is to estimate the maximum number of points in a shared bucket in the distributed two-party DP case. We propose an algorithm for this task in \cref{alg:distributed-min}.

\begin{algorithm}
\caption{Private Common-Frequency Estimation: $\mathcal{A}_\mathsf{PCF}$}
\label{alg:distributed-min}
\textbf{Input:} Alice holds $S_A=(X_1,\dots,X_n)\in[m]^n$, Bob holds $S_A=(Y_1,\dots,Y_n)\in[m]^n$; parameters $\epsilon,\beta$.\\
\textbf{Output:} estimate $\hat f$ of $f=\max_{k\in[m]}\min\{A_k,B_k\}$.
\begin{algorithmic}[1]
  \State $R \gets \lceil 4n^2/\beta\rceil$.
  \If{$m \le R$}
     \State $g \gets \mathrm{id}_{[m]}$,\quad $R \gets m$. \Comment{\textcolor{black}{domain already small: no hashing ($\mathrm{id}_{[m]}$ is the identity map)}}
  \Else
     \State draw $g:[m]\to[R]$ uniformly at random from public coins. \Comment{\textcolor{black}{public hash}}
  \EndIf
  \State Alice releases $\tilde A_z \gets \big|\{i:g(X_i)=z\}\big| + \Lap(2/\epsilon)$ for all $z\in[R]$.
  \State Bob releases $\tilde B_z \gets \big|\{i:g(Y_i)=z\}\big| + \Lap(2/\epsilon)$ for all $z\in[R]$.
  \State \Return $\hat f \gets \max_{z\in[R]}\min\{\tilde A_z,\tilde B_z\}$.
\end{algorithmic}
\end{algorithm}
\begin{theorem}[Private Common-Frequency Estimation]
\label{thm:distributed-common-heavy-hitter}
For every $n,m\in\Naturals$, $\varepsilon>0$, and $\beta\in(0,1)$,
\cref{alg:distributed-min} is an $\varepsilon$-two-party
differentially private protocol. Moreover, for every pair of inputs
$S_A,S_B\in[m]^n$, with probability at least $1-\beta$,
$\left|\widehat f-f(S_A,S_B)\right|
\le
\frac{2}{\varepsilon}\ln\left(\frac{4r}{\beta}\right)$ where 
$
r:=\min\left\{m,\left\lceil\frac{4n^2}{\beta}\right\rceil\right\}.
$
In particular,
\[
\Pr\left(\left|\widehat f-f(S_A,S_B)\right|\leq\frac{2\log(16n^2/\beta)}{\varepsilon}\right)\geq 1-\beta.
\]
\end{theorem}
\begin{proof}
For $z\in[r]$, let
\[
A_z^g:=\bigl|\{i\in[n]:g(X_i)=z\}\bigr|,
\qquad
B_z^g:=\bigl|\{i\in[n]:g(Y_i)=z\}\bigr|.
\]

We first prove privacy. For every fixed public hash function $g$, changing
one record changes the corresponding hashed histogram in $\ell_1$-distance
by at most $2$. Thus, by \cref{lem:laplace}, Alice's release
$(\widetilde A_z)_{z\in[r]}$ is $\varepsilon$-DP in $S_A$, and Bob's
release $(\widetilde B_z)_{z\in[r]}$ is $\varepsilon$-DP in $S_B$.
All subsequent computations are post-processing. Therefore, Bob's view is
$\varepsilon$-DP in $S_A$ and Alice's view is $\varepsilon$-DP in $S_B$,
as required.

It remains to prove accuracy. Let
$\mathcal U:=\supp(S_A)\cup\supp(S_B)$, and we have $|\mathcal U|\le 2n$. If $r=m$, then $g$ is the identity. Otherwise, by a union bound,
\[
\Pr\!\left[g\text{ is not injective on }\mathcal U\right]
&\le
\binom{2n}{2}\frac1r \\ 
&\le
\frac{\beta}{2}.
\]
On the complementary event, every occupied hash bucket contains records
from a single element of $[m]$. Consequently,
$
\max_{z\in[r]}\min\{A_z^g,B_z^g\}=\max_{j\in[m]}\min\{A_j,B_j\}=f(S_A,S_B).
$
Set
$
\tau:=\frac{2}{\varepsilon}\ln\left(\frac{4r}{\beta}\right).
$
There are $2r$ independent Laplace variables, each with scale
$2/\varepsilon$. Hence, by \cref{lem:laplace}
\[
\Pr\!\left[
\max_{z\in[r]}
\max\left\{
|\widetilde A_z-A_z^g|,
|\widetilde B_z-B_z^g|
\right\}>\tau
\right]
\le
2r e^{-\varepsilon\tau/2}
=
\frac{\beta}{2}.
\]
On the complementary event, for every \(z\in[r]\), $\left|
\min\{\widetilde A_z,\widetilde B_z\}
-
\min\{A_z^g,B_z^g\}
\right|
\le \tau.$ Taking the maximum over \(z\) therefore gives
$
\left|
\widehat f-
\max_{z\in[r]}\min\{A_z^g,B_z^g\}
\right|
\le \tau.
$
Combining the hashing and noise events proves the stated guarantee.
\end{proof}
\subsubsection{Main Algorithm}\label{sec:main-alg-thresh}
In this section building up on various lemma we proved in the previous two sections, we state the main algorithm and its guarantees. Before, doing it, we discuss a simple baseline that shows why simple idea fails

\paragraph{Simple baseline: private interior point.}  A natural approach is for Alice to privately select a point lying within the range of her witness block and send it to Bob. Bob can then determine
whether his witness block lies above or below this point. We show that this idea already solves the promise problem, but at a $\log(m)$ cost that our main algorithm improves to $\log\log(m)$.
For $\theta\in\dataspace$, define
$
\text{depth}_{S_A}(\theta)
:=
\min\!\left\{
|\{x\in S_A:x\leq\theta\}|,
|\{x\in S_A:x\geq\theta\}|
\right\}.
$
This score has sensitivity one. Moreover,
$
\max_{\theta\in\dataspace}\text{depth}_{S_A}(\theta)
\geq \left\lceil\frac{k}{2}\right\rceil,
$
since every median of $S_A$ has depth at least $\lceil k/2\rceil$. Alice applies the exponential mechanism with privacy $\varepsilon$ and
score $\text{depth}_{S_A}$, and sends the resulting point $\hat\theta$ to Bob.
The exponential-mechanism guarantee implies that, if
$
k\geq C\frac{\log(2m/\beta)}{\varepsilon}
$, with probability at
least $1-\beta/2$, for a sufficiently large universal constant $C$, then
$\text{depth}_{S_A}(\hat\theta)>0$ and hence
$\hat\theta\in[\min S_A,\max S_A]$. Then, Bob releases $\widetilde c_B
:=
|\{x\in S_B:x<\hat\theta\}|
+\Lap(1/\varepsilon)$ and outputs $\mathsf{Match}$ if $\widetilde c_B<k/2$ and
$\mathsf{Far}$ otherwise.  It follows that the baseline solves the $t$-promise problem for
$
t\geq C\frac{\log(2m/\beta)}{\varepsilon}.
$
The dependence on $m$ comes entirely from privately selecting a point
from an $m$-element domain. As we will see, our main algorithm improves exponentially over this baseline by avoiding this direct
selection by searching over a different object.

\begin{algorithm}
\caption{Data Valuation for Threshold: $\mathcal{A}_\mathsf{val-thr}$}
\label{alg:thresh-main}
\textbf{Input:} Alice holds $S_A=(X_1,\dots,X_n)\in \left(\{0,1\}^b\right)^n$, Bob holds $S_A=(Y_1,\dots,Y_n)\in \left(\{0,1\}^b\right)^n$; parameters $\epsilon,\beta$.\\
\textbf{Output:} 
\begin{algorithmic}[1]
\State $T\gets\lceil\log_2 b\rceil$, $\epsilon_0 \gets \epsilon/2T$, $\beta_0 \gets \beta/2T$
\State $\displaystyle
\alpha\gets
20\max\left\{
\frac{2}{\epsilon_0}
\ln\left(
\frac{
4\left\lceil 4n^2/\beta_0\right\rceil
}{\beta_0}
\right),
\;
\frac{28}{\epsilon}\ln\frac{4}{\beta}
\right\}.
$
\State Set $L\gets0$, $H\gets b$,
$\widehat F(0)\gets0$, and $\widehat F(b)\gets n$. \Comment{\textcolor{black}{Initialization of binary search}}
  \While{$H-L>1$} \Comment{\textcolor{black}{Binary search step}}
    \State $\ell\gets\lfloor(L+H)/2\rfloor$.
    \State Form the prefix datasets
    \[
    S_A^\ell
    \gets
\bigl(\prefix_\ell(X_1),\ldots,\prefix_\ell(X_n)\bigr),
    \qquad
    S_B^\ell
    \gets
\bigl(\prefix_\ell(Y_1),\ldots,\prefix_\ell(Y_n)\bigr).
    \]
    \State $\widehat F(\ell) \gets n -  \Alg_\PCF(S_A^\ell,S_B^\ell;\varepsilon_0,\beta_0)$. \Comment{\textcolor{black}{See \cref{alg:distributed-min}}}
    \If{$\widehat F(\ell)<\alpha$}
        \State $L\gets\ell$.
    \Else
        \State $H\gets\ell$.
    \EndIf
\EndWhile

\State $\widehat\Delta\gets\widehat F(H)-\widehat F(L)$.

\If{$\widehat\Delta\ge\alpha/2$}
    \Comment{\textcolor{black}{large-jump case}}
    \State Alice releases
    \[
    \begin{aligned}
    \widehat V_A
    \gets {}&
    \max_{p\in\{0,1\}^{L}}
    \left\{
        6|A_p^{L}|+|A_{p0}^{H}|
    \right\}-
    \max_{p\in\{0,1\}^{L}}
    \left\{
        6|A_p^{L}|+|A_{p1}^{H}|
    \right\}
    +\Lap(28/\epsilon).
    \end{aligned}
    \]
    \State Bob releases
    \[
    \begin{aligned}
    \widehat V_B
    \gets {}&
    \max_{p\in\{0,1\}^{L}}
    \left\{
        6|B_p^{L}|+|B_{p0}^{H}|
    \right\}-
    \max_{p\in\{0,1\}^{L}}
    \left\{
        6|B_p^{L}|+|B_{p1}^{H}|
    \right\}
    +\Lap(28/\epsilon).
    \end{aligned}
    \]
\Else
    \Comment{\textcolor{black}{cumulative-escape case}}
    \State Alice releases
    \[
    \begin{aligned}
    \widehat V_A
    \gets {}&
    \max_{p\in\{0,1\}^{H}}
    \bigg\{
        6|A_p^{H}|
        +\sum_{\substack{q\in\{0,1\}^{H}\\q<p}}
        |A_q^{H}|
    \bigg\}-
    \max_{p\in\{0,1\}^{H}}
    \bigg\{
        6|A_p^{H}|
        +\sum_{\substack{q\in\{0,1\}^{H}\\q>p}}
        |A_q^{H}|
    \bigg\}
    +\Lap(28/\epsilon).
    \end{aligned}
    \]
    \State Bob releases
    \[
    \begin{aligned}
    \widehat V_B
    \gets {}&
    \max_{p\in\{0,1\}^{H}}
    \bigg\{
        6|B_p^{H}|
        +\sum_{\substack{q\in\{0,1\}^{H}\\q<p}}
        |B_q^{H}|
    \bigg\}-
    \max_{p\in\{0,1\}^{H}}
    \bigg\{
        6|B_p^{H}|
        +\sum_{\substack{q\in\{0,1\}^{H}\\q>p}}
        |B_q^{H}|
    \bigg\}
    +\Lap(28/\epsilon).
    \end{aligned}
    \]
\EndIf
 \State \Return $\mathsf{Match}$ if $\widehat V_A\ge\widehat V_B$; otherwise \Return $\mathsf{Far}$.
\end{algorithmic}
\end{algorithm}

\paragraph{Overview of the Technical Ideas of Our Algorithm.} In this section, we provide an overview of the ideas in \cref{alg:thresh-main}. We organize the ordered domain as a binary trie as defined in \cref{def:tree}. The first step is two perform private private binary search in order to find
\[
\ell^\star = \min\{\ell \in [b]: F(\ell)\geq \alpha\},
\]
where $F: [b]\to \Naturals$ defined as $F(\ell)=n - \max_{p \in \{0,1\}^\ell} \min\{A_p^\ell,B_p^\ell\}$. Notice that for every depth $\ell$,  $\{A_p^\ell\}_{p \in \{0,1\}^\ell}$ and $\{B_p^\ell\}_{p \in \{0,1\}^\ell}$ can be seen as a histogram. The straightforward implementation of DP histogram estimation incurs an error of $O(\ell/\epsilon_0)$ which can be $O(\log(m)/\epsilon_0)$. The main observation here is that we only need to estimate $\max_{p \in \{0,1\}^\ell} \min\{A_p^\ell,B_p^\ell\}$ without estimating the whole histogram. To do so, we develop \cref{alg:distributed-min} where the main idea is to use public random hashing to design a distributed algorithm with two-view DP guarantee such that  the error due to DP to estimate $\max_{p \in \{0,1\}^\ell} \min\{A_p^\ell,B_p^\ell\}$ is independent of the domain size.

\begin{figure}[htbp]
    \centering
    \begin{subfigure}{0.49\textwidth}
        \centering
        \includegraphics[width=\textwidth]{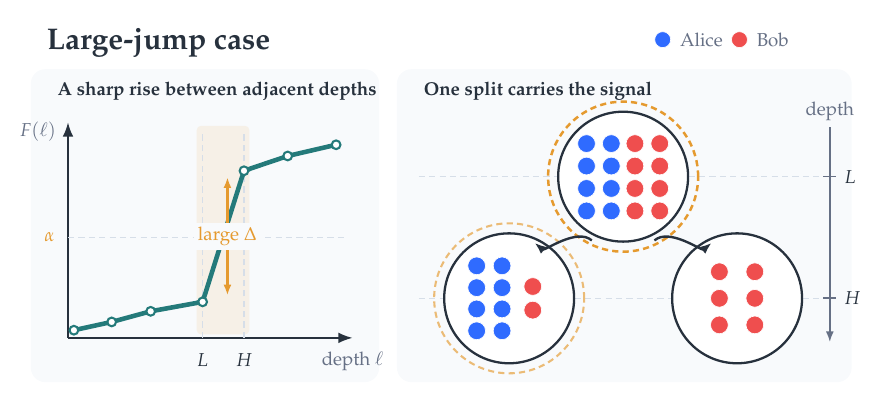}
        \caption{Large jump case}
        \label{fig:sub1}
    \end{subfigure}
    \hfill %
    \begin{subfigure}{0.49\textwidth}
        \centering
        \includegraphics[width=\textwidth]{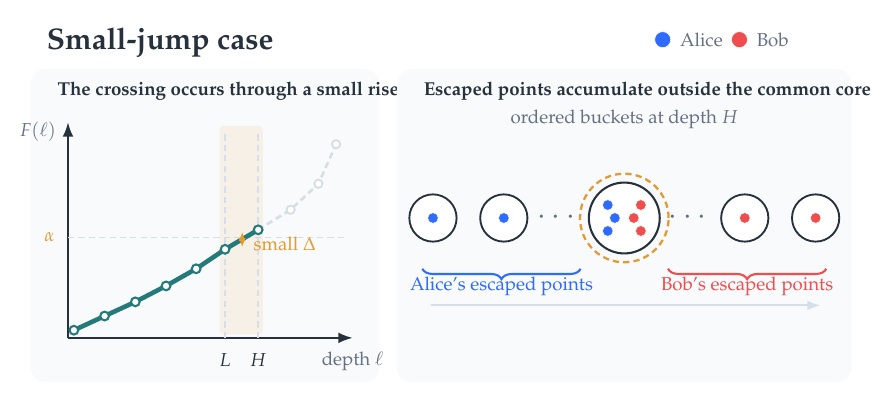}
        \caption{Small jump case}
        \label{fig:sub2}
    \end{subfigure}
    
    \caption{Intuition behind the two-cases of the second stage of \cref{alg:thresh-main}}
    \label{fig:thr-both}
\end{figure}

Once the binary search returns two consecutive depths $L$ and $H$, we consider whether the increase from $F(L)$ to $F(H)$ is large or small. For intuition, suppose first that the common nodes at these depths are known. If the increase is large, then many points leave the common node between depths $L$ and $H$. Since $H=L+1$, this information is concentrated in a single split, and comparing the numbers of Alice and Bob points in its two children reveals their relative order. To gain intuition behind this case, see \cref{fig:sub1}. Our algorithm is based on privately finding the \emph{orientation} of split on Alice and Bob's side and comparing them. 

If the increase is small, the information is not concentrated in the final split. Instead, because $F(H)$ is already above the noise level, sufficiently many points have accumulated outside the common node by depth $H$, while the common node itself remains heavy. Under the Match promise, Alice's points outside this node lie to its left and Bob's points lie to its right; under the Far promise, these directions are reversed. To gain intuition see \cref{fig:sub2}. Thus, comparing the total mass on the two sides reveals the correct answer. The actual implementation is more challenging because the common node depends on the private datasets and cannot be released explicitly; our statistics are designed to recover these two comparisons without identifying it.

\begin{theorem}[Upper bound for thresholds]
\label{thm:threshold-upper-bound}
There exists a universal constant $C>0$ such that the following holds.
Let $\dataspace\subseteq\Reals$ be a finite ordered domain with
$|\dataspace|=m\geq4$, let $n\in\Naturals$, let
$\epsilon\in(0,1]$ and $\beta\in(0,1/3]$, and let
$t\in[n]$ satisfy
\[
t
\geq
C\frac{\log\log m}{\epsilon}\left(\log\left(\frac{1}{\beta}\right) + \log\left(\frac{\log\log m}{\epsilon}\right)\right)
\]
Also, assume $n \geq C \frac{\log\log m}{\epsilon}
\log\left(\frac{\log\log m}{\beta\epsilon}\right)$. Then $\mathcal{A}_{\mathsf{val\text{-}thr}}$ (in \cref{alg:thresh-main}) is an
$\epsilon$-private tester for $\hypclass_{\mathrm{thr}}$ with resolution $t$ with failure probability at most $\beta$. 
\end{theorem}
\begin{proof}
The proof has three parts: \emph{private binary search}, and then, after that we have two cases, \emph{large-jump case} and \emph{cumulative-escape case}.

\paragraph{Step 1: private binary search}. The goal of this step is to approximate
\[
\ell^\star = \min\{\ell \in [b]: F(\ell)\geq \alpha\}.
\]
Define the following bad event
\begin{equation} \label{eq:bad-event-bs}
\mathcal{B}_{bs}=\{\exists t\leq T: |F(\ell_t)-\hat{F}(\ell_t)|\geq \alpha/20\},
\end{equation}
where $\{\ell_t\}_{t\in [T]}$ are itself random variable denoting the indices chosen by the private binary search step. Also, $T = O\left(\log(b)\right)$ and determinstic. Then, we write
\[
\Pr\left(\mathcal{B}_{bs}\right) &\leq \sum_{t=}^{T} \Pr\left(|F(\ell_t)-\hat{F}(\ell_t)|\geq \alpha/20\right)\\
&\leq \sum_{t=}^{T} \EE_{\ell_t}\left[\Pr\left(|F(\ell_t)-\hat{F}(\ell_t)|\geq \alpha/20 \big|  \ell_t\right)\right]\\
&\leq T.\beta_0\\
&\leq \beta.
\]
where the first step is due to the union bound, and the second step is by the law of the total probability, then, in the third step, we use the fact that each execution of the $\Alg_{\PCF}$ uses a fresh randomness independent of the history of the algorithm to obtain the result.

Notice that the private binary search maintains $L,H$ such that $H-L=1$ and $\hat{F}(L)\leq \alpha\leq \hat{F}(H)$. Under the event of $\mathcal{B}_{\text{bs}}^{c}$, we have
\begin{equation} \label{eq:ub-lb-fh-fl}
\begin{aligned}
F(L)&\leq \hat{F}(L) + \frac{\alpha}{20}\leq \frac{21 \alpha}{20},\\
F(H)&\geq \hat{F}(H) - \frac{\alpha}{20} \geq \frac{19 \alpha}{20}.
\end{aligned}
\end{equation}
\paragraph{Step~2-Case~1: Large Jump.} 
Suppose that $\widehat{\Delta}\geq\alpha/2$. Throughout the proof, we assume we are under $\mathcal{B}_{bs}^c$ in \cref{eq:bad-event-bs}. First, we have $\Delta \geq \widehat{\Delta}-\frac{\alpha}{10} \geq \frac{2\alpha}{5}$. Since $n\geq4\alpha$, we have $F(L)\leq\frac{21\alpha}{20}\leq\frac{21n}{80}<n.$ Therefore, $n-F(L)>0$, so there is a node $p\in\{0,1\}^L$ satisfying $\min\left\{|A_p^L|,|B_p^L|\right\}=n-F(L)$. By \cref{lem:unique-common-node}, this is the unique common node at
depth $L$.   

We next show that the maximization used in this case are attained at the common node. We prove correctness when
$\opt_{\hypclass_{\mathrm{thr}}}(S_A,S_B)=0$; the Far case is
symmetric. To show that the argmax is achieved at the common node consider an arbitrary $q\neq p$ where $q \in \{0,1\}^L$. Then, for every $j \in \{0,1\}$
\[
6 |A_{p}| + |A_{pj}| - (6 |A_{q}| + |A_{qj}|) &= 6\left(|A_{p}|-|A_{q}|\right) + |A_{pj}|-|A_{qj}| \\
&\geq 6\left(n-F(L)-F(L)\right) - F(L)\\
&\geq 6n - 13 F(L)\\
&\geq 6n - 13 \frac{21\alpha}{20},
\]
where the second step is by $|A_p|=n-F(L)$, $|A_q|\leq F(L)$, $|A_{pj}| \geq 0$, and $|A_{qj}| \leq |A_q|\leq F(L)$. The last step follows from \cref{eq:ub-lb-fh-fl}. It shows that assuming $n \geq 4 \alpha$ the last term is positive and hence $p$ is argmax. Similarly it holds for Bob's maximization part. Therefore, 
\[
 \max_{p\in\{0,1\}^{L}}
    \left\{6|A_p^{L}|+|A_{p0}^{H}| \right\}-\max_{p\in\{0,1\}^{L}}\left\{6|A_p^{L}|+|A_{p1}^{H}|\right\}=|A_{p0}^{H}| - |A_{p1}^{H}|,
\]
and similarly 
\[
 \max_{p\in\{0,1\}^{L}}
    \left\{6|B_p^{L}|+|B_{p0}^{H}| \right\}-\max_{p\in\{0,1\}^{L}}\left\{6|B_p^{L}|+|B_{p1}^{H}|\right\}=|B_{p0}^{H}| - |B_{p1}^{H}|.
\]
Under the Match promise, every Alice point is smaller than every Bob point. Therefore, Alice cannot have a point in $p1$ while Bob has a point in $p0$.  Equivalently, $|A_{p1}^{H}|\cdot |B_{p0}^{H}|=0$. Suppose first that $|A_{p1}^{H}|=0$; the case
$|B_{p0}^{H}|=0$ follows similarly. Since $\Delta>0$, we must have
$|A_{p0}^{H}|>|B_{p0}^{H}|$. Indeed, otherwise the common frequency
would remain $|A_{p0}^{H}|$ when passing from depth $L$ to depth $H$,
contradicting $\Delta>0$. Therefore, $n-F(L)
=
\min\left\{
|A_{p0}^{H}|,
|B_{p0}^{H}|+|B_{p1}^{H}|
\right\}$ whereas $n-F(H)=|B_{p0}^{H}|$. It follows that
\begin{align*}
\Delta
&=F(H)-F(L)\\
&=
\min\left\{
|A_{p0}^{H}|-|B_{p0}^{H}|,
|B_{p1}^{H}|
\right\}.
\end{align*}

Consequently, the difference between Alice's and Bob's unnoised
statistics is
\begin{align*}
&\left(|A_{p0}^{H}|-|A_{p1}^{H}|\right)
-
\left(|B_{p0}^{H}|-|B_{p1}^{H}|\right)\\
&\qquad=
|A_{p0}^{H}|-|B_{p0}^{H}|+|B_{p1}^{H}|\\
&\qquad\geq
2\min\left\{
|A_{p0}^{H}|-|B_{p0}^{H}|,
|B_{p1}^{H}|
\right\}\\
&\qquad=
2\Delta
\geq
\frac{4\alpha}{5}.
\end{align*}
The case $|B_{p0}^{H}|=0$ follows symmetrically and gives the same
lower bound. Thus, under the Match promise, the unnoised statistic has
positive margin at least $4\alpha/5$. The final noise is smaller than
this margin under the corresponding good event, and hence the
algorithm outputs $\mathsf{Match}$.

Under the Far promise, the order is reversed, and the same argument
gives a negative margin of magnitude at least $4\alpha/5$. Therefore, the algorithm outputs $\mathsf{Far}$.

For the privacy proof, each maximization has sensitivity at most $7$, since changing one record changes the term multiplied by $6$ by at most $6$ and the remaining count by at most $1$. Therefore, the difference of the two maximizations has sensitivity at most $14$, and allocating privacy budget $\epsilon/2$ to its release requires noise $\Lap(14/(\epsilon/2))=\Lap(28/\epsilon)$.

\paragraph{Step~2-Case~2: Cumulative Escape.}
Suppose that $\widehat{\Delta}<\alpha/2$. Under
$\mathcal{B}_{\mathrm{bs}}^c$, we have
$\Delta\leq\widehat{\Delta}+\alpha/10<3\alpha/5$. Therefore, by
\cref{eq:ub-lb-fh-fl},
\[
F(H)=F(L)+\Delta
<
\frac{21\alpha}{20}+\frac{3\alpha}{5}
=
\frac{33\alpha}{20}.
\]
Since $n\geq4\alpha$, we have
$F(H)<33n/80<n$. Therefore, there is a node
$p\in\{0,1\}^H$ satisfying
$\min\{|A_p^H|,|B_p^H|\}=n-F(H)$. By
\cref{lem:unique-common-node}, this is the unique common node at depth
$H$. We next show that the maximizations used in this case are attained at
$p$. As before, we prove correctness under the Match promise; the Far
case follows symmetrically. Consider an arbitrary
$q\in\{0,1\}^H$ with $q\neq p$. Then
\begin{align*}
&6|A_p^H|+|A_{<p}^H|
-
\left(6|A_q^H|+|A_{<q}^H|\right)\\
&\qquad\geq
6\left(n-F(H)-F(H)\right)-n\\
&\qquad=
5n-12F(H)\\
&\qquad>
5n-12\frac{33n}{80}
=
\frac{n}{20}>0.
\end{align*}
Here we used $|A_p^H|\geq n-F(H)$, $|A_q^H|\leq F(H)$,
$|A_{<p}^H|\geq0$, and $|A_{<q}^H|\leq n$. The same argument, with
$A_{>p}^H$ and $A_{>q}^H$ in place of $A_{<p}^H$ and $A_{<q}^H$,
shows that the second maximization is also attained at $p$. The same
conclusion holds for Bob. Therefore,
\[
\max_{q\in\{0,1\}^H}
\left\{6|A_q^H|+|A_{<q}^H|\right\}
-
\max_{q\in\{0,1\}^H}
\left\{6|A_q^H|+|A_{>q}^H|\right\}
=
|A_{<p}^H|-|A_{>p}^H|,
\]
and similarly,
\[
\max_{q\in\{0,1\}^H}
\left\{6|B_q^H|+|B_{<q}^H|\right\}
-
\max_{q\in\{0,1\}^H}
\left\{6|B_q^H|+|B_{>q}^H|\right\}
=
|B_{<p}^H|-|B_{>p}^H|.
\]

Under the Match promise, every Alice point is smaller than every Bob
point. Since $p$ contains points from both parties, Alice cannot have
a point to the right of $p$, and Bob cannot have a point to the left
of $p$. Hence $|A_{>p}^H|=|B_{<p}^H|=0$. Moreover,
$\min\{|A_p^H|,|B_p^H|\}=n-F(H)$ implies
\[
\max\left\{n-|A_p^H|,n-|B_p^H|\right\}=F(H).
\]
Consequently, the difference between Alice's and Bob's unnoised
statistics is
\begin{align*}
&\left(|A_{<p}^H|-|A_{>p}^H|\right)
-
\left(|B_{<p}^H|-|B_{>p}^H|\right)\\
&\qquad=
|A_{<p}^H|+|B_{>p}^H|\\
&\qquad\geq
F(H)
\geq
\frac{19\alpha}{20}.
\end{align*}
Thus, under the Match promise, the unnoised statistic has positive
margin at least $19\alpha/20$. The final noise is smaller than this
margin under the corresponding good event, and hence the algorithm
outputs $\mathsf{Match}$.

Under the Far promise, the order is reversed, and the same argument
gives a negative margin of magnitude at least $19\alpha/20$.
Therefore, the algorithm outputs $\mathsf{Far}$.

For the privacy proof, each maximization has sensitivity at most $7$, since changing one record changes the term multiplied by $6$ by at most $6$ and the remaining count by at most $1$. Therefore, the difference of the two maximizations has sensitivity at most $14$, and allocating privacy budget $\epsilon/2$ to its release requires noise $\Lap(14/(\epsilon/2))=\Lap(28/\epsilon)$.
\end{proof}

\begin{corollary}[Mixed labels for thresholds]
\label{remark:mixed-label-thr}
There exists a universal constant $C>0$ such that the following holds. Let $\dataspace\subseteq\Reals$ be a finite ordered domain with $|\dataspace|=m\geq4$, and suppose that each party $P\in{A,B}$ holds a labeled dataset $S_P\in(\dataspace\times{\pm1})^n$. Define 
$
\opt_{\hypclass_{\mathrm{thr}}}^{\mathrm{mix}}(S_A,S_B)
=
\min_{\theta\in\Reals}
\sum_{P\in\{A,B\}}
\sum_{(x,y)\in S_P}
\indic{h_\theta(x)\neq y}.
$
Let $\epsilon\in(0,1]$, $\beta\in(0,1/3]$, and $t\in[n]$. If

$$
t\geq
C\frac{\log\log(m+2)}{\epsilon}
\log\left(\frac{\log\log(m+2)}{\beta\epsilon}\right),
$$
then there exists an $\epsilon$-private tester that distinguishes $\opt_{\hypclass_{\mathrm{thr}}}^{\mathrm{mix}}(S_A,S_B)=0$ from $\opt_{\hypclass_{\mathrm{thr}}}^{\mathrm{mix}}(S_A,S_B)\geq t$ with failure probability at most $\beta$.
\end{corollary}

\begin{proof}
For $P\in{A,B}$, let $S_P^-$ and $S_P^+$ denote the multisets of negative and positive points held by $P$, respectively. For $P,Q\in{A,B}$, write $q_{PQ}:=\opt_{\hypclass_{\mathrm{thr}}}(S_P^-,S_Q^+)$. We first observe that

$$
\max_{P,Q\in\{A,B\}}q_{PQ}
\leq
\opt_{\hypclass_{\mathrm{thr}}}^{\mathrm{mix}}(S_A,S_B)
\leq
\sum_{P,Q\in\{A,B\}}q_{PQ}.
$$

The first inequality holds because each $q_{PQ}$ optimizes the threshold error over a subset of the pooled records. For the upper bound, for each pair $(P,Q)$, let $\theta_{PQ}$ achieve $q_{PQ}$ and let $E_{PQ}$ be the records it misclassifies, so $|E_{PQ}|=q_{PQ}$. Set $R=\bigcup_{P,Q}E_{PQ}$. Consider any negative point $x$ and positive point $y$ that do not belong to $R$. If $x\in S_P^-$ and $y\in S_Q^+$, then both points are correctly classified by $\theta_{PQ}$, and hence $x<\theta_{PQ}\leq y$. Therefore, every surviving negative point is smaller than every surviving positive point, so the records outside $R$ are jointly realizable by a threshold. It follows that $\opt_{\hypclass_{\mathrm{thr}}}^{\mathrm{mix}}(S_A,S_B)\leq |R|\leq\sum_{P,Q}q_{PQ}$.

Let $\tau=\lceil t/4\rceil$. In the Match case, every $q_{PQ}$ is zero. In the Far case, the preceding upper bound gives $\sum_{P,Q}q_{PQ}\geq t$, and hence some $q_{PQ}$ is at least $\tau$. We therefore perform four tests and return $\mathsf{Far}$ if any of them returns $\mathsf{Far}$.

The cross-party quantities $q_{AB}$ and $q_{BA}$ are tested using \cref{alg:thresh-main}, with Bob playing the negative-holder role in the second execution. To obtain equal-length inputs, fix public points $x_-<\min\dataspace$ and $x_+>\max\dataspace$. In a cross-party test, the negative holder replaces every positive record by $x_-$, while the positive holder replaces every negative record by $x_+$. The resulting datasets have size $n$, and their threshold optimum is exactly the corresponding $q_{PQ}$. Indeed, an optimal threshold may be chosen in $(x_-,x_+]$, in which case all dummy records are classified correctly. This preprocessing is recordwise and therefore maps neighboring labeled datasets to neighboring padded datasets. Notice that the enlarged domain has size $m+2$.

For a same-party quantity $q_{PP}$, replacing one labeled record changes the error of every threshold by at most one, and hence $q_{PP}$ has sensitivity one. Party $P$ releases $\widetilde q_{PP}:=q_{PP}+\Lap(4/\epsilon)$ and returns $\mathsf{Far}$ precisely when $\widetilde q_{PP}\geq\tau/2$. This release is $\epsilon/4$-differentially private and distinguishes $q_{PP}=0$ from $q_{PP}\geq\tau$ with failure probability at most $\beta/4$, provided that $\tau\geq(8/\epsilon)\log(4/\beta)$.

We run the two cross-party testers with parameters $(\epsilon/4,\beta/4)$ and allocate the same privacy and failure budgets to the two same-party tests. By composition, the resulting protocol is $\epsilon$-two-party differentially private, and by a union bound its failure probability is at most $\beta$. After increasing the universal constant $C$, the assumptions of the corollary ensure that $\tau$ satisfies the conditions of \cref{thm:threshold-upper-bound} with domain size $m+2$ and parameters $(\epsilon/4,\beta/4)$. This proves the result.
\end{proof}

\section{Halfspaces in High Dimensions}

In this part we focus on the halfspace in high dimensions. More precisely, $\dataspace\subseteq \Reals^d$. Also, without loss of generality, we assume Alice and Bob add a dummy dimension to their datapoints and therefore we can represent the class of halfspaces with 
\begin{equation} \label{eq:hs-def}
\hypclass_{\text{hs}}=\{h_{w}: \Reals^d \to \{\pm 1\}~|~ \exists w \in \Reals^{d}~\text{s.t.}~\text{$h_{w}(x)=\text{sign}\left(\inner{w}{x}\right)$}\}.
\end{equation}

\subsection{Lower bound for half spaces}

In this section, we prove an information-theoretic lower bound for $\hypclass_{\text{hs}}$. The argument for the lower bound is rather technical and it involves many steps. We first provide an overview of the proof and defer the formal proof to Appendix in \cref{appx:pf-hs-lb}. In this informal overview, we only focus on the case of large $n$, i.e., $n\geq C \frac{md \log(m/d)}{\epsilon}$ to highlight main ideas. With slight modifications, we can extend the lower bound to all values of $n$ as shown in \cref{appx:hs-sample-sizes} and \cref{thm:hs-sample-size}.

\subsubsection{Overview of the lower bound proof.} \label{sec:hs-lower-overview}

The proof follows the same template as that of
\cref{thm:threshold-lower-bound}. We take a binary predicate with high
information complexity and embed it into the promise problem, so that the
verdict of the promise problem determines the value of the predicate, and
then derive the lower bound from the connection between DP and information complexity from
\citep{McGregorMironovPitassiReingoldTalwarVadhan10}. For half spaces,
however, constructing an embedding with a sufficiently large gap is
substantially more delicate.

Our starting viewpoint is that each of Alice's and Bob's data points may be
read as a \emph{linear constraint} on $w\in\Reals^d$, namely
$\inner{w}{x}\ge0$ or $\inner{w}{x}<0$. Viewed in the dual space, the
problem therefore becomes a linear program with finitely many
constraints: in the Match case the system is feasible, whereas in the Far
case it remains infeasible even after removing any $t-1$ of the
constraints. Achieving this robustness in the Far case is the main
technical difficulty of the proof, and it is what leads us to use
error-correcting codes.

\paragraph{Step~1: from inner product in a finite field to LP constraints.}
Let $s$ be an integer determined later and $q=2^s$. Alice holds a secret
$u\in\mathbb{F}_q^k$ and Bob holds a secret $v\in\mathbb{F}_q^k$, and they
wish to compute
\begin{equation}
\label{eq:inner-prod-trace-fq}
F(u,v)=\bigoplus_{i=1}^{k}
\operatorname{Tr}_{\mathbb{F}_q/\mathbb{F}_2}(u_iv_i)\in\{0,1\}.
\end{equation}
$F$ can be thought of as the trace of an inner product over $\mathbb{F}_q$. The internal information complexity of computing $F$ is $\Theta(k\log q)$; for completeness we provide a proof in \cref{lem:ic-trace}. 
We now build a linear program whose feasibility decides
\cref{eq:inner-prod-trace-fq}. The main challenge here is that the computation in \cref{eq:inner-prod-trace-fq} consists of non-linear components over $\mathbb{F}_q$ while we want to convert it to a feasibility problem in $\Reals^d$. To do so, we embed the $q$ elements of $\mathbb{F}_q$ as the vertices of a convex polygon $P_q\subset\Reals^2$ lying on a parabola (so each symbol uses two real coordinates) and represent each finite-field relation by the convex hull of its satisfying assignments (for details see \cref{fig:probala}).
The resulting LP constraints come in two kinds: \underline{A public family}, known to both parties and independent of $u$ and $v$, performs the
computation itself: it holds the variables in the right ranges, forms the
$k$ trace products, and accumulates their parity, with one last row
requiring the parity to be zero. \underline{A private family} encodes Alice's and Bob's secrets: each coordinate of $u$ and $v$
contributes exactly one constraint, which we call a \emph{pin} (the naming is inspired by our way of converting elements of $\mathbb{F}_q$ into a linear program in \cref{fig:probala}). Once all $2k$ of these are present, the fixed rows propagate
the computation and the program is feasible if and only if $F(u,v)=0$. We give the details of
constructing such a program in \cref{appx:pf-hs-lb}.
Reading this back through \cref{def:private-tester}, we obtain a Match
instance whenever $F(u,v)=0$ and a Far instance whenever $F(u,v)=1$.
This feasibility equivalence alone certifies only a Far gap of one. Repeating each linear constraint $L$ times
guarantees $\opt_{\hypclass_{\mathrm{hs}}}\ge L$, but the encoding
diameter of \cref{lem:metric-dp-to-information} rises to $2kL$. Then, by the connection of the IC and encoding diameter from \cref{lem:metric-dp-to-information}, we obtain only $t=\Omega(\log q/\varepsilon)$.
The shortcoming of the simple repetition is that the Far gap grows only with the replication $L$, while the diameter
grows with $kL$. Can we instead design a program that is already robust to
deleting $D \geq 1$ of its private pins (i.e., erasing $D$ symbols) before any replication?

\paragraph{Step~2: turning code robustness into LP rigidity.}
What we need is that, as long as the public constraints remain enforced,
removing fewer than $D$ pins still leaves the final answer intact. We
therefore encode $u$ and $v$ using a public \emph{systematic linear code}
with generator $G=[\id{k}\mid A]\in\mathbb{F}_q^{k\times K}$ where $K>k$. Let $c = uG$ and $c'=vG$; since the code is systematic, the first $k$ coordinates of $c$
and $c'$ are $u$ and $v$ themselves, so the trace products of
\cref{eq:inner-prod-trace-fq} are still formed from these coordinates
alone. Three properties are needed from the code and its parity-check
representation: constant rate, so that $k=\Theta(K)$; linear
\emph{LP-erasure rigidity}, meaning the LP tolerates up to $D=\Theta(K)$ removed pins; and constant check
degree, for a reason that will become clear below. 

The key subtlety is that local parity-check constraints only enforce a convex relaxation of the code, which can admit fractional solutions even when the code has large minimum distance. We therefore use a code whose parity-check graph is a constant-degree unique-neighbor expander \citep{feldman2005using}: whenever fewer than $D=\Theta(K)$ pins are removed, any nonempty set of unpinned coordinates has an adjacent check touching it in a single coordinate, whose remaining pinned coordinates force that variable back to its true symbol (\cref{lem:forced-relation}). Thus the public constraints still uniquely force $c$ and $c'$.
Two counts govern the size of the construction, and separating them is
what makes a large alphabet affordable.
First, since each $\mathbb{F}_q$ symbol is embedded in $\Reals^2$, the program uses $d_{\mathrm{LP}}:=4K+2k+1=\Theta(K)$ variables ($2K$ per codeword, $k$ trace, and $k+1$ parity variables), \emph{independent of $q$}, allowing us to choose $K=\Theta(d)$. Second, each distinct LP inequality corresponds to a point in the public data space $\mathcal{X}$ of size $m$. Because each check involves only $\rho=O(1)$ symbols, its convex hull in $\Reals^{2\rho}$ is described by $q^{O(\rho^2)}=q^{O(1)}$ inequalities; across all $O(K)$ checks and computation gates, this yields at most $CKq^{a}$ distinct records for universal constants $C,a$ (replication increases multiplicities, not distinct records). The alphabet therefore costs records but never dimensions, which is what will let us
take $q$ large.

\paragraph{Step~3: producing the Far gap.}

Converting each LP inequality into a labeled data point, we repeat each private pin $L$ times and each public constraint $J$ times. When $F(u,v)=0$, all constraints are simultaneously satisfiable. When $F(u,v)=1$, removing fewer than $\min\{J,DL\}$ constraints leaves at least one copy of every public constraint intact and removes all $L$ copies of fewer than $D$ private pins; by Step~2, the remaining LP remains infeasible. Hence
\[
F(u,v)=0 \ \Longrightarrow\ \opt_{\hypclass_{\mathrm{hs}}}(S_A,S_B)=0,
\qquad
F(u,v)=1 \ \Longrightarrow\
\opt_{\hypclass_{\mathrm{hs}}}(S_A,S_B)\ge\min\{J,DL\}.
\]
For a hypothetical $t$-tester, take $J=\lceil t\rceil$ and $L=\lceil t/D\rceil$, which guarantees a Far gap of at least $t$. Since only the $K$ private pins depend on each party's secret, the encoding diameter is $\Delta_A,\Delta_B\le KL = O(t+K)$ (as $D=\Theta(K)$). The information lower bound and \cref{lem:metric-dp-to-information} then give
\[
\Omega(K\log q) \le O\bigl(\varepsilon(t+K)\bigr),
\]
which forces $t=\Omega(K\log q/\varepsilon)$ for sufficiently small $\varepsilon$. Finally, we choose $K=\Theta(d)$ and the largest power of two $q$ satisfying $BKq^a\log_2 q\le m/8$ for a universal constant $B$. This keeps the number of distinct records at most $m$ and gives $\log q=\Theta(\log(m/d))$, yielding $t=\Omega\bigl(\frac{d\log(m/d)}{\varepsilon}\bigr)$. Moreover, when $t=O(K\log q/\varepsilon)$, the total number of repeated records is $O(Kq^a J + KL) = O(md/\varepsilon)$, which fits within datasets of size $n\ge Cmd/\varepsilon$ after padding with input-independent records.

\begin{theorem}[Lower bound for halfspaces]
\label{thm:hs-lower-bound}
There are universal constants $c,C,C_0,\varepsilon_0>0$ and
$d_0\in\Naturals$ such that, for every $d,m\in\Naturals$ with
$d\ge d_0$ and $m\ge C_0d$, there is a public domain
$\dataspace_{d,m}\subseteq\Reals^d$ with $|\dataspace_{d,m}|=m$
such that for every $0<\varepsilon\le\varepsilon_0$,
every integer $n \geq C_0 \frac{md\log(m/d)}{\epsilon}$, and every resolution $t\in[n]$,
an $\varepsilon$-private tester for $\hypclass_{\mathrm{hs}}$ at dataset
size $n$ over $\dataspace_{d,m}$ must satisfy
\[
  t\ge c\frac{d\log(m/d)}{\varepsilon}.
\]
The domain is chosen independently of $n,t,\varepsilon$.
\end{theorem}

\begin{proof}
Take the fixed domain from \cref{lem:hs-fixed-domain}, and let
$R=\Theta(d)$ be its largest block length. Let $B,a$ be the constants
in \cref{lem:hs-scale}. Among powers of two
$q\ge2$, choose the largest satisfying
\begin{equation}
\label{eq:hs-retuned-alphabet}
  BRq^a\log_2q\le m/8.
\end{equation}
For $C_0$ sufficiently large, $q=2$ is admissible. Every such alphabet
is included in $\dataspace_{d,m}$, since $q^a\le m/(8BR)$.

Put $s=\log_2q\ge1$. Maximality gives
\[
  \frac{m}{8BR}<(2q)^a(s+1)
  \le 2^{a+1}q^a s.
\]
Taking logarithms and using $\log_2s\le s$ shows that
$s=\Omega(\log(m/R))=\Omega(\log(m/d))$.
The reverse bound follows from \eqref{eq:hs-retuned-alphabet} and
$R=\Theta(d)$, so $s=\Theta(\log(m/d))$.

By \cref{lem:hs-scale}, the construction at scale $(R,q)$ forces
$t\ge bRs/\varepsilon$ whenever
$n\varepsilon\ge A R^2q^a s$, for universal $A,b>0$.
The retuned alphabet satisfies
\[
  A R^2q^a s\le\frac{A}{8B}Rm\le Cdm
\]
for a sufficiently large universal $C$. Thus the assumed sample size
suffices, with each dataset padded to exactly $n$ as in
\cref{lem:hs-scale}. Since $R=\Theta(d)$ and
$s=\Theta(\log(m/d))$, we obtain the claimed lower bound.
\end{proof}

\begin{remark}
The surprising behavior of this lower bound compared to the optimal gap for thresholds in \cref{thm:threshold-lower-bound,thm:threshold-upper-bound} is that the dependence on $m$ is exponentially larger, i.e., $\log\log(m)$ for threshold and $\log(m)$ for half spaces. Also, $t$ exhibits (at least) a linear dependence on $d$.
\end{remark}

\subsection{Algorithm for Half Spaces}
\subsubsection{Structural Results and Tools}
As our lower bound in \cref{thm:hs-lower-bound} shows we need to assume that $\dataspace=\{x_1,\dots,x_m\}\subset \Reals^d$ is a finite set. Let $|\dataspace|=m$. Towards developing our algorithm, we mainly focus on the dual space: consider the dual space denoted by $\mathcal{W}=\Reals^d$  where we map each $h_w \in \hypclass_{\text{hs}}$ to $w \in \Reals^d$. 

The cardinality of $\mathcal{W}$ is unbounded. However, we only need to consider the projection of $\dataspace$ using $\hypclass_{\text{hs}}$, which means that two $w\sim w'$ are equivalent if and only if the predictions of $h_{w}$ and $h_{w'}$ over $\dataspace$ be the same. The next lemma shows that considering this \emph{equivalent class}, we can reduce $\mathcal{W}$ to a discrete set. The proof is an application of Sauer–Shelah lemma \citep{sauer1972density,shelah1972combinatorial}. 
\begin{lemma}[Finite dual discretization]
\label{lem:finite-dual-discretization}
Let $\dataspace=\{x_1,\ldots,x_m\}\subseteq\Reals^d$. There exists a
finite set $\mathcal{W}_{\mathrm{disc}}\subseteq\Reals^d$ such that, for
every $w\in\Reals^d$, some $w'\in\mathcal{W}_{\mathrm{disc}}$ satisfies
$h_w(x_i)=h_{w'}(x_i)$ for every $i\in[m]$. Moreover,
$
|\mathcal{W}_{\mathrm{disc}}|
\leq
(2m)^d.
$
\end{lemma}

For each $w \in \mathcal{W}_{\mathrm{disc}}$, define the local error of $h$ on Alice and Bob datasets as
\[
f_A(w)=\sum_{x \in S_A} \indic{h_w(x) \neq -1}~\qquad~f_B(w)=\sum_{x \in S_B} \indic{h_w(x)\neq 1},\]
and using this notation, we have $\opt_{\hypclass_{\text{hs}}}=\min_{w \in \mathcal{W}_{\mathrm{disc}}}\{f_A(w) + f_B(w)\}$. For every $k \in [n]$, we also define the set of local low-error classifiers on Alice and Bob's side as
\[
G_A^{k}=\{w \in \mathcal{W}_{\mathrm{disc}}~:~f_A(w)\leq k\} ~\qquad~G_B^{k}=\{w \in \mathcal{W}_{\mathrm{disc}}~:~f_B(w)\leq k\}.\]
Note that $G_A^{k}$ and $G_B^{k}$ as sets live in the dual space. Also, since $\mathcal{W}_{\mathrm{disc}}$ is a finite set, these sets are generally non-convex in $\Reals^d$. However, when we consider the convex hull of them, we have an important property.
\begin{lemma}
\label{lem:disjoint-hull}
Fix $n,d\in\Naturals$, datasets $S_A,S_B\in\dataspace^n$, and
$t\in[n]$. Suppose that $\opt_{\hypclass_{\mathrm{hs}}}(S_A,S_B)\geq t$. Then, for every $k\in\Naturals_0$ satisfying $k\leq\frac{t-1}{2(d+1)}$, we have
$
\conv(G_A^k)
\cap
\conv(G_B^k)
=
\emptyset.
$
\end{lemma}

\begin{proof}
Suppose toward a contradiction that there exists
$
v\in
\operatorname{conv}(G_A^k)
\cap
\operatorname{conv}(G_B^k).
$
By Carath\'eodory's theorem \cref{lem:carathedory}, there exist
$a_1,\ldots,a_r\in G_A^k$, where $r\leq d+1$, such that $v$ is a
convex combination of $a_1,\ldots,a_r$. If $v$ misclassifies a point
in $S_A$, then at least one $a_i$ also misclassifies that point.
Consequently,
\[
f_A(v)
\leq
\sum_{i=1}^r f_A(a_i)
\leq
(d+1)k.
\]
Applying the same argument to a representation of $v$ as a convex
combination of at most $d+1$ points from $G_B^k$ gives $f_B(v)\leq(d+1)k$. Therefore,
\[
f_A(v)+f_B(v)\leq2(d+1)k\leq t-1,
\]
contradicting
$\opt_{\hypclass_{\mathrm{hs}}}(S_A,S_B)\geq t$.
\end{proof}

Our algorithm employes a geometric tool that is called \emph{container}. It is a special case of the cover where we also assume that the cover \emph{includes} the set. We use the following result from \cite{braverman2019convex,dutta2021uniform}.

\begin{lemma}[Halfspace containers~\citep{braverman2019convex,dutta2021uniform}]
\label{lem:halfspace-containers}
There exists a universal constant $c>0$ such that the following holds.
For every finite set $\mathcal{S}\subseteq\Reals^d$ and every $\eta\in(0,1)$,
there exists a family $\mathcal{C}(\mathcal{S},\eta)\subseteq 2^\mathcal{S}$ satisfying
$\
|\mathcal{C}(\mathcal{S},\eta)| \leq \left(\frac{2}{\eta}\right)^{cd}.
$
Moreover, for every affine halfspace $H\subseteq\Reals^d$, there exists
$C\in\mathcal{C}(\cal{S},\eta)$ such that
\[
H\cap \mathcal{S}\subseteq C
\qquad\text{and}\qquad
|C \setminus H|\leq\eta|\mathcal{S}|.
\]
\end{lemma}

\begin{remark}
An special case of \cref{lem:halfspace-containers} that will be used in the sequel is as follows: consider a half space whose positive region is less than the whole space, i.e., $|H\cap \mathcal{S}|\leq |\mathcal{S}|/2$. Then, from $|C \setminus H|$, we have $|C \setminus H|=|C|-|H \cap \mathcal{S}|\geq |C|-\frac{|\mathcal{S}|}{2}$. Therefore, we have $|C| \leq (\frac{1}{2}+\eta)|S|$.
\end{remark}

\subsubsection{Main Algorithm} 
In this section, we present the algorithm for half spaces. We begin with providing an overview of the technical ideas. 

\paragraph{Overview of the algorithm.} Our algorithm is a private elimination procedure over the finite dual
space $\mathcal{W}_{\mathrm{disc}}$. At round $r$, the parties maintain a public set $\mathcal{W}_r$ of candidate halfspaces and attempt to replace it by a substantially smaller set. Ideally, a party would find a halfspace in the dual space that contains all of its low-error candidates but at most half of $\mathcal{W}_r$. Such a halfspace defines a safe cut, but the family of possible halfspaces is too large to search privately. The container lemma in \cref{lem:halfspace-containers} replaces these halfspace cuts by a small public family: whenever such a halfspace exists, the family contains a set covering it and
containing at most $3|\mathcal{W}_r|/4$ candidates.

For $P\in\{A,B\}$, let $G_P^k(\mathcal{W}_r)$ denote the candidates in $\mathcal{W}_r$
with local error at most $k$. We assign each container a
low-sensitivity cost measuring the quality of the best halfspace candidate it
excludes; its cost is zero exactly when it contains all of
$G_P^k(\mathcal{W}_r)$. Alice and Bob can therefore use the exponential
mechanism to privately search the public container family for a safe
shrinking cut.

The high level goal is as follows: The two promise cases interact with this mechanism in opposite ways, and the stopping rule reads off which occurred: a Match instance keeps $\mathcal{W}_r$ nonempty forever, while a Far instance is driven to $\mathcal{W}_r = \emptyset$ within $\log\left(|\mathcal{W}_{\text{disc}}|\right)$ rounds, thanks to \cref{lem:finite-dual-discretization} that shows $|\mathcal{W}_{\text{disc}}|$ is finite.

\paragraph{Far case.} By \cref{lem:disjoint-hull},
$\conv(G_A^k(\mathcal{W}_r))$ and $\conv(G_B^k(\mathcal{W}_r))$ are disjoint. A separating
hyperplane therefore places one party's low-error region in a
halfspace containing at most half of $\mathcal{W}_r$. By
\cref{lem:halfspace-containers}, that party has a zero-cost real
container of size at most $3|\mathcal{W}_r|/4$ in the public menu. On a high-probability good
event that we define later,  each party selects a real container,
not necessarily the particular zero-cost one. Since every real member
of the menu has size at most $3|\mathcal{W}_r|/4$, the active set shrinks by a
constant factor. See \cref{fig:dual-space}.

\paragraph{Match case.} Here the mechanism above can fail: the two low-error
regions may overlap and be scattered across $\mathcal{W}_r$, so neither
is guaranteed to lie in a halfspace holding at most half the candidates. We
therefore add a dummy action $\bot$ to the menu, which proposes no cut
and lets a party abstain when no safe container exists; both parties
abstaining is the protocol's signal to output $\mathsf{Match}$. The cost
of $\bot$ is fixed at $k/2$, wedged between the cost of a safe container
($0$) and that of any container excluding a consistent classifier
($\ge k+1$). This calibration makes abstention the preferred move
exactly when no safe cut exists. Concretely, fix a classifier $h^\star$ consistent
with both datasets. On a high probability event, that we define in the proof, each party either selects $\bot$
or selects a real container that still contains $w^\star$, since any
container excluding $w^\star$ costs at least $k+1 > k/2$. If both select
$\bot$ the protocol outputs $\mathsf{Match}$; otherwise $w^\star$
survives into $\mathcal{W}_{r+1}$. If the protocol reachs the $\log(|\mathcal{W}_{\text{disc}}|$ step, then the resulting active set is not empty since it contains $w^\star$. See \cref{fig:dual-space}.

\begin{algorithm}
\caption{Data Valuation for Half Spaces: $\mathcal{A}_\mathsf{val-hs}$}
\label{alg:hs-main}
\textbf{Input:} Alice holds $S_A\in \dataspace^n$, Bob holds $S_B\in \dataspace^n$; parameters $\epsilon,\beta$.\\
\textbf{Output:} $\mathsf{Far}$ or $\mathsf{Match}$.
\begin{algorithmic}[1]
\State $\mathcal{W}_{0} \gets \mathcal{W}_{\text{disc}}$
\State $Q\gets
\left\lceil\log_{4/3}|\mathcal{W}_{\mathrm{disc}}|\right\rceil+2$
\State $\alpha\gets
\displaystyle
\frac{2Q}{\epsilon}
\left(
\ln(8^{cd}+1)+\ln\frac{2Q}{\beta}
\right)$ \Comment{\textcolor{black}{$c$ is a universal constant.}}
\State $k\gets
\displaystyle
\left\lfloor\frac{t-1}{2(d+1)}\right\rfloor$ \Comment{\textcolor{black}{see \cref{lem:disjoint-hull}}}.
\State $\epsilon_0 \gets \epsilon/Q$.
 \For{$r = 0, \dots, Q-1$}
    \State Construct container family $\mathcal{C}\big(\mathcal{W}_r,\frac{1}{4}\big)$ and define the \emph{public menu} $$\mathcal{R}_{r} = \left\{ C \in \mathcal{C}\big(\mathcal{W}_r,\frac{1}{4}\big)~\big|~|C|\leq \frac{3}{4}|\mathcal{W}_r|\right\} \cup \{\bot\}.$$ \Comment{\textcolor{black}{ See \cref{lem:halfspace-containers}}}
    \State For $P\in\{A,B\}$ and $C\in \mathcal{W}_r$, define
    \[
    q_{P,\mathcal{W}_r}(C;S_P)
    \gets
    \begin{cases}
    \displaystyle
    \max_{h\in \mathcal{W}_r\setminus C}
    \bigl(k+1-f_P(h)\bigr)_+,
    & C\neq\bot,\\[2mm]
    \frac{k}{2},
    & C=\bot,
    \end{cases}
    \]
    \State Alice samples $C_{A,r} \sim \mathsf{EM}\bigg(q_{A,\mathcal{W}_r},\mathcal{R}_{r},S_A,\epsilon_0\bigg)$ and releases it. \Comment{\textcolor{black}{ See \cref{lem:exponential-mechanism}}}
    \State Bob samples $C_{B,r} \sim \mathsf{EM}\bigg(q_{B,\mathcal{W}_r},\mathcal{R}_{r},S_B,\epsilon_0\bigg)$ and releases it. \Comment{\textcolor{black}{ See \cref{lem:exponential-mechanism}}}
    \If{$C_{A,r} = C_{B,r} = \bot$}
        \State \Return $\mathsf{Match}$
    \EndIf
    \If{$C_{A,r} \neq \bot$}
        \State $\mathcal{W}_{r+1} \gets C_{A,r}$ 
    \ElsIf{$C_{A,r} = \bot$ \textbf{and} $C_{B,r} \neq \bot$}
        \State $\mathcal{W}_{r+1} \gets C_{B,r}$ 
    \EndIf
\EndFor

 \State \Return $\mathsf{Far}$ if $\mathcal{W}_{Q}=\emptyset$; otherwise \Return $\mathsf{Match}$.
\end{algorithmic}
\end{algorithm}

\begin{theorem}
\label{thm:halfspace-upper-bound}
Let $c$ be a universal constant. Fix a finite data space $\dataspace\subseteq\Reals^d$ with
$|\dataspace|=m$, and let $\epsilon \in (0,1]$, $\beta\in(0,1)$, and
$t\in[n]$. Let
\[
t \geq 
c\left(
\frac{d^2\log(2m)}{\epsilon}
\left(
d+\log\frac{d\log(2m)}{\beta}
\right)
\right).
\]
then $\mathcal{A}_{\mathsf{val-hs}}$ in \cref{alg:hs-main} is an $\epsilon$-private tester for
the $t$-promise problem over $\hypclass_{\mathrm{hs}}$ with success
probability at least $1-\beta$. 
\end{theorem}
\begin{proof}
The privacy proof is straightforward: in each round, Alice and Bob use exponential mechanism with $\epsilon_0$. The maximum number of rounds is $Q$, therefore, the composition theorem of pure-DP shows that the output satisfies $\epsilon$ two-view DP. Next we turn to the utility proof.

First, we define an event and then the rest of the analysis is based on assuming the event holds. For $P\in\{A,B\}$ and $r\in\{0,\ldots,Q-1\}$, lets define an event that shows the protocol reaches round $r$ and the exponential mechanism is accurate. More precisely,
\[
\mathcal{B}_{P,r}
:=
\{r<T\}
\cap
\left\{
q_{P,\mathcal{W}_r}(C_{P,r};S_P)
-
\min_{C\in\mathcal{R}_r}
q_{P,\mathcal{W}_r}(C;S_P)
>
\alpha
\right\}.
\]
Then define
\begin{equation} \label{eq:good-event-hs}
\mathcal{G}:=\bigcap_{r=0}^{Q-1}\left(\mathcal{B}_{A,r}^{\,c}\cap\mathcal{B}_{B,r}^{\,c}\right).
\end{equation}
Thus, on $\mathcal{G}$, every exponential-mechanism call that is
actually executed has additive cost error at most $\alpha$. It is easy to see that by the union bound and \cref{lem:exponential-mechanism}, we have $\Pr(\mathcal{G}^c)\leq\sum_{r=0}^{Q-1}\left(\Pr(\mathcal{B}_{A,r})+\Pr(\mathcal{B}_{B,r})\right)\leq
2Q\cdot\frac{\beta}{2Q}=
\beta$. For the rest of the proof, we assume $\mathcal{G}$ holds.

 \paragraph{Case 1: Far holds.}
We first assume that
$\opt_{\hypclass_{\mathrm{hs}}}(S_A,S_B)\geq t$. Consider any round $r$
reached by the protocol with $\mathcal{W}_r\neq\emptyset$. Since
$k=\lfloor(t-1)/(2(d+1))\rfloor$, \cref{lem:disjoint-hull} gives
$\conv(G_A^k(\mathcal{W}_r))\cap\conv(G_B^k(\mathcal{W}_r))=\emptyset$.

If both low-error sets are nonempty, strict separation gives opposite
affine halfspaces containing $\conv(G_A^k(\mathcal{W}_r))$ and $\conv(G_B^k(\mathcal{W}_r))$,
respectively. We may perturb the separating hyperplane so that its
boundary avoids the finite set $\mathcal{W}_r$. The two halfspaces then partition
$\mathcal{W}_r$, and one contains at most $|\mathcal{W}_r|/2$ points. Thus, for some $P\in\{A,B\}$, there
is an affine halfspace $H$ such that
$G_P^k(\mathcal{W}_r)\subseteq H\cap \mathcal{W}_r$ and
$|H\cap \mathcal{W}_r|\leq |\mathcal{W}_r|/2$.

By \cref{lem:halfspace-containers}, there is a container
$C^\star\in\mathcal{C}(\mathcal{W}_r,1/4)$ containing $H\cap \mathcal{W}_r$ and satisfying
$|C^\star|\leq3|\mathcal{W}_r|/4$. Hence $C^\star\in\mathcal{R}_r$,
$G_P^k(\mathcal{W}_r)\subseteq C^\star$, and therefore
$q_{P,\mathcal{W}_r}(C^\star;S_P)=0$. On the event $\mathcal{G}$, the container
released by party $P$ consequently satisfies
$q_{P,\mathcal{W}_r}(C_P^r;S_P)\leq\alpha$. Since
$q_{P,\mathcal{W}_r}(\bot;S_P)=k/2>\alpha$, it follows that
$C_{P,r}\neq\bot$. Thus, the protocol cannot return $\mathsf{Match}$ in
this round.

The protocol therefore either returns $\mathsf{Far}$ because the active
set is empty or replaces $\mathcal{W}_r$ by a real member of $\mathcal{R}_r$.
In the latter case, $|W_{r+1}|\leq3|\mathcal{W}_r|/4$. Consequently, if the
protocol completes all $Q$ rounds, then
$|W_Q|\leq(3/4)^Q|W_0|<1$, where the last inequality follows from
$Q=\lceil\log_{4/3}|W_0|\rceil+2$. Since $W_Q$ is finite, we must have
$W_Q=\emptyset$, and the protocol returns $\mathsf{Far}$.

\paragraph{Case 2: Match holds.}
Assume that $\opt_{\hypclass_{\mathrm{hs}}}(S_A,S_B)=0$, and fix
$w^\star\in W_0$ such that $f_A(w^\star)=f_B(w^\star)=0$. We claim that
if the protocol reaches round $r$ without having already returned
$\mathsf{Match}$, then $w^\star\in \mathcal{W}_r$. The claim holds initially
because $W_0=\hypclass_{\mathrm{hs}}$.

Suppose that $w^\star\in \mathcal{W}_r$. For each $P\in\{A,B\}$, since
$\bot\in\mathcal{R}_r$ and $q_{P,\mathcal{W}_r}(\bot;S_P)=k/2$, the event
$\mathcal{G}$ implies that
$q_{P,\mathcal{W}_r}(C_P^r;S_P)\leq k/2+\alpha\leq k-1$, where the last inequality
follows from $k\geq2\alpha+2$. However, every real container
$C\in\mathcal{R}_r$ excluding $w^\star$ has
$q_{P,\mathcal{W}_r}(C;S_P)\geq(k+1-f_P(h^\star))_+=k+1$. Thus, every real
container released by either party contains $w^\star$.

If both parties release $\bot$, the protocol correctly returns
$\mathsf{Match}$. Otherwise, $W_{r+1}$ is one of the released real
containers and therefore contains $w^\star$. This proves the claim by
induction. Consequently, the active set never becomes empty. If the
protocol does not terminate earlier, then $w^\star\in W_Q$, and the
protocol returns $\mathsf{Match}$ after the final round.

Combining the two cases, the protocol is correct whenever
$\mathcal{G}$ holds. Since $\Pr(\mathcal{G})\geq1-\beta$, its success
probability is at least $1-\beta$.
\end{proof}

\begin{corollary}[Mixed labels for halfspaces]
\label{remark:mixed-label-hs}
Let $c$ be the universal constant in \cref{thm:halfspace-upper-bound}. Fix a finite data space $\dataspace\subseteq\Reals^d$ with $|\dataspace|=m$, and suppose that each party $P\in{A,B}$ holds a labeled dataset $S_P\in(\dataspace\times{\pm1})^n$. Define $
\opt_{\hypclass_{\mathrm{hs}}}^{\mathrm{mix}}(S_A,S_B)
:=
\min_{w\in\Reals^d}
\sum_{P\in\{A,B\}}
\sum_{(x,y)\in S_P}
\indic{h_w(x)\neq y}.
$
Let $\epsilon\in(0,1]$, $\beta\in(0,1)$, and $t\in[n]$. If

$$
t\geq
c\frac{d^2\log(2m)}{\epsilon}
\left(
d+\log\frac{d\log(2m)}{\beta}
\right),
$$

then \cref{alg:hs-main}, after redefining the local error as $f_P(w):=\sum_{(x,y)\in S_P}\indic{h_w(x)\neq y}$, is an $\epsilon$-private tester that distinguishes $\opt_{\hypclass_{\mathrm{hs}}}^{\mathrm{mix}}(S_A,S_B)=0$ from $\opt_{\hypclass_{\mathrm{hs}}}^{\mathrm{mix}}(S_A,S_B)\geq t$ with success probability at least $1-\beta$.
\end{corollary}

\begin{proof}
The finite dual discretization is unchanged, since it depends only on the predictions over $\dataspace$. Redefine $G_P^k(\mathcal{W}):={w\in\mathcal{W}:f_P(w)\leq k}$. To verify \cref{lem:disjoint-hull}, suppose that $v\in\conv(G_A^k(\mathcal{W}))\cap\conv(G_B^k(\mathcal{W}))$. By Carath'eodory's theorem, write $v=\sum_{i=1}^r\lambda_i a_i$ for some $r\leq d+1$ and $a_1,\ldots,a_r\in G_A^k(\mathcal{W})$. If $h_v$ misclassifies a labeled record, then at least one $h_{a_i}$ also misclassifies it. Therefore, $f_A(v)\leq\sum_{i=1}^r f_A(a_i)\leq(d+1)k$. The same argument gives $f_B(v)\leq(d+1)k$, and hence, for $k=\lfloor(t-1)/(2(d+1))\rfloor$,

$$
\opt_{\hypclass_{\mathrm{hs}}}^{\mathrm{mix}}(S_A,S_B)
\leq f_A(v)+f_B(v)
\leq 2(d+1)k
\leq t-1,
$$

contradicting the Far assumption. Thus, the utility proof of \cref{thm:halfspace-upper-bound} applies unchanged.

Finally, replacing one labeled record changes $f_P(w)$ by at most one for every $w$, so $q_{P,\mathcal{W}}(C;S_P)$ retains sensitivity one. Therefore, the same exponential-mechanism and composition arguments give $\epsilon$-two-party differential privacy.
\end{proof}

\subsubsection{Complementary bounds for small data regime}

An interesting aspect of the guarantee in \cref{thm:halfspace-upper-bound} is that $t$ does not depend on the
dataset size $n$. However, the cost of such a result is that we need $n \geq \tilde{O}\left(\frac{d^3 \log(m)}{\epsilon}\right)$. In this section, we show that when $n$ is small, there are simple algorithms that extend the private tester to almost all non-trivial dataset sizes.

\paragraph{A tester with $t=n$.} In this section, we give a simple algorithm to show that, as long as $n\gtrsim 1/\epsilon^2$, we can design a tester that achieves $t=n$. Note that this is the largest resolution one can ask for: classifier $w=0$ labels every point as positive and makes exactly $n$ mistakes, and therefore $\opt_{\hypclass_{\text{hs}}}(S_A,S_B)\leq n$ for all datasets.

The starting point is a simple characterization of when $\opt_{\hypclass_{\text{hs}}}(S_A,S_B)=n$. Note that we assume the homogeneous halfspaces in \cref{eq:hs-def} without a bias term.   For $x\neq 0$, let $R(x)=\{\lambda x:\lambda>0\}$ denote the ray through $x$, and let $\mathcal{R}=\{R(x):x\in\dataspace\}$ be the set of rays that contain at least one point of the domain. Since the domain is finite, this set is well-defined. Notice that $\mathcal{R}$ is public.

For every $R\in\mathcal{R}$, let $a_R$ and $b_R$ denote the number of Alice's and Bob's records that lie on $R$, respectively. The main observation is that a homogeneous halfspace assigns the same label to all the points on a ray. Using this observation, it is straightforward to see that
\begin{equation}
\label{eq:hs-t-equals-n-char}
\begin{aligned}
\opt_{\hypclass_{\text{hs}}}(S_A,S_B)=n &\iff a_R=b_R~\text{for all}~R\in\mathcal{R},\\
\opt_{\hypclass_{\text{hs}}}(S_A,S_B)=0 &\implies a_Rb_R=0~\text{for all}~R\in\mathcal{R}.
\end{aligned}
\end{equation}
In other words, in the Far case Alice and Bob have exactly the same histogram over the rays, while in the Match case their histograms have disjoint supports. Therefore, at $t=n$ the testing problem reduces to checking whether two histograms over $\mathcal{R}$ are identical, and this can be done using shared randomness as we show in \cref{alg:hs-t-equals-n}. The next theorem summarize the privacy and utility guarantee of this tester.

\begin{algorithm}
\caption{Testing at Resolution $t=n$ for Half Spaces: $\mathcal{A}_\mathsf{max-hs}$}
\label{alg:hs-t-equals-n}
\textbf{Input:} Alice holds $S_A=(X_1,\dots,X_n)\in\dataspace^n$, Bob holds $S_B=(Y_1,\dots,Y_n)\in\dataspace^n$; parameter $\epsilon$.\\
\textbf{Output:} $\mathsf{Far}$ or $\mathsf{Match}$.
\begin{algorithmic}[1]
\State Let $k$ be a large enough constant, and set $\tau\gets\sqrt{n}/2$.
\State Draw independent and uniform signs $\sigma_{j,R}\in\{\pm1\}$ for $j\in[k]$ and $R\in\mathcal{R}$. \Comment{{public randomness}}
\State Alice computes $a_R\gets|\{i\in[n]:X_i\in R\}|$ for every $R\in\mathcal{R}$, and releases
\[
Z_{A,j}\gets\sum_{R\in\mathcal{R}}\sigma_{j,R}\,a_R+\Lap(2k/\epsilon)\qquad\text{for all } j\in[k].
\]
\State Bob computes $b_R\gets|\{i\in[n]:Y_i\in R\}|$ for every $R\in\mathcal{R}$, and releases
\[
Z_{B,j}\gets\sum_{R\in\mathcal{R}}\sigma_{j,R}\,b_R+\Lap(2k/\epsilon)\qquad\text{for all } j\in[k].
\]
\If{$\max_{j\in[k]}|Z_{A,j}-Z_{B,j}|>\tau$}
    \State \Return $\mathsf{Match}$
\Else
    \State \Return $\mathsf{Far}$
\EndIf
\end{algorithmic}
\end{algorithm}

\begin{theorem}
\label{thm:hs-t-equals-n-mb}
There exists a universal constant $C>0$ such that the following holds. Fix a finite data space $\dataspace\subseteq\Reals^d\setminus\{0\}$ and $\epsilon>0$, and let $n\geq C/\epsilon^2$. Then there exists a non-interactive algorithm (presented in \cref{alg:hs-t-equals-n}) that is $\epsilon$-private tester with resolution $t=n$ for $\hypclass_{\text{hs}}$.
\end{theorem}

\begin{remark}[Mixed labels]
Our tester at resolution $t=n$ also applies when each agent holds
both positive and negative examples. We retain the assumption
that $\dataspace\subseteq\Reals^d\setminus\{0\}$, and write $N_-$
for the total number of negatively labeled records across the
two datasets. Unlike the disjoint-label setting, this number
need not equal $n$. We first observe that the all-positive classifier makes exactly
$N_-$ mistakes. Thus, in the Far case, we must have $N_-\ge n$.
This suggests a simple first step: each agent releases a noisy
count of its negative records, and the agents add these counts.
If $N_-<n/2$, this already distinguishes the input from Far,
using the threshold $3n/4$. It remains to handle the case in
which there are at least $n/2$ negative records.

For this case, we modify the random signs used by the tester.
Independently for each line through the origin meeting
$\dataspace$, we assign a random sign to one ray and the
opposite sign to the other. Let $\sigma_R$ denote the sign
assigned to ray $R$. Each agent $P$ releases a noisy version of
\[
  Z_P=\sum_{(x,y)\in S_P}y\,\sigma_{R(x)},
\]
and the agents add their releases. Thus, the contribution of a
record depends on both its ray and its label.

The main observation is that these contributions cancel in
the Far case. Indeed, when $\opt=n$, every normal orthogonal
to no domain point gives exactly $n$ errors. Moving the normal
across the hyperplane perpendicular to one line shows that the
positive-label count minus the negative-label count must agree
on its two rays. Since these rays receive opposite signs,
their contributions sum to zero.

In the Match case, consider a line containing a negative record.
A classifier realizing the pooled data must place all negative
records on this line on one ray and all positive records on
the opposite ray. Consequently, multiplying by the labels makes
all contributions from this line have the same sign. Since the
signs on different lines are independent, the pooled sketch has
variance at least $N_-\ge n/2$. The same moment argument as
before therefore gives a sketch of magnitude at least
$\sqrt n/2$ with constant probability.

We repeat the sketch a constant number of times. Replacing one
labeled record changes each sketch by at most two and the
negative-label count by at most one, so Laplace noise of scale
$O(1/\varepsilon)$ suffices for privacy. Combining the count
test with the sketches gives an $\varepsilon$-private tester
whenever $n\ge C/\varepsilon^2$, for $0<\varepsilon\le1$ and a
universal constant $C$, independently of $d$ and $m$.
\end{remark}

\paragraph{A tester using private synthetic data generation.}
A simple application of the mechanism in
\citep{blum2013learning} gives a complementary bound with a better
dependence on the dimension. The idea is to have each party privately
release a small synthetic dataset that approximates its empirical
error for every halfspace. Then, using it, we can approximate $\opt$.

\begin{theorem}[Halfspace testing via the BLR net mechanism]
\label{thm:halfspace-query-release}
There exists a universal constant $C>0$ such that the following holds.
Fix $d\geq1$, $m,n\in\Naturals$, and $\epsilon\in(0,1]$.
For every public data space $\dataspace\subseteq\Reals^d$ with
$|\dataspace|=m$ and every $t\in[n]$ satisfying
\begin{equation}
\label{eq:halfspace-blr-gap}
t\geq C\left(
\frac{n^2d\log(2m)}{\epsilon}
\right)^{1/3},
\end{equation}
there exists a non-interactive, randomized,
$\epsilon$-two-party DP tester for $\hypclass_{\mathrm{hs}}$
with resolution $t$.
\end{theorem}

\begin{remark}
At first glance, the dependence on $d$, $\log m$, and $1/\epsilon$ in \cref{thm:halfspace-query-release} scales as $\bigl(\frac{d\log(2m)}{\epsilon}\bigr)^{1/3}$, which might appear to contradict our lower bound of $t=\Omega\bigl(\frac{d\log(m/d)}{\epsilon}\bigr)$ from \cref{thm:hs-lower-bound}. However, there is no contradiction because the bound in \cref{eq:halfspace-blr-gap} also depends on $n$ and can be written as a weighted geometric mean of $n$ and $\frac{d\log(2m)}{\epsilon}$:
\[
\left(\frac{n^2d\log(2m)}{\epsilon}\right)^{1/3}
=
n^{2/3}\left(\frac{d\log(2m)}{\epsilon}\right)^{1/3}.
\]
Recall that any non-trivial tester requires $t\leq n$ (since $\opt_{\hypclass_{\mathrm{hs}}}(S_A,S_B)\leq n$ holds trivially for all datasets of size $n$). For the right-hand side of \cref{eq:halfspace-blr-gap} to be at most $n$, we must have $n \gtrsim \frac{d\log(2m)}{\epsilon}$. In this regime, the geometric mean always lies between $\frac{d\log(2m)}{\epsilon}$ and $n$, namely
\[
\frac{d\log(2m)}{\epsilon}
\lesssim
\left(\frac{n^2d\log(2m)}{\epsilon}\right)^{1/3}
\leq n.
\]
Conversely, when $n \ll \frac{d\log(2m)}{\epsilon}$, the right-hand side of \cref{eq:halfspace-blr-gap} exceeds $n$, so the range of admissible $t\in[n]$ in \cref{thm:halfspace-query-release} is empty. Thus, whenever \cref{thm:halfspace-query-release} is non-vacuous, its resolution $t$ is always consistent with \cref{thm:hs-lower-bound}.
\end{remark}

\section*{Acknowledgment}
Mahdi Haghifam would like to thank Alireza Fallah and Adam Smith for discussions on this project.

\section*{AI Disclosure}
GPT-5 and GPT-6 were used to explore ideas for proving lower bound for halfspaces in \cref{thm:hs-lower-bound}. Authors also used AI tools for generating the figures in the paper as well as polishing, editing, and proof-reading. 

\printbibliography

\newpage
\appendix

\section{On the in-population error induced by OPT}\label{app:generalization}

The joint empirical loss $\opt_{\hypclass}$ can be directly linked to an in-distribution loss. Suppose the agents want to minimize the loss with respect to a target distribution $P\in\Delta(\dataspace\times\{\pm1\})$. We define the minimal population error on this distribution for a hypothesis class $\hypclass$ as follows: 
\begin{equation}
  R_{\hypclass}^{*}(P)
  :=\inf_{h\in\hypclass}\Pr_{(X,Y)\sim P}[h(X)\neq Y].
  \label{eq:joint-population-risk}
\end{equation}
Clearly, if the agents require target population error at most $\tau\in(0,1)$, then $R_{\hypclass}^{*}(P)>\tau$ implies that no shared classifier in $\hypclass$, including the Empirical Risk Minimizer (ERM) on their joint data, can satisfy them. In the following remark, we highlight how distinguishing between the conditions on $\opt_\hypclass$ in \eqref{eq:hypotheses} allows us to  detect a large $R_{\hypclass}^{*}(P)$, and vice-versa under distributional assumptions on $S_A,\ S_B$.

\begin{remark}[Independently sampled training datasets]\label{rem:iid_samples}
Suppose each agent's dataset consists of $n$ i.i.d.\ samples from its population distribution $P_A,\ P_B\in\Delta(\dataspace\times\{\pm1\})$, respectively, and the two datasets are independent. Define the minimal average population error by $\widebar R^*_{\mathcal{H}} := R^*_{\mathcal{H}}(\tfrac{P_A + P_B}{2})$. Let $v:=\operatorname{VC}(\mathcal{H})$ and assume $1\le v\le n$. Applying relative VC uniform convergence \citep[Thm.~7]{bousquet2003introduction} separately to the two datasets gives, with probability at least $1-\delta$,
\[
\left|\frac{\opt_\hypclass(S_A,S_B)}{2n}
-\widebar R^*_{\mathcal{H}}\right|
\le
\sqrt{\eta_n\min\left\{
\frac{\opt_\hypclass(S_A,S_B)}{2n},
\widebar R^*_{\mathcal{H}}
\right\}}+\eta_n,
\]
where $\eta_n:=C(v\log(en/v)+\log(1/\delta))/n$ for a sufficiently large universal constant $C$. Thus, the joint empirical loss bounds the minimal average population error from both above and below. In particular, for any $\tau\in(0,1)$, on this event, $\opt_\hypclass(S_A,S_B)>2n(\tau+\sqrt{\tau\eta_n}+\eta_n)$ implies $\widebar R^*_{\mathcal{H}}>\tau$, ruling out any shared classifier in $\mathcal{H}$ with average population error at most $\tau$ across $P_A$ and $P_B$. Conversely, $\opt_\hypclass(S_A,S_B)\le 2n(\tau-\sqrt{\tau\eta_n}-\eta_n)$ implies $\widebar R^*_{\mathcal{H}}\le\tau$. In particular, zero joint empirical loss implies $\widebar R^*_{\mathcal{H}}\le\eta_n$ on the same event. Thus, under independent sampling, the joint empirical loss $\opt_\hypclass(S_A,S_B)$ characterizes the minimal average population error $\widebar R^*_{\mathcal{H}}$ and importantly, when the target distribution satisfies $P=\tfrac{P_A+P_B}{2}$, we have $\widebar R^*_{\mathcal{H}}=R^*_{\mathcal{H}}(P)$, so this remark directly relates the joint empirical loss to the minimal target population error.
\end{remark}

We go back to our running examples and apply \Cref{rem:iid_samples} to demonstrate how, in real applications,  a large joint empirical loss $\opt_{\hypclass}(S_A,S_B)$ certifies that $R_{\hypclass}^{*}>\tau$ with high probability.

\begin{myexample}[Learning from disparate label sets]
Remember that Alice's labels are all $-1$, and Bob's are all $+1$, but their data otherwise follow the target distribution conditioned on the label. Formally, let $P_A:=P_{\cdot\mid y=-1}$ and $P_B:=P_{\cdot\mid y=1}$, and suppose $S_A\sim P_A^n$ and $S_B\sim P_B^n$ independently. If both labels have equal probability under $P$, then $\tfrac12(P_A+P_B)=P$. By \Cref{rem:iid_samples}, with probability at least $1-\delta$, $\operatorname{opt}_{\mathcal{H}}(S_A,S_B)=0$ implies $R^*_{\mathcal{H}}(P)\le\eta_n$, whereas $\operatorname{opt}_{\mathcal{H}}(S_A,S_B)>t$ implies $R^*_{\mathcal{H}}(P)>t/(2n)-\eta_n$. Thus, the joint empirical loss allows the agents to conclude whether a shared classifier in $\mathcal{H}$ can achieve low target error. 
\end{myexample}

\begin{myexample}[Identifying conflicting labeling practices]
Remember that the agents encounter the same distribution of emails but may assign different labels to the same email. Let $D$ denote their common feature distribution and $f_A,f_B\in\hypclass$ their deterministic labeling functions, so each task is individually realizable. Let $P_A$ and $P_B$ be the distributions of $(X,f_A(X))$ and $(X,f_B(X))$, respectively, for $X\sim D$. We take the target distribution to be $P=\tfrac12(P_A+P_B)$, a symmetric choice that gives the two agents' test errors equal weight. Suppose $S_A\sim P_A^n$ and $S_B\sim P_B^n$ independently. By \Cref{rem:iid_samples}, with probability at least $1-\delta$, $\opt_\hypclass(S_A,S_B)=0$ implies that a shared classifier in $\hypclass$ can achieve test error at most $\eta_n$ under $P$, whereas $\opt_\hypclass(S_A,S_B)>2n(\tau+\eta_n)$ rules out any shared classifier in $\hypclass$ with test error at most $\tau$ under $P$.
\end{myexample}

\section{Helper Lemma}

\begin{lemma}[Laplace mechanism~\citep{dwork2006calibrating}]
\label{lem:laplace}
Let $f: \dataspace^n \to \Reals^d$ be a vector-valued function with $\ell_1$-sensitivity at most
$\Delta$, meaning $\|f(X)-f(X')\|_1\le \Delta$ for all neighboring $X,X'$, i.e., two $X,X' \in \dataspace^n$ that differ only in a single element.  The mechanism $\tilde{f}(X) = f(X)+Z$, where $Z \sim \left(\Lap(\Delta/\varepsilon)\right)^{\otimes d}$, is $epsilon$-DP.  Here $\Lap(b)$ denotes the
mean-zero Laplace distribution with density $(2b)^{-1}e^{-|z|/b}$. Moreover, for every $X \in \dataspace^n$ and $\beta \in (0,1]$, we have $\Pr\left(|\tilde{f}(X)-f(X)| \geq \frac{\Delta}{\epsilon} \log(1/\beta) \right)\leq \beta$.
\end{lemma}

\begin{lemma}[A constant-degree expander \cite{Richardson_Urbanke_2008}]
\label{lem:hs-expander}
There are universal constants $\delta>0$ and $K_0\in\Naturals$ such
that, for every even $K\ge K_0$, there is a simple bipartite graph with
$K$ variable nodes of degree four and $K/2$ check nodes of degree eight
satisfying
\[
  |\Gamma(T)|>2|T|
  \qquad\text{whenever }\varnothing\ne T\subseteq[K]
  \text{ and }|T|<\delta K.
\]
Here $\Gamma(T)$ is the set of checks adjacent to $T$. Consequently,
every such $T$ has a check adjacent to exactly one of its variables.
\end{lemma}

\begin{lemma}[Carath\'eodory's theorem \citep{caratheodory1907variability}]\label{lem:carathedory}
Let $\dataspace\subseteq\Reals^d$. Every point $x\in\conv(\dataspace)$ can be written as
a convex combination of at most $d+1$ points of $\dataspace$. That is for every $x\in\conv(\dataspace)$, there
exist $n\leq d+1$, points $x_1,\ldots,x_n\in \dataspace$, and coefficients
$\lambda_1,\ldots,\lambda_n\geq0$ with
$\sum_{i=1}^n\lambda_i=1$ such that $x=\sum_{i=1}^n\lambda_i x_i.$
\end{lemma}

\begin{lemma}[Exponential mechanism \citep{mcsherry2007mechanism}]
\label{lem:exponential-mechanism}
Let $\mathcal{R}$ be a finite output space, and let
$q:\dataspace^\star\times\mathcal{R}\to\Reals$ be a cost function with
sensitivity at most $\Delta>0$; that is, for every pair of neighboring
datasets $S,S'$ and every $r\in\mathcal{R}$,
$|q(S,r)-q(S',r)|\leq\Delta$. The exponential mechanism outputs $r\in\mathcal{R}$ with probability
$
\Pr[\mathsf{EM}(S)=r]
=
\frac{\exp\!\left(-\frac{\epsilon q(S,r)}{2\Delta}\right)}
{\sum_{r'\in\mathcal{R}}
 \exp\!\left(-\frac{\epsilon q(S,r')}{2\Delta}\right)}.
$ Then $\mathsf{EM}$ is $\epsilon$-DP.
Moreover, for every $\beta\in(0,1)$, with probability at least $1-\beta$,
\[
q\!\left(S,\mathcal{M}_{\mathrm{exp}}(S)\right)
\leq
\min_{r\in\mathcal{R}}q(S,r)
+
\frac{2\Delta}{\epsilon}
\left(
\ln|\mathcal{R}|+\ln\frac{1}{\beta}
\right).
\]
\end{lemma}

\begin{lemma}[From LP constraints to halfspace datasets]
\label{lem:hs-lp-conversion}
Fix $p,M \in \Naturals$. Let $\mathcal{R} = \{a_i^\top x \le b_i\}_{i\in [M]}$ be a finite set of linear constraints on $x\in\Reals^p$. There exists a constant $\eta > 0$, depending only on $\mathcal{R}$, with the following property. For each $i\in[M]$, define the points
\[
  x_i^- := (a_i, -b_i - \eta)\in\Reals^{p+1}
  \qquad\text{and}\qquad
  x_i^+ := (-a_i, b_i)\in\Reals^{p+1},
\]
together with $x_0^- := -e_{p+1}$ and $x_0^+ := e_{p+1}$ in $\Reals^{p+1}$.
Then for any subsets $I_A, I_B \subseteq [M]$, if we form the datasets
\[
  S_A := \{x_i^- : i\in I_A\}\cup\{x_0^-\}
  \qquad\text{and}\qquad
  S_B := \{x_i^+ : i\in I_B\}\cup\{x_0^+\},
\]
the linear system $\{a_i^\top x \le b_i : i\in I_A\cup I_B\}$ is feasible in $\Reals^p$ if and only if $\opt_{\hypclass_{\mathrm{hs}}}(S_A, S_B) = 0$ (that is, there exists $h_w\in\hypclass_{\mathrm{hs}}$ such that $h_w(x)=-1$ for all $x\in S_A$ and $h_w(x)=+1$ for all $x\in S_B$).
\end{lemma}
\begin{proof}
We first choose $\eta>0$ depending only on $\mathcal{R}$. For each subset $T\subseteq[M]$ such that $\{a_i^\top x\le b_i : i\in T\}$ is infeasible, Farkas' lemma gives a vector $y_T\in\Reals_{\ge 0}^T$ with $\sum_{i\in T}(y_T)_i = 1$, $\sum_{i\in T}(y_T)_i a_i = 0$, and $\gamma_T := -\sum_{i\in T}(y_T)_i b_i > 0$. Since $[M]$ has only finitely many subsets, we may choose $\eta > 0$ smaller than $\min\{\gamma_T : T\subseteq[M]\text{ is infeasible}\}$ (or $\eta=1$ if every subset is feasible).
Now fix any $I_A, I_B\subseteq[M]$ and let $S := I_A\cup I_B$.
Suppose first that there exists $x\in\Reals^p$ satisfying $a_i^\top x\le b_i$ for all $i\in S$, and consider $h_w\in\hypclass_{\mathrm{hs}}$ with $w=(x,1)\in\Reals^{p+1}$. Then $\inner{w}{x_0^-}=-1<0$, $\inner{w}{x_0^+}=1\ge 0$, and for every $i\in I_A$ and $j\in I_B$,
\[
  \inner{w}{x_i^-} = a_i^\top x - b_i - \eta \le -\eta < 0,
  \qquad
  \inner{w}{x_j^+} = -a_j^\top x + b_j \ge 0.
\]
Hence $h_w(x)=-1$ for all $x\in S_A$ and $h_w(x)=+1$ for all $x\in S_B$, so $\opt_{\hypclass_{\mathrm{hs}}}(S_A, S_B)=0$.
Conversely, suppose $\opt_{\hypclass_{\mathrm{hs}}}(S_A, S_B)=0$, and let $w=(v,\lambda)\in\Reals^p\times\Reals$ be such that $h_w(x)=-1$ for all $x\in S_A$ and $h_w(x)=+1$ for all $x\in S_B$. Since $x_0^-\in S_A$, we have $\inner{w}{x_0^-}=-\lambda<0$, so $\lambda>0$. Setting $\tilde{x}:=v/\lambda\in\Reals^p$, the conditions $\inner{w}{x_i^-}<0$ for $i\in I_A$ and $\inner{w}{x_j^+}\ge 0$ for $j\in I_B$ divide by $\lambda>0$ to yield
\[
  a_i^\top\tilde{x} < b_i + \eta \quad (\forall i\in I_A)
  \qquad\text{and}\qquad
  a_j^\top\tilde{x} \le b_j \quad (\forall j\in I_B).
\]
In particular, $a_i^\top\tilde{x} \le b_i + \eta$ for all $i\in S$. By our choice of $\eta$, the system $\{a_i^\top x\le b_i : i\in S\}$ cannot be infeasible, since otherwise
\[
  0 = \sum_{i\in S}(y_S)_i a_i^\top\tilde{x} \le \sum_{i\in S}(y_S)_i(b_i+\eta) = -\gamma_S + \eta < 0,
\]
a contradiction. Thus $\{a_i^\top x\le b_i : i\in I_A\cup I_B\}$ is feasible in $\Reals^p$.
\end{proof}

\begin{lemma}[Information complexity of the trace inner product] 
\label{lem:ic-trace}
Let $q=2^s$, and let $\mu$ be the uniform product distribution over
$\mathbb{F}_q^k\times\mathbb{F}_q^k$. For $F(u,v)
=
\bigoplus_{i=1}^k
\operatorname{Tr}_{\mathbb{F}_q/\mathbb{F}_2}(u_iv_i),
$, we have
$
\operatorname{IC}_{\mu,1/10}(F)
=
\Omega(k\log q).
$
\end{lemma}

\begin{proof}
The trace pairing $(a,b)\mapsto
\operatorname{Tr}_{\mathbb{F}_q/\mathbb{F}_2}(ab)$ is nondegenerate.
Therefore, there exist dual $\mathbb{F}_2$-bases
$\{\alpha_1,\ldots,\alpha_s\}$ and $\{\beta_1,\ldots,\beta_s\}$ of
$\mathbb{F}_q$ satisfying
\[
\operatorname{Tr}_{\mathbb{F}_q/\mathbb{F}_2}(\alpha_r\beta_\ell)
=
\indic{r=\ell}.
\]
Write
$
u_i=\sum_{r=1}^s x_{i,r}\alpha_r,
v_i=\sum_{r=1}^s y_{i,r}\beta_r$ where ,
$x_{i,r},y_{i,r}\in\mathbb{F}_2$.
Then
\[
\operatorname{Tr}_{\mathbb{F}_q/\mathbb{F}_2}(u_iv_i)
=
\bigoplus_{r=1}^s x_{i,r}y_{i,r},
\]
and consequently
$
F(u,v)
=
\bigoplus_{i=1}^k\bigoplus_{r=1}^s x_{i,r}y_{i,r}
=
\operatorname{IP}_{ks}(x,y).
$

The changes of basis are bijections, so under $\mu$, the corresponding
$x,y\in\{0,1\}^{ks}$ are independent and uniform. Moreover, any protocol
for $F$ induces a protocol for $\operatorname{IP}_{ks}$ with the same
error and the same internal information cost. It is folklore that binary
inner product on $N$ independent uniform bits has constant-error internal
information complexity $\Omega(N)$. Therefore, $\operatorname{IC}_{\mu,1/10}(F) =\Omega(ks)=\Omega(k\log q)$.
\end{proof}

\section{Embedding a codeword into an LP}
\label{appx:hs-lb-lp-construction}

In this section, we develop the machinery that turns a pair of codewords
into an instance of the promise problem. Recall from
\cref{sec:hs-lower-overview} that each data point imposes one linear
constraint on $w$, so that $\opt_{\hypclass_{\hs}}=0$ means the resulting
system is feasible, while $\opt_{\hypclass_{\hs}}\ge t$ means it remains
infeasible even after removing any $t-1$ constraints. Our goal is
therefore a linear program whose feasibility encodes the value of a hard
function, and in which each private \textcolor{black}{symbol} contributes exactly one
inequality.

Our representation is based on the fundamental-polytope approach from the literature on LP decoding of linear codes \citep{feldman2005using,flanagan2009linear}. In that approach, each row of a parity-check matrix is represented by the convex hull of its satisfying local assignments, and the resulting local polytopes are intersected to obtain an LP relaxation of the code. We adapt this machinery in two ways. First, we represent every symbol of \(\mathbb F_q\) as an exposed vertex of a two-dimensional polygon, rather than using the standard indicator representation. Second, we interpret the symbol-selecting inequalities as private data records and the parity-check inequalities as public records.

\paragraph{The function we want to compute.}
\textcolor{black}{Fix the public parity-check matrix
$H\in\Field_q^{M\times K}$ with $M=K/2$, retaining all its rows,
and let $\code_H=\{c\in\Field_q^K:Hc^\top=0\}$ have dimension $k$.}
We assume that the code is
systematic, i.e.,  the first $k$ coordinates carry the message and
the remaining $K-k$ are determined by the checks. Alice holds a codeword
$c\in\code_H$, Bob holds a codeword $c'\in\code_H$, and they wish to
compute
\[
F(c,c')=\bigoplus_{i\in[k]}\Tr(c_ic_i')\in\{0,1\},
\]
where $\Tr$ denotes the trace from $\Field_q$ to $\Field_2$. The
computation has three parts: each party's input must be a genuine
codeword, that is, satisfy every parity check of $H$; for every message
coordinate $i\in[k]$, the bit $\Tr(c_ic_i')$ must be computed from the
pair $(c_i,c_i')$; and these $k$ bits must be combined by parity. We then
plant a public condition on the output, namely that the resulting parity
equals zero. Feasibility of the whole system is then equivalent to
$F(c,c')=0$.

\begin{figure}
\centering
\includegraphics[width=0.5\textwidth]{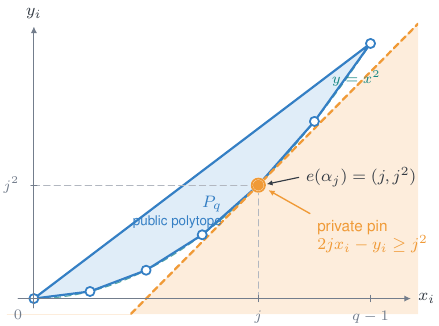}
\caption{Encdoing of the field's element into $\Reals$}
\label{fig:probala}
\end{figure}

\paragraph{Encoding a field element.}
Fix an enumeration $\Field_q=\{\alpha_0,\ldots,\alpha_{q-1}\}$ and set
$e(\alpha_j):=(j,j^2)\in\Reals^2$. To represent a field-valued variable
we introduce a pair of real variables $z_i=(z_i^{(1)},z_i^{(2)})$ and
impose the public constraint $z_i\in P_q$, where $P_q$ is the convex hull
of the $q$ points $e(\alpha_0),\ldots,e(\alpha_{q-1})$. Since these
points lie on a parabola, they are in convex position, and $P_q$ is
described by
\[
0\le z_i^{(1)}\le q-1,
\qquad
z_i^{(2)}\le(q-1)\,z_i^{(1)},
\qquad
z_i^{(2)}\ge(2j+1)\,z_i^{(1)}-j(j+1)
\ \text{ for } j=0,\ldots,q-2.
\]
See
\cref{fig:probala}. This constraint says only that $z_i$ is \emph{some}
mixture of field elements. The private input supplies one further
inequality that selects a particular vertex.

\begin{lemma}[Pinning a field element]
\label{lem:pin}
Let $\ell_j(u,v):=2ju-v$. Then for every $s\in\{0,\ldots,q-1\}$,
\[
\ell_j\bigl(e(\alpha_s)\bigr)=2js-s^2=j^2-(s-j)^2\le j^2,
\]
with equality if and only if $s=j$. Consequently, the constraints
$z_i\in P_q$ and $\ell_j(z_i)\ge j^2$ together force $z_i=e(\alpha_j)$.
\end{lemma}

\begin{proof}
The displayed identity is immediate. Since $z_i\in P_q$, we may write
$z_i=\sum_s\lambda_se(\alpha_s)$ with $\lambda_s\ge0$ and
$\sum_s\lambda_s=1$. As $\ell_j$ is linear,
\[
\ell_j(z_i)=\sum_s\lambda_s\,\ell_j\bigl(e(\alpha_s)\bigr)
=j^2-\sum_s\lambda_s(s-j)^2 .
\]
The constraint $\ell_j(z_i)\ge j^2$ therefore forces
$\sum_s\lambda_s(s-j)^2\le0$. Every term is nonnegative, so
$\lambda_s=0$ for all $s\ne j$, whence $\lambda_j=1$ and
$z_i=e(\alpha_j)$.
\end{proof}

We call $\ell_j(z_i)\ge j^2$ the \emph{pin} for $\alpha_j$; see
\cref{fig:probala}. The inequalities describing $P_q$ are public and
identical for every input, and only the choice among the $q$ candidate
pins depends on the data. Thus one private inequality specifies one of
$q$ field elements, which is exactly the accounting we need.

\paragraph{Imposing a public relation.}
The pins determine each symbol individually, but we also need to impose
relations among several variables, such as a parity check or the trace of
a product. Such relations are not linear, so we cannot write them
directly. The following device converts any finite relation into linear
inequalities.

Let $\mathcal{R}$ be a public finite set of tuples
$(a_1,\ldots,a_r,b_1,\ldots,b_{r'})\in\Field_q^r\times\{0,1\}^{r'}$,
listing the assignments we wish to allow. Replace each field symbol $a_i$
by the point $e(a_i)$, we describe the construction above, and leave the bits unchanged. Then, let
$Q_{\mathcal{R}}$ be the convex hull of the resulting points in
$\Reals^{2r+r'}$. These points are rational, so $Q_{\mathcal{R}}$ is a
rational polytope and admits a finite description by linear inequalities,
which we add to the linear program as public rows.
\textcolor{black}{If the polytope is lower-dimensional, this description
includes equations for its affine hull, each written as two
inequalities. This also covers $q=2$.}
In the linear program
this reads
\[
(z_1,\ldots,z_r,\;\beta_1,\ldots,\beta_{r'})\in Q_{\mathcal{R}},
\]
where the $z_i$ are symbol variables and the $\beta_j$ are real variables
playing the role of bits. By itself this constraint is weak, as it says
only that the tuple is some mixture of allowed assignments. The point is
that pinning some of the variables collapses the mixture.

\begin{lemma}[Forced relation]
\label{lem:forced-relation}
\textcolor{black}{Fix nonnegative integers $r,r'$ with $r+r'>0$, and
use the convention $[0]=\varnothing$.}
Let $\mathcal{R}\subseteq\Field_q^{r}\times\{0,1\}^{r'}$ be a finite
relation and let $Q_{\mathcal{R}}$ be as above. Fix index sets
$S\subseteq[r]$ and $S'\subseteq[r']$, together with symbols
$a_i\in\Field_q$ for $i\in S$ and bits $b_j\in\{0,1\}$ for $j\in S'$, and
suppose that $\mathcal{R}$ contains exactly one tuple $\mathcal{T}^\star$
whose $i$-th symbol equals $a_i$ for every $i\in S$ and whose $j$th bit
equals $b_j$ for every $j\in S'$. If a point
$(z,\beta)\in Q_{\mathcal{R}}$ satisfies $z_i=e(a_i)$ for every $i\in S$
and $\beta_j=b_j$ for every $j\in S'$, then $(z,\beta)$ equals the point
generated by $\mathcal{T}^\star$; in particular every remaining
coordinate is forced to the value it takes in $\mathcal{T}^\star$.
\end{lemma}

\begin{proof}
We want to show that in any convex decomposition of $(z,\beta)$ over the
generating points, all the weight sits on the single generator coming
from $\mathcal{T}^\star$. 

Write $(z,\beta)=\sum_{\mathcal{T}\in\mathcal{R}}
\lambda_{\mathcal{T}}\,v_{\mathcal{T}}$ with $\lambda_{\mathcal{T}}\ge0$
and $\sum_{\mathcal{T}}\lambda_{\mathcal{T}}=1$, where $v_{\mathcal{T}}$
denotes the point generated by the tuple $\mathcal{T}$. Define
\[
L:=\sum_{i\in S}\ell_{a_i}+\sum_{j\in S'}\sigma_{b_j},
\qquad
\sigma_0(\beta):=-\beta,
\quad
\sigma_1(\beta):=\beta,
\]
where \textcolor{black}{$\ell_{\alpha_j}$ denotes $\ell_j$,}
$\ell_{a_i}$ acts on the $i$th symbol block and $\sigma_{b_j}$ on
the $j$th bit coordinate. By \cref{lem:pin}, $\ell_{a_i}$ is maximized
over $\{e(\alpha):\alpha\in\Field_q\}$ exactly at $e(a_i)$, and clearly
$\sigma_{b_j}$ is maximized over $\{0,1\}$ exactly at $b_j$. Writing $M$
for the resulting maximum, we therefore have $L(v_{\mathcal{T}})\le M$
for every $\mathcal{T}\in\mathcal{R}$, with equality if and only if
$\mathcal{T}$ agrees with the fixed assignment on $S\cup S'$. On the
other hand $L(z,\beta)=M$ by hypothesis, so linearity gives
\[
\sum_{\mathcal{T}}\lambda_{\mathcal{T}}
\bigl(M-L(v_{\mathcal{T}})\bigr)=0 .
\]
Every summand is nonnegative, so $\lambda_{\mathcal{T}}=0$ whenever
$L(v_{\mathcal{T}})<M$; that is, only tuples agreeing with the fixed
assignment receive positive weight. By hypothesis $\mathcal{T}^\star$ is
the only such tuple, so $\lambda_{\mathcal{T}^\star}=1$ and
$(z,\beta)=v_{\mathcal{T}^\star}$.
\end{proof}

\paragraph{The four families of constraints.}
We can now assemble the linear program. Every part of the computation
becomes an instance of the device above, and the private inputs enter
through the pins alone.

\underline{\emph{Symbols.}} For every $i\in[K]$ we introduce $z_i\in\Reals^2$ with
$z_i\in P_q$, and Alice's private input contributes the pin forcing
$z_i=e(c_i)$. Bob contributes an identical family for a second collection
$z_1',\ldots,z_K'$.

\underline{\emph{Parity checks.}} For every row $a$ of $H$, we take $\mathcal{R}$ to
be the set of assignments to the coordinates in the support of $a$ that
satisfy the check, and impose membership in the corresponding polytope.
Applying this to both collections guarantees that pinned inputs are
codewords.

\underline{\emph{Trace products.}} For every $i\in[k]$ we introduce a real variable
$r_i$ and take $\mathcal{R}=\{(a,b,\Tr(ab)):a,b\in\Field_q\}$. Once $z_i$
and $z_i'$ are pinned, \cref{lem:forced-relation} forces
$r_i=\Tr(c_ic_i')\in\{0,1\}$.

\underline{\emph{Parity of the bits.}} We introduce $p_0,\ldots,p_k$, impose the
public row $p_0=0$, and for every $j\in[k]$ take $\mathcal{R}$ to be the
four even-parity binary triples, so that the constraint reads
$(p_{j-1},r_j,p_j)\in Q_\oplus$. Once $p_{j-1}$ and $r_j$ are bits,
\cref{lem:forced-relation} forces $p_j=p_{j-1}\oplus r_j$.

Finally we plant the public row $p_k=0$. Notice that these families are
not independent: the pins force the symbols, the symbols force the trace
bits, and the trace bits together with the running parities force the
next parity. Each gadget produces integral values only because the
previous one already has, so the constraints must be read as a chain
rather than as a set.

\paragraph{Size of the embedding.}

\color{black}
\begin{lemma}[Dimension and number of rows]
\label{lem:embedding-size}
The linear program above uses $4K+2k+1$ real variables. If $H$ has
$M$ rows of maximum support size $\rho$, the public rows together
with all candidate private pins admit a rational description using
at most
\[
  O(Kq)+M q^{O(\rho^2)}+O(kq^4+k)
\]
inequalities. For the chosen $(4,8)$-biregular graph, $M=K/2$ and
$\rho=8$, so both the public-row count and the full catalogue size
are at most $BKq^a$ for universal $B,a$.
\end{lemma}

\begin{proof}
The symbol blocks contribute $4K$ variables, the trace bits contribute
$k$, and the running parities contribute $k+1$.

For the rows, use a rational facet description in each polytope's
affine hull, together with equations defining that affine hull.
The Upper Bound Theorem \citep{mcmullen1970maximum} bounds the number
of facets of a polytope with $N$ vertices and intrinsic dimension
$h\ge1$ by $O_h(N^{\lfloor h/2\rfloor})$. In ambient dimension
$D$, at most $D$ affine-hull equations add at most $2D$ inequalities;
a zero-dimensional polytope needs only these equations. All local
ambient dimensions here are bounded by $\max\{2\rho,5\}$.

\emph{Symbol constraints and pins.} Each of the $2K$ symbol blocks
uses $O(q)$ public inequalities for $P_q$, including the affine-hull
equation when $q=2$. Each block has $q$ possible pins, although only
one pin per block is selected for a dataset. Thus the public symbol
rows and all $2Kq$ candidate pins use $O(Kq)$ catalogue rows.

\emph{Parity checks.} Fix a check with $r_a$ nonzero entries. Its valid
local assignments are the solutions of one linear equation in $r_a$
unknowns over $\Field_q$: any $r_a-1$ of the symbols may be chosen
freely, and the remaining one is then determined, since its coefficient
is nonzero. There are therefore $q^{r_a-1}$ valid assignments, so the
associated polytope has at most $q^{r_a-1}$ vertices. Each symbol is
represented by a point of $\Reals^2$, so this polytope lies in dimension
at most $2r_a$. Applying the Upper Bound Theorem with $N=q^{r_a-1}$ and
$D=2r_a$ bounds the facet count by
$O\!\left(q^{r_a(r_a-1)}\right)$; at most $4r_a$ inequalities
describe the affine hull. Sum over all $M$ checks for both words,
including redundant checks, to obtain $M q^{O(\rho^2)}$ rows.

\emph{Trace products.} The trace polytope is the convex hull of the $q^2$
points $\bigl(e(a),e(b),\Tr(ab)\bigr)$ for $a,b\in\Field_q$, which lie in
$\Reals^{5}$. The Upper Bound Theorem with $N=q^2$ and $D=5$ gives
$O(q^{4})=q^{O(1)}$ facets, and there is one such polytope for each of
the $k$ message coordinates. Including at most ten inequalities for
each affine hull, these contribute $O(kq^4)$ public rows.

\emph{Parity accumulation.} The polytope $Q_\oplus$ is the convex hull of
four points in $\Reals^{3}$ and therefore has a constant-size
description; there is one for each of the $k$ steps of the chain, and the
endpoint equations $p_0=0$ and $p_k=0$ contribute four inequalities. Together these
give $O(k)$ inequalities, which is absorbed into the last term.

With $\rho=8$, one may take $a=56$, increasing a universal
multiplicative constant to cover every family. All descriptions are
rational and fixed before the inputs are specified. The conversion
in \cref{lem:hs-lp-conversion} adds one homogenizing coordinate and
two anchor types; these are counted in
\cref{lem:hs-encoding-resource}. Repetitions affect the dataset
lengths, but not this count of domain points.
\end{proof}
\color{black}

\begin{remark}[Why the parabola]
\label{rem:parabola}
The embedding $e(\alpha_j)=(j,j^2)$ is chosen for three reasons, all
visible in \cref{fig:probala}. First, a parabola is strictly convex, so
no field element lies in the convex hull of the others; had some
$e(\alpha_j)$ fallen inside, no single inequality could isolate it, and
one record could not specify a symbol. Second, the tangent line at
$e(\alpha_j)$ touches the parabola only there and has integer
coefficients, so the pin $2j\,z_i^{(1)}-z_i^{(2)}\ge j^2$ is rational and
uniform in $j$; this matters because every inequality eventually becomes
a point of the public data space. Third, two dimensions is the fewest in
which $q\ge3$ points can be in convex position, so the embedding is
optimal in the resource that is scarce for us. The alphabet size
therefore costs inequalities, never coordinates. 
\end{remark}

\section{Proof of the lower bound for half spaces} \label{appx:pf-hs-lb}

\paragraph{Step~1: construction.} {\color{black}\hypersetup{linkcolor=black}Fix an even
$K\ge K_0$, where $K_0$ is the universal minimum in
\cref{lem:hs-expander}, increased so that $\delta K_0\ge2$.
Let $q=2^s$ with $s\ge1$. For each such $K$, fix once and for all a
graph supplied by that lemma.}
\textcolor{black}{This is a} $(4,8)$-biregular graph on $K$ variable nodes and $M=K/2$ check
nodes satisfying, for a universal constant $\delta>0$,
\begin{equation}
\label{eq:expansion}
|\Gamma(T)|>2|T|
\qquad\text{for every } T\subseteq[K] \text{ with } 0<|T|<\delta K ,
\end{equation}
where $\Gamma(a)\subseteq[K]$ denotes the support of check $a$ and
$\Gamma(T)$ the set of checks meeting $T$. The same chosen graph is used for every alphabet and every
dataset size at this block length. Let
$H\in\{0,1\}^{M\times K}$ be its incidence matrix, viewed over
$\Field_q$, and define
\[
\code_H:=\{c\in\Field_q^K:Hc^\top=0\},
\qquad
k:=\dim_{\Field_q}(\code_H).
\]
Since $\operatorname{rank}(H)\le M=K/2$, we have $k\ge K/2$. After relabeling the
coordinates, fix a systematic generator
$G=[\id{k}\mid A]\in\Field_q^{k\times K}$.

Alice holds $u\in\Field_q^{k}$ and Bob holds $v\in\Field_q^{k}$. Set
$c=uG$ and $c'=vG$, and define
\begin{equation}
\label{eq:inner-prod-trace-appx}
F(u,v)=\bigoplus_{i=1}^{k}\Tr_{\Field_q/\Field_2}(u_iv_i).
\end{equation}
Since $G$ is systematic,
$F(u,v)=\bigoplus_{i=1}^{k}\Tr_{\Field_q/\Field_2}(c_ic_i')$, so the
trace products may be formed from the first $k$ coordinates of the
codewords. \textcolor{black}{We also write $F(c,c')$ for this value.}

By \cref{appx:hs-lb-lp-construction}, the pair $(c,c')$ determines a
rational linear program $\Lambda(c,c')$ with $4K+2k+O(1)$ variables and
at most $CKq^{a}$ possible distinct rows, for universal constants $C,a$.
Its only input-dependent rows are the $K$ pins determined by $c$ and the
$K$ pins determined by $c'$. We retain all $M$ parity checks,
including redundant ones, since the rigidity proof uses the graph
structure rather than a basis. Write $P_H$ for the set of
$z\in(\Reals^2)^{K}$ satisfying every public symbol and parity-check row,
so that $P_H\subseteq P_q^{K}$ and $e(c)\in P_H$ for every
$c\in\code_H$; note that $P_H$ is only a relaxation of $\code_H$ and may
contain further points.

\paragraph{Step~2: Correctness and the gap.} Let $\Lambda(c,c')$ denote the linear program consisting of all public rows
together with the $2K$ private pins \textcolor{black}{determined} by $c$ and $c'$.

\begin{lemma}[Feasibility decides $F$]
\label{lem:feasibility-decides-F}
For every $c,c'\in\code_H$, the system $\Lambda(c,c')$ is feasible if and
only if $F(c,c')=0$.
\end{lemma}

\begin{proof}
Suppose $F(c,c')=0$. Set $z_i=e(c_i)$ and $z_i'=e(c_i')$ for $i\in[K]$,
$r_i=\Tr(c_ic_i')$ for $i\in[k]$, and $p_j=\bigoplus_{i\le j}r_i$. Each
pin is satisfied by \cref{lem:pin}; each parity-check polytope is
satisfied because \textcolor{black}{$Hc^\top=H(c')^\top=0$}, so the corresponding local assignment is
valid and its embedding is one of the generators; the trace and parity
polytopes are satisfied for the same reason; and $p_k=F(c,c')=0$
satisfies the planted row.

Conversely, let $(z,z',r,p)$ be a solution of $\Lambda(c,c')$. By
\cref{lem:pin} the pins force $z_i=e(c_i)$ and $z_i'=e(c_i')$ for every
$i$. For each $i\in[k]$ the trace relation assigns a unique bit to the
pair $(c_i,c_i')$, so \cref{lem:forced-relation} forces
$r_i=\Tr(c_ic_i')$. Since $p_0=0$ and each $r_i$ is a bit, induction on
$j$ with \cref{lem:forced-relation} applied to $Q_\oplus$ forces
$p_j=\bigoplus_{i\le j}r_i$. In particular $p_k=F(c,c')$, and the planted
row gives $F(c,c')=0$.
\end{proof}

The feasibility equivalence by itself gives no quantitative guarantee
after pins are removed. The next lemma supplies that guarantee: the
codeword remains forced after a linear number of erasures, even inside
the fractional relaxation $P_H$.

\begin{lemma}[LP-erasure rigidity]
\label{lem:lp-rigidity}
Let $c\in\code_H$, let $S\subseteq[K]$ with $|S|<\delta K$, and let
$z\in P_H$ satisfy $z_i=e(c_i)$ for every $i\notin S$. Then $z=e(c)$.
\end{lemma}

\begin{proof}
Let $T:=\{i\in[K]:z_i\neq e(c_i)\}$ and suppose $T\neq\varnothing$. Since
$T\subseteq S$ we have $0<|T|<\delta K$, so \cref{eq:expansion} gives
$|\Gamma(T)|>2|T|$. Every coordinate lies in exactly four checks, so $T$
is incident to $4|T|$ edges; if every check in $\Gamma(T)$ met $T$ at
least twice, then $4|T|\ge2|\Gamma(T)|$, contradicting expansion. Hence
some check $a$ has $\Gamma(a)\cap T=\{i^\star\}$ for a single $i^\star$.

Every $j\in\Gamma(a)\setminus\{i^\star\}$ lies outside $T$, so
$z_j=e(c_j)$. As $H_{a,i^\star}\neq0$, the relation of check $a$ assigns
a unique symbol to coordinate $i^\star$ given the other seven, so
\cref{lem:forced-relation} forces $z_{i^\star}=e(c_{i^\star})$,
contradicting $i^\star\in T$. Hence $T=\varnothing$.
\end{proof}

\paragraph{Step~3: from the LP to fixed-size datasets.} We now convert the linear program $\Lambda(c,c')$ into size-$n$ datasets for Alice and Bob via \cref{lem:hs-lp-conversion}. Let $\mathcal{R}_{K,q}$ be the finite set of all linear constraints $a_i^\top x\le b_i$ appearing in the public polytopes (with each equality written as two inequalities), together with all $q$ candidate pins at each of Alice's and Bob's $K$ symbol positions. Applying \cref{lem:hs-lp-conversion} to $\mathcal{R}_{K,q}$ yields a constant $\eta_{K,q}>0$ and the corresponding points $x_i^-, x_i^+\in\Reals^{p+1}$, along with the anchors $x_0^-=-e_{p+1}$ and $x_0^+=e_{p+1}$. Let $\dataspace_{K,q}\subset\Reals^{p+1}$ denote the resulting finite set of points; note that $\dataspace_{K,q}$ depends only on $(K,q)$ and is independent of the inputs $(u,v)$ and the testing parameters.

Let $R_{\mathrm{pub}}$ be the number of public LP rows. For positive
integers $J,L$, Alice's encoder $E_A^{J,L}(u)$ consists of $L$ copies
of each of her $K$ selected pin records and $J$ copies of $x_0^-$.
Bob's encoder $E_B^{J,L}(v)$ consists of $L$ copies of each of his
$K$ selected pin records, $J$ copies of each public row record, and
$J$ copies of $x_0^+$. In particular, their lengths are
\begin{equation}
\label{eq:hs-exact-encoder-lengths}
  N_A=KL+J,\qquad N_B=KL+(R_{\mathrm{pub}}+1)J.
\end{equation}
For these intermediate samples, $\opt$ denotes the same pooled-error
sum, allowing unequal lengths; only the padded size-$n$ inputs below
are passed to the tester.
Put $D=\lfloor\delta K\rfloor\ge\delta K/2$.

\begin{lemma}[Gap, diameter, and encoding resources]
\label{lem:gap}
\label{lem:hs-encoding-resource}
There are universal constants $B,a\ge1$ such that, for every even
$K\ge K_0$ and every $q=2^s\ge2$, the preceding construction has a
fixed domain $\dataspace_{K,q}$ in dimension at most $6K+2$ with
$|\dataspace_{K,q}|\le BKq^a$ and $R_{\mathrm{pub}}+1\le BKq^a$.
For every $J,L\in\Naturals$ and every pair of secrets,
\[
 F(u,v)=0\ \Longrightarrow\ \opt_{\hypclass_{\hs}}=0,
 \qquad
 F(u,v)=1\ \Longrightarrow\ \opt_{\hypclass_{\hs}}\ge\min\{J,DL\}.
\]
Each party's encoder has length at most $BKq^aJ+KL$ and replacement
diameter at most $KL$. Whenever both lengths are at most $n$, the
encoders can be padded to exactly $n$ records per party with the same
promise and diameter guarantees. The domain depends only on $(K,q)$.
\end{lemma}

\begin{proof}
The LP has $p=4K+2k+1\le6K+1$ real variables.
\Cref{lem:embedding-size} bounds its public rows and all candidate
pins by $O(Kq^a)$ for a fixed universal $a$.
\Cref{lem:hs-lp-conversion} adds one coordinate and two anchor types.
Taking $B$ large enough proves the domain and row counts, and
\eqref{eq:hs-exact-encoder-lengths} gives the dataset lengths.

If $F(u,v)=0$, \cref{lem:feasibility-decides-F} gives an exact feasible
LP witness. Its homogenization classifies every record, including
the anchors, correctly by \cref{lem:hs-lp-conversion}.

Suppose $F(u,v)=1$ and a halfspace makes fewer than
$\min\{J,DL\}$ errors. No public row or anchor can be misclassified,
because each appears $J$ times. Let $S_A,S_B\subseteq[K]$ be the
positions of misclassified pin records. Each such record appears
$L$ times, so $|S_A|,|S_B|<D$. Keep every public row and precisely
the pins outside $S_A,S_B$. The halfspace is consistent with all
these records and the normalization anchor. By
\cref{lem:hs-lp-conversion}, the corresponding original closed LP
subsystem has an exact witness $(z^\star,z'^\star,r^\star,p^\star)$.
The retained original pins give
$z_i^\star=e(c_i)$ for $i\notin S_A$ and
$z_i'^\star=e(c_i')$ for $i\notin S_B$.
Both symbol vectors satisfy all original public rows, so
\cref{lem:lp-rigidity} forces $z^\star=e(c)$ and $z'^\star=e(c')$.
The trace and parity relations now force $p_k^\star=F(u,v)=1$,
contradicting the retained public condition $p_k^\star=0$.

For the diameter, fix an ordering with one block of $L$ positions
for each pin and fixed positions for the public rows and anchors.
Changing a party's secret replaces at most $KL$ records; all other
positions are identical. This bounds the replacement distance also
when the datasets are regarded as multisets.

Finally, append $n-N_A$ copies of $x_0^-$ to Alice and $n-N_B$ copies
of $x_0^+$ to Bob. These multiplicities depend only on the public
parameters. Every Match witness already satisfies the anchors,
while adding records cannot decrease the Far optimum. The fixed
padding changes neither replacement diameter. Group privacy applies
on the full size-$n$ input space, so the padded encoders can be used
directly in \cref{lem:metric-dp-to-information}.
\end{proof}

With $J=\lceil t\rceil$ and $L=\lceil J/D\rceil$, the Far gap is at
least $t$, the diameter is $O(t+K)$, and the sample cost is
$O(Kq^a(t+1)+t+K)$ per party.

\section{Sample-size-dependent lower bounds}
\label{appx:hs-sample-sizes}

The proof of \Cref{thm:hs-lower-bound} gives a lower bound of
$\Omega(d\log(m/d)/\varepsilon)$ whenever
$n\gtrsim md/\varepsilon$. To obtain bounds for smaller datasets,
we return to the encoders in \cref{appx:pf-hs-lb}. Recall that Bob
repeats each of its $O(Kq^a)$ public records $t$ times, where $a$
is a universal constant. These repetitions contribute
$O(Kq^at)$ records, so the choice of block length and alphabet
must also account for the dataset size. For smaller datasets,
we shorten the block length or reduce the alphabet. This makes
the encoders shorter, at the price of a weaker lower bound of
order $K\log q/\varepsilon$.

We call a pair $(r,q)$ a \emph{scale}, where $r$ is an even block
length and $q$ is a power of two. At scale $(r,q)$, we prove that
\[
  t\gtrsim\frac{r\log q}{\varepsilon}
  \qquad\text{provided that}\qquad
  n\varepsilon\gtrsim r^2q^a\log q.
\]
Indeed, when $t$ is below the claimed lower bound, the encoders
use $O(r^2q^a\log q/\varepsilon)$ records, which fit into datasets
of size $n$ under this condition. For $n\varepsilon\le d^2$,
with $n\varepsilon$ above a universal constant, we use the binary
alphabet and choose $r=\Theta(\sqrt{n\varepsilon})$. This gives
$t\gtrsim\sqrt{n/\varepsilon}$. For $n\varepsilon>d^2$, we use
$r=\Theta(d)$ and choose the largest alphabet allowed by the
sample and domain sizes. This gives
\[
  t\gtrsim\frac d\varepsilon
  \log\!\left(
    2+\min\left\{\frac md,\frac{n\varepsilon}{d^2}\right\}
  \right).
\]
In particular, the bound reaches
$\Omega(d\log(m/d)/\varepsilon)$ once
$n\varepsilon\gtrsim md$.

We also want the bounds for different sample sizes to hold on
the same public domain. We therefore choose a domain
$\dataspace_{d,m}$ containing all the scales needed for the two
regimes, before $n,t,\varepsilon$ are specified. Each scale
$(r,q)$ uses $O(rq^a)$ points. By taking geometrically spaced
block lengths and alphabets, we can include these scales while
keeping the total number of points at most $m$.
All logarithms in this section are to base two.

\begin{lemma}[Lower bound at a single scale]
\label{lem:hs-scale}
There are universal constants $A,B,a\ge1$, $b>0$,
$\varepsilon_0\in(0,1]$, and an even integer $r_0\ge2$ such
that the following holds. For every even $r\ge r_0$ and every
$q=2^s$ with $s\in\Naturals$, there is a public domain
$\dataspace_{r,q}\subseteq\Reals^{6r+2}$ that depends only on
$(r,q)$ and has $|\dataspace_{r,q}|\le Brq^a$, such that for
every $0<\varepsilon\le\varepsilon_0$ and every
$n\in\Naturals$ satisfying
\begin{equation}
\label{eq:hs-scale-sample-condition}
  n\varepsilon\ge Ar^2q^a\log q,
\end{equation}
every $\varepsilon$-private tester for
$\hypclass_{\mathrm{hs}}$ over $\dataspace_{r,q}$ with
resolution $t$ at dataset size $n$ satisfies
\begin{equation}
\label{eq:hs-scale-lower-bound}
  t\ge b\,\frac{r\log q}{\varepsilon}.
\end{equation}
\end{lemma}

\begin{proof}
Let $K_0$ be the minimum block length used in
\cref{appx:pf-hs-lb}. Choose an even $r_0\ge K_0$ large
enough that $\delta r_0\ge2$, where $\delta$ is the expansion
constant of \cref{lem:hs-expander}. For every even $r\ge r_0$,
we use the domain $\dataspace_{K,q}$ of
\cref{lem:hs-encoding-resource} with $K=r$, padding its
coordinates with zeros if necessary. The bounds on its
dimension and size follow from that lemma, and the domain
depends only on $(r,q)$. Recall also that
$D=\lfloor\delta r\rfloor\ge\delta r/2$.

Let $0<\varepsilon\le\varepsilon_0$ and let $n$ satisfy
\eqref{eq:hs-scale-sample-condition}, where
$\varepsilon_0\le1$ will be fixed at the end of the proof.
Suppose toward a contradiction that there is an
$\varepsilon$-private tester with resolution
$t<b\,r\log q/\varepsilon$, where $b\le1$ will also be fixed
at the end. Since $t$ is an integer, we apply
\cref{lem:hs-encoding-resource} with $J=t$ and
$L=\lceil t/D\rceil$. These choices give $DL\ge t$.
Thus the encodings of two secrets $u,v$ satisfy
$\opt_{\hypclass_{\mathrm{hs}}}=0$ when $F(u,v)=0$, and
$\opt_{\hypclass_{\mathrm{hs}}}\ge t$ when $F(u,v)=1$.

First, we check that the encoders fit into $n$ records.
Each encoder has length at most $Brq^at+rL$, including
the public records and anchors. Since
$rL\le rt/D+r\le2t/\delta+r$, we obtain
\[
  Brq^at+rL
  \le
  \left(B+\frac2\delta+1\right)
  \frac{r^2q^a\log q}{\varepsilon}
  \le n.
\]
The first inequality uses $t<r\log q/\varepsilon$,
$r,q^a,\log q\ge1$, and $\varepsilon\le1$. The second follows
from \eqref{eq:hs-scale-sample-condition}, provided that
$A\ge B+2/\delta+1$. By \cref{lem:hs-encoding-resource},
we can therefore pad both encoders to exactly $n$ records
without changing their promise or encoding diameters.

Next, let $k\ge r/2$ be the dimension of the code, and let
$(u,v)$ be uniform on
$\mathbb{F}_q^k\times\mathbb{F}_q^k$. Running the tester on
the padded encodings, and reading $\mathsf{Match}$ as $0$
and $\mathsf{Far}$ as $1$, gives a protocol computing
$F(u,v)$ with error at most $1/10$. By \cref{lem:ic-trace},
its internal information cost is at least $c_Ir\log q$
for a universal $c_I>0$.

On the other hand, changing one party's secret replaces at
most $rL$ of its records. Hence
\begin{equation}
\label{eq:hs-scale-diameter}
  \Delta_A+\Delta_B
  \le2rL
  \le\frac{4t}{\delta}+2r
  \le C_\Delta(t+r)
\end{equation}
for a universal $C_\Delta$. Applying
\cref{lem:metric-dp-to-information} to the same protocol gives
\[
  c_Ir\log q\le C_I\varepsilon(t+r),
  \qquad
  C_I:=\frac{C_\Delta}{\ln2}.
\]
We now fix
\[
  \varepsilon_0:=\min\left\{1,\frac{c_I}{2C_I}\right\},
  \qquad
  b:=\min\left\{1,\frac{c_I}{4C_I}\right\}.
\]
Since $\varepsilon\le\varepsilon_0$ and $\log q\ge1$, the
term $C_I\varepsilon r$ is at most $c_Ir\log q/2$.
Rearranging therefore gives
\[
  t\ge\frac{c_I}{2C_I}\frac{r\log q}{\varepsilon}
  >b\,\frac{r\log q}{\varepsilon},
\]
contradicting the assumption on $t$.
\end{proof}

\begin{remark}[Larger domains]
\label{rem:hs-scale-larger-domain}
\Cref{lem:hs-scale} also applies to testers over any domain
$\dataspace\subseteq\Reals^d$, with $d\ge6r+2$, containing
a copy of $\dataspace_{r,q}$ padded with zeros. Indeed,
restricting a tester to datasets supported on this copy
preserves privacy and its completeness and soundness
guarantees. Padding with zeros also preserves
$\opt_{\hypclass_{\mathrm{hs}}}$: a normal $w\in\Reals^d$
makes the same predictions on the copy as its first $6r+2$
coordinates, and every normal in $\Reals^{6r+2}$ extends
to $\Reals^d$ by adding zero coordinates.
\end{remark}

We next construct a single domain containing the scales
needed for both regimes.

\begin{lemma}[A single domain for all scales]
\label{lem:hs-fixed-domain}
Let $A,B,a,r_0$ be the constants of \cref{lem:hs-scale}.
There are universal constants $\kappa>0$, $C_0\ge1$, and
$d_0\in\Naturals$ such that the following holds for every
$d\ge d_0$ and $m\ge C_0d$. There are an integer $h\ge0$
and a public domain $\dataspace_{d,m}\subseteq\Reals^d$
with $|\dataspace_{d,m}|=m$, depending only on $(d,m)$,
such that $R:=r_02^h$ satisfies
\[
  \frac{\kappa d}{2}<R\le\kappa d,
  \qquad
  6R+2\le d,
  \qquad
  A2^aR^2\le\frac{d^2}{16},
\]
and $\dataspace_{d,m}$ contains a copy of
$\dataspace_{r,q}$ padded with zeros for every scale
$(r,q)$ in
\begin{equation}
\label{eq:hs-fixed-scale-union}
  \bigl\{(r_02^j,2):0\le j\le h\bigr\}
  \ \cup\
  \bigl\{(R,2^s):s\ge1,\ 2^{as}\le m/(8BR)\bigr\}.
\end{equation}
\end{lemma}

\begin{proof}
Fix
\[
  \kappa:=\min\left\{
    \frac1{12},\frac1{4\sqrt{A2^a}}
  \right\},
  \qquad
  C_0:=8B2^a.
\]
Let $R$ be the largest element of
$\{r_02^j:j\ge0\}$ that is at most $\kappa d$.
Choose an integer $d_0\ge\max\{4,r_0/\kappa\}$, so that
$r_0\le\kappa d$ and this maximum exists. Maximality gives
$2R>\kappa d$. The remaining bounds on $R$ follow from
the choice of $\kappa$:
\[
  6R+2\le d/2+2\le d,
  \qquad
  A2^aR^2\le A2^a\kappa^2d^2\le d^2/16.
\]

For each scale in \eqref{eq:hs-fixed-scale-union}, fix one
construction of $\dataspace_{r,q}$, including its graph,
inequality descriptions, and the constant $\eta$ of
\cref{lem:hs-lp-conversion}. Pad each domain with zeros to
embed it in $\Reals^d$. All these embeddings fit because
$6r+2\le6R+2\le d$.

We count the points contributed by the two families
separately. The binary scales $(r_02^j,2)$ use at most
\[
  \sum_{j=0}^{h}B2^ar_02^j<2B2^aR
\]
points. Since $m\ge C_0d\ge8B2^aR$, this is less than
$m/4$.

For the alphabet scales $(R,2^s)$, let $Q$ be the largest
power of two satisfying $Q^a\le m/(8BR)$. The inequality
$m\ge8B2^aR$ ensures that $Q\ge2$ exists. These scales use
at most
\[
  BR\sum_{s=1}^{\log Q}2^{as}
  \le\frac{BRQ^a}{1-2^{-a}}
  \le2BRQ^a
  \le\frac m4
\]
points, where we used $a\ge1$.

The union therefore has fewer than $m/2$ points. We add
distinct unused points until the domain has exactly $m$
points. None of the encoded datasets uses these additional
points. All choices have been made using only $(d,m)$,
before the testing parameters are specified.
\end{proof}

We now choose a scale for each sample size.

\begin{theorem}[Lower bounds as a function of the sample size]
\label{thm:hs-sample-size}
There are universal constants $c>0$, $N_0\ge1$, $C_0\ge1$,
$d_0\in\Naturals$, and $\varepsilon_0>0$ such that the
following holds. Let $d\ge d_0$ and $m\ge C_0d$, and let
$\dataspace_{d,m}$ be the domain of
\cref{lem:hs-fixed-domain}. For every
$0<\varepsilon\le\varepsilon_0$ and every $n\in\Naturals$
with $n\varepsilon\ge N_0$, every $\varepsilon$-private
tester for $\hypclass_{\mathrm{hs}}$ over
$\dataspace_{d,m}$ with resolution $t$ at dataset size $n$
satisfies
\begin{equation}
\label{eq:hs-sample-size-lower-bounds}
  t\ge c
  \begin{cases}
    \displaystyle\sqrt{n/\varepsilon},
    &N_0\le n\varepsilon\le d^2,\\[6pt]
    \displaystyle\frac d\varepsilon
    \log\!\left(
      2+\min\left\{\frac md,\frac{n\varepsilon}{d^2}\right\}
    \right),
    &n\varepsilon>d^2.
  \end{cases}
\end{equation}
\end{theorem}

\begin{proof}
Use the constants and scales of
\cref{lem:hs-scale,lem:hs-fixed-domain}, and set
$N:=n\varepsilon$ and $N_0:=A2^ar_0^2$. Enlarge $d_0$ if necessary so that $d^2\ge N_0$.
Every scale used below belongs to
\eqref{eq:hs-fixed-scale-union}. Thus
\cref{lem:hs-scale} applies to testers over
$\dataspace_{d,m}$ by
\cref{rem:hs-scale-larger-domain}.

\paragraph{Case 1: $N_0\le N\le d^2$.}
We use the binary alphabet and choose $r$ to be the largest
element of $\{r_02^j:0\le j\le h\}$ satisfying
$A2^ar^2\le N$. Such an $r$ exists because
$A2^ar_0^2=N_0\le N$.

To obtain the claimed bound, we need
$r=\Omega(\sqrt N)$. If $r<R$, the next scale $2r$
violates the constraint, so
\[
  A2^a(2r)^2>N
  \qquad\Longrightarrow\qquad
  r>\frac{\sqrt N}{2\sqrt{A2^a}}.
\]
If $r=R$, then
$r>\kappa d/2\ge\kappa\sqrt N/2$, since $N\le d^2$.
Thus $r\ge c_r\sqrt N$ for
$c_r:=\min\{1/(2\sqrt{A2^a}),\kappa/2\}$.

For $q=2$, the inequality $N\ge A2^ar^2$ is precisely
the sample-size condition in \cref{lem:hs-scale}.
Applying that lemma gives
\[
  t\ge b\,\frac r\varepsilon
  \ge bc_r\,\frac{\sqrt{n\varepsilon}}{\varepsilon}
  =bc_r\sqrt{\frac n\varepsilon}.
\]

\paragraph{Case 2: $N>d^2$.}
We use the largest block length $R$ and choose the alphabet
to satisfy both the domain-size and sample-size constraints.
Define
\[
  Z:=\min\left\{\frac m{8BR},\frac N{AR^2}\right\},
  \qquad
  X:=\min\left\{\frac md,\frac N{d^2}\right\},
\]
and let $q=2^s$ be the largest power of two satisfying
\begin{equation}
\label{eq:hs-sample-alphabet-choice}
  q^a\log q\le Z.
\end{equation}
The choice $q=2$ is admissible. Indeed,
$m/(8BR)\ge2^a$ because $m\ge C_0d\ge8B2^aR$,
while
\[
  \frac N{AR^2}>16\cdot2^a
\]
follows from $A2^aR^2\le d^2/16<N/16$.

Since $\log q\ge1$, our choice gives
\[
  q^a\le q^a\log q\le Z\le\frac m{8BR}.
\]
Thus $(R,q)$ is one of the scales included in
$\dataspace_{d,m}$. We also have
$N\ge AR^2q^a\log q$, so \cref{lem:hs-scale} gives
\[
  t\ge b\,\frac{R\log q}{\varepsilon}.
\]
It remains to show that $\log q$ is large enough. By maximality, $2q$ violates
\eqref{eq:hs-sample-alphabet-choice}, and therefore $
  Z<(2q)^a\log(2q)=2^{a(s+1)}(s+1).$
Using $Z\ge2$ and $\log(s+1)\le s$, we obtain
$
  \log(2+Z)
  \le1+\log Z
  <1+a+(a+1)s
  \le2(a+1)s.
$
Also, $R\le\kappa d$ implies $Z\ge c_ZX$ for a universal
$c_Z>0$, and hence
$\log(2+Z)\ge c'\log(2+X)$ for a universal $c'>0$.
Combining these inequalities gives
$\log q\ge c_q\log(2+X)$, where
$c_q:=c'/(2(a+1))$.

Finally, $R>\kappa d/2$, so
\[
  t\ge b\,\frac{R\log q}{\varepsilon}
  \ge\frac{b\kappa c_q}{2}\,
      \frac d\varepsilon\log(2+X).
\]
This gives the claimed bound for $N>d^2$.
Taking $c:=\min\{bc_r,b\kappa c_q/2\}$ completes the proof.
\end{proof}

The two expressions agree up to universal constants when
$n\varepsilon\asymp d^2$, where both give
$\Omega(d/\varepsilon)$. The theorem includes the resolution
$t=n$. As with the large-sample result, the sample-size
conditions are sufficient and are not asserted to be sharp.

\section{Additional Algorithms for Small-dataset Regime}

\begin{proof}[Proof of \cref{thm:hs-t-equals-n-mb}]
We write $\opt=\opt_{\hypclass_{\text{hs}}}(S_A,S_B)$. We first prove privacy and then turn to the utility proof.

We first prove the privacy guarantee. Fix the public signs. Replacing one record of Alice moves one unit of count from one ray to another, so it changes each $\sum_{R}\sigma_{j,R}a_R$ by at most two. Therefore, the vector of the $k$ sums has $\ell_1$-sensitivity at most $2k$, and by \cref{lem:laplace} and composition property of DP \citep{dwork2006calibrating}, Alice's release is $\epsilon$-DP. The same argument works for Bob. Since the public signs are independent of the datasets and the output is a post-processing of the two releases, the protocol satisfies $\epsilon$-two-party DP.

To show this algorithm achieves the claimed resolution, for $j\in[k]$, let $D_j=\sum_{R}\sigma_{j,R}(a_R-b_R)$ be the difference of the two sums before adding the noise. Since $k$ is a constant, the tail bound of the Laplace distribution and a union bound over the $2k$ noise terms show that, with probability at least $0.95$, all of them are $O(1/\epsilon)$ in absolute value. In the rest of the proof we assume that this event holds. Then, by choosing $C$ large enough, every noise term is at most $\sqrt{n}/8$, and hence $|Z_{A,j}-Z_{B,j}-D_j|\leq\sqrt{n}/4$ for every $j$.

Under the Far promise, \cref{eq:hs-t-equals-n-char} gives $a_R=b_R$ for every $R$. Thus $D_j=0$ and $|Z_{A,j}-Z_{B,j}|\leq\sqrt{n}/4<\tau$ for every $j$, and the algorithm outputs $\mathsf{Far}$.

Under the Match promise, \cref{eq:hs-t-equals-n-char} gives $a_Rb_R=0$ for every $R$. Let $V=\sum_{R}(a_R-b_R)^2$. Since the counts are integers with disjoint supports,
\[
V=\sum_{R}(a_R^2+b_R^2)\geq\sum_{R}(a_R+b_R)=2n.
\]
The signs are independent, so $\EE[D_j^2]=V$ and $\EE[D_j^4]\leq 3V^2$. By the Paley--Zygmund inequality,
\[
\Pr\left[D_j^2\geq V/2\right]\geq\frac{(\EE[D_j^2])^2}{4\,\EE[D_j^4]}\geq\frac{1}{12},
\]
and in particular $|D_j|\geq\sqrt{n}$ with probability at least $1/12$. The $k$ repetitions are independent, so if $k$ is large enough, with probability at least $0.95$ there exists $j\in[k]$ with $|D_j|\geq\sqrt{n}$. For this $j$, we have $|Z_{A,j}-Z_{B,j}|\geq\sqrt{n}-\sqrt{n}/4>\tau$, and the algorithm outputs $\mathsf{Match}$.

In both cases, the failure probability is at most $0.1$, which completes the proof.
\end{proof}
\begin{proof}[Proof of \cref{thm:halfspace-query-release}]
First of all, by Sauer--Shelah lemma (see \cref{lem:finite-dual-discretization}), we can restrict $\hypclass_{\mathrm{hs}}$ to a finite discretization $\hypclass_0\subseteq\hypclass_{\mathrm{hs}}$ since $\dataspace$ is public and each party can find the projection of the hypothesis class on $\dataspace$.

For any finite multiset $D$ over $\dataspace$, define the local error at Alice and Bob as
\[
f_A(h;D):=\sum_{x\in D}\indic{h(x)\neq-1},
\qquad
f_B(h;D):=\sum_{x\in D}\indic{h(x)\neq+1}.
\]
Using this notation, we can rewrite the optimal error as
\[
\opt_{\hypclass_{\mathrm{hs}}}(S_A,S_B)
=
\min_{h\in\hypclass_0}
\bigl(f_A(h;S_A)+f_B(h;S_B)\bigr).
\]

Then, we show how we can use the mechanism of \citep{blum2013learning} to perform the test.
Fix a sufficiently large universal constant $c_0>0$, and set
\[
s:=
\left\lceil
\frac{c_0 n^2 d}{t^2}
\right\rceil
\qquad\text{and}\qquad
\mathcal{D}_s:=\dataspace^s.
\]
The family $\mathcal{D}_s$ is public and has cardinality $m^s$.
We claim that, for every party $P\in\{A,B\}$ and every
$S_P\in\dataspace^n$, some $D^\star_P\in\mathcal{D}_s$ satisfies
\[
\max_{h\in\hypclass_0}
\left|
f_P(h;S_P)-\frac{n}{s}f_P(h;D^\star_P)
\right|
\leq \frac{t}{16}.
\]

Indeed, view $S_P$ as defining an empirical distribution $\mathrm{Unif}(S_P)$ over $\dataspace$. Sample $s$ records independently and uniformly from $S_P$, with replacement, and let $D\in\mathcal{D}_s$ be the resulting multiset. Notice that $\frac{1}{n}f_P(h;S_P)$ is the population error of $h$ under $\mathrm{Unif}(S_P)$ and $\frac{1}{s}f_P(h;D)$ is its empirical average over the $s$ draws. Since the class of halfspaces $\hypclass_{\mathrm{hs}}$ in $\Reals^d$ has VC dimension $d$, standard VC uniform convergence guarantees that for a sufficiently large universal constant $c_0>0$, a sample of size $s=\lceil c_0 d/\alpha^2\rceil$ with $\alpha=t/(16n)$ satisfies
\[
\Pr\left[
\max_{h\in\hypclass_0}
\left|
\frac{1}{n}f_P(h;S_P)-\frac{1}{s}f_P(h;D)
\right|>\frac{t}{16n}
\right]
\leq\frac12.
\]
Multiplying both sides inside the probability by $n$ shows that such a $D^\star_P\in\mathcal{D}_s$ exists.

Then, we explain the protocol. Each party $P$ independently applies the exponential mechanism
with privacy parameter $\epsilon$ to the public family
$\mathcal{D}_s$, using the cost
\[
q_P(S_P,D):=
\max_{h\in\hypclass_0}
\left|
f_P(h;S_P)-\frac{n}{s}f_P(h;D)
\right|.
\]
Let $D_P$ be its released synthetic dataset. After exchanging
their releases, the parties compute
\[
\widehat{\opt}
:=
\frac{n}{s}
\min_{h\in\hypclass_0}
\bigl(f_A(h;D_A)+f_B(h;D_B)\bigr).
\]
They output $\mathsf{Match}$ if
$\widehat{\opt}\leq t/2$, and $\mathsf{Far}$ otherwise.

The privacy proof is straightforward:
For every fixed candidate $D$, changing one input record changes
$f_P(h;S_P)$ by at most one for every $h$. Therefore,
$q_P(S_P,D)$ has sensitivity at most one, and each release is
$\epsilon$-DP by \cref{lem:exponential-mechanism}.
The public parameters $n$ and $s$ remain fixed, so the same
sensitivity bound holds on the full domain of finite input
multisets.

Then, we prove the utility guarantee. By \cref{lem:exponential-mechanism}, for each party $P$, with
probability at least $19/20$,
\[
\begin{aligned}
q_P(S_P,D_P)
&\leq
\min_{D\in\mathcal{D}_s}q_P(S_P,D)
+\frac{2}{\epsilon}
\left(\log|\mathcal{D}_s|+\log20\right)\\
&\leq
\frac{t}{16}
+\frac{2}{\epsilon}\left(s\log m+\log20\right)\\
&\leq \frac{t}{8}.
\end{aligned}
\]
The last inequality follows from
\cref{eq:halfspace-blr-gap}, for a sufficiently large universal
constant $C$, using
$s=O(n^2d/t^2)$ and $t\leq n$.

By a union bound, both releases satisfy this guarantee with
probability at least $0.9$. On this event, the total error of
every halfspace is approximated within $t/4$. Taking the minimum
over $h$ therefore gives
\[
\left|
\widehat{\opt}
-\opt_{\hypclass_{\mathrm{hs}}}(S_A,S_B)
\right|
\leq\frac{t}{4}.
\]
If the pooled data are realizable, then
$\widehat{\opt}\leq t/4$, so the protocol outputs
$\mathsf{Match}$. If
$\opt_{\hypclass_{\mathrm{hs}}}(S_A,S_B)\geq t$, then
$\widehat{\opt}\geq3t/4$, so it outputs $\mathsf{Far}$.
This proves completeness and soundness.
\end{proof}

\end{document}

%% file: headers.tex
\pdfoutput=1
\usepackage{amsmath,amsthm,wrapfig}
	\usepackage[utf8]{inputenc} 
\usepackage{amsmath, amssymb,bm, cases, mathtools, thmtools}
\usepackage{verbatim}
\usepackage{graphicx}\graphicspath{{figures/}}
\usepackage{multicol}
\usepackage{tabularx}
\usepackage{mathrsfs} 
\usepackage{caption}
\usepackage{algorithm}
\usepackage{algorithmicx}
\usepackage[noend]{algpseudocode}
\usepackage{array,booktabs,arydshln,xcolor}

\usepackage[%
		minnames=1,maxnames=99,maxbibnames=99,minalphanames=1,maxalphanames=99,
		style=alphabetic,
		sorting=anyt,
		sortcites=false, %
		doi=false,url=false,
		uniquename=init,
		giveninits=true,
		hyperref,natbib,
		backend=biber]{biblatex}
\renewbibmacro{in:}{%
	\ifentrytype{article}{}{\printtext{\bibstring{in}\intitlepunct}}}

\usepackage[T1]{fontenc}
\usepackage{inconsolata}
\usepackage{dsfont}
\usepackage{hyperref}
\usepackage{listings} %

\usepackage{enumitem}
\usepackage{booktabs}       %
\usepackage{nicefrac}       %

\usepackage{lineno}
\usepackage{bbm}

\usepackage{caption}
\usepackage{subcaption}

\usepackage{datetime}

\DeclareMathAlphabet\EuRoman{U}{eur}{m}{n}
\SetMathAlphabet\EuRoman{bold}{U}{eur}{b}{n}

\declaretheorem[style=plain,numberwithin=section,name=Theorem]{theorem}

\declaretheorem[style=plain,sibling=theorem,name=Lemma]{lemma}

\declaretheorem[style=plain,sibling=theorem,name=Corollary]{corollary}

\declaretheorem[style=definition,sibling=theorem,name=Definition]{definition}
\declaretheorem[style=definition,qed=$\triangleleft$,sibling=theorem,name=Example]{example}
\declaretheorem[style=remark,qed=$\triangleleft$,sibling=theorem,name=Remark]{remark}
\numberwithin{theorem}{section}

\usepackage{xparse}
\usepackage{xstring}
\usepackage{xspace}

%% file: defs.tex
\newcommand{\dataspace}{\mathcal X}
\newcommand\optparen[1]{\ifthenelse{\equal{#1}{}}{}{(#1)}}

\newcommand{\Naturals}{\mathbb{N}}

\newcommand{\Reals}{\mathbb{R}}

\newcommand{\conv}{\textrm{conv}}

\DeclareMathOperator{\supp}{supp}

\DeclareMathOperator*{\newlim}{\mathrm{lim}\vphantom{\mathrm{infsup}}}
\DeclareMathOperator*{\newmin}{\mathrm{min}\vphantom{\mathrm{infsup}}}
\DeclareMathOperator*{\newmax}{\mathrm{max}\vphantom{\mathrm{infsup}}}
\DeclareMathOperator*{\newinf}{\mathrm{inf}\vphantom{\mathrm{infsup}}}

\renewcommand{\lim}{\newlim}
\renewcommand{\min}{\newmin}
\renewcommand{\max}{\newmax}
\renewcommand{\inf}{\newinf}

\def\EE{\mathbb{E}}

\newcommand{\hypclass}{\mathcal{H}}

\newcommand{\e}{\mathrm{e}}
\newcommand{\id}[1]{\mathbb{I}_{#1}\xspace}

\newcommand{\Alg}{\mathcal{A}}

\newcommand{\lcrx}[4][{-1}]{
	\IfEq{#1}{-1}{\left #2 {{{{#3}}}} \right #4}{
   	\IfEq{#1}{0}{#2 {{{{#3}}}} #4}{
	\IfEq{#1}{1}{\bigl #2 {{{{#3}}}} \bigr #4}{
	\IfEq{#1}{2}{\Bigl #2 {{{{#3}}}} \Bigr #4}{
	\IfEq{#1}{3}{\biggl #2 {{{{#3}}}} \biggr #4}{
	\IfEq{#1}{4}{\Biggl #2 {{{{#3}}}} \Biggr #4}{
    \GenericWarning{"4th argument to lcrx must be -1, 0, 1, 2, 3, or 4"}
    }}}}}}}
\newcommand{\inner}[3][{-1}]{\lcrx[#1] < {{#2},{#3}} >}

\newcommand{\indic}[1]{\mathbf{1}\left\{#1\right\}}

\renewcommand{\epsilon}{\varepsilon}

\newcommand{\Field}{\mathbb{F}}
\newcommand{\Tr}{\operatorname{Tr}}
\newcommand{\hs}{\text{hs}}
\newcommand{\code}{\mathcal{C}}

%% file: defs-valuation.tex
\newcommand{\Lap}{\mathrm{Lap}}
\newcommand{\opt}{\mathrm{opt}}

\newcommand{\PCF}{\mathsf{PCF}}
\newcommand{\prefix}{\mathrm{Prefix}}